\documentclass{article} % For LaTeX2e
\usepackage{iclr2027_conference,times}

\usepackage{amsmath,amsfonts,bm}

\def\eqref#1{equation~\ref{#1}}
\def\1{\bm{1}}

\DeclareMathAlphabet{\mathsfit}{\encodingdefault}{\sfdefault}{m}{sl}
\SetMathAlphabet{\mathsfit}{bold}{\encodingdefault}{\sfdefault}{bx}{n}

\usepackage{comment}  % to use comment feature
\usepackage{paralist} % use \begin{compactitem}
\usepackage{booktabs}
\usepackage[table]{xcolor}
\usepackage{amssymb}
\usepackage{amsmath}
\usepackage{multirow}
\usepackage{enumitem}
\usepackage{tikz}
\usepackage{pgfplots}
\usepackage{subcaption}
\usepackage{xcolor}
\usepackage{colortbl}
\usepackage[table]{xcolor}
\usepackage{tabularx}
\usepackage{hyperref}

\usepackage{listings}
\usepackage{booktabs}
\usepackage{booktabs}
\usepackage[table]{xcolor}
\usepackage{amssymb}
\usepackage{amsmath}
\usepackage{multirow}
\usepackage{enumitem}
\usepackage{tikz}
\usepackage{pgfplots}
\usepackage{subcaption}
\usepackage{xcolor}
\usepackage{colortbl}
\usepackage{booktabs}
\usepackage[table]{xcolor}
\usepackage{tabularx}
\usepackage{booktabs}
\usepackage{epigraph}
\usepackage[table]{xcolor}
\usepackage{amssymb}   
\usepackage{pifont}    
\usepackage{wrapfig}   % 提供 wraptable 环境
\usepackage{graphicx}  % 提供 \resizebox
\usepackage{booktabs}  % 提供 \toprule、\midrule、\bottomrule
\usepackage[table]{xcolor} % 提供 \rowcolor 和 \cellcolor

\usepackage{longtable}
\usepackage{fvextra}
\usepackage{placeins}

\definecolor{meiSage}{HTML}{E5EFEB}
\definecolor{meiBlue}{HTML}{648CAF}
\definecolor{meiRose}{HTML}{BD7B89}
\definecolor{meiRule}{HTML}{A9B6BA}

\definecolor{srtlightgray}{RGB}{243,244,246}
\definecolor{srtmidgray}{RGB}{229,231,235}
\definecolor{srtblue}{RGB}{30,100,200}
\definecolor{srtpurple}{RGB}{88,28,135}
\definecolor{srtteal}{RGB}{8,80,65}
\definecolor{srtamber}{RGB}{99,56,6}
\definecolor{srtpink}{RGB}{114,36,62}
\definecolor{srtgreen}{RGB}{39,80,10}
\definecolor{srtcoral}{RGB}{113,43,19}

\definecolor{audiocolor}{HTML}{DDF0F7}   % 浅蓝灰，audio context
\definecolor{neutralcolor}{HTML}{D6EAF8} % 蓝，neutral probe
\definecolor{biascolor}{HTML}{FADBD8}    % 红，biased probe
\definecolor{answercolor}{HTML}{D5F5E3}  % 绿，正确答案
\definecolor{metacolor}{HTML}{F2F3F4}    % 浅灰，元信息行
\usepackage{array}
\definecolor{tableheader}{RGB}{45, 62, 80}
\definecolor{tablerowalt}{RGB}{242, 245, 248}
\definecolor{headertext}{RGB}{255, 255, 255}
\definecolor{accentblue}{RGB}{31, 119, 180}
\definecolor{posstrong}{RGB}{198, 219, 239}
\definecolor{posmedium}{RGB}{107, 174, 214}
\definecolor{poslight}{RGB}{222, 235, 247}
\definecolor{neglight}{RGB}{252, 219, 199}
\definecolor{neutral}{RGB}{247, 247, 247}
\definecolor{srtblue}{RGB}{219,234,254}
\definecolor{srtlightgray}{RGB}{243,244,246}
\definecolor{srtdark}{RGB}{30,64,175}
\definecolor{srtorange}{RGB}{194,65,12}
\definecolor{srtgreen}{RGB}{6,95,70}
\definecolor{srtpurple}{RGB}{88,28,135}

\lstdefinestyle{prompt}{
  basicstyle=\ttfamily\footnotesize,
  breaklines=true,
  breakatwhitespace=false,
  columns=fullflexible,
  frame=single,
  framerule=0.3pt,
  xleftmargin=0.5em,
  xrightmargin=0.5em
}
\usepackage{subcaption}

\title{All the Facts, Still the Wrong Answer: \\ Integration Gaps in Vision-Language Models}

\author{Yuyang Dai$^{1}$, Yuxia Wang$^{2}$, Zhuohan Xie$^{3}$\\
$^{1}$Florida State University\\
$^{2}$INSAIT, Sofia University ``St. Kliment Ohridski''\\
$^{3}$MBZUAI\\
\texttt{y9657422@gmail.com}, \texttt{yuxia.wang@insait.ai}\\
\texttt{zhuohan.xie@mbzuai.ac.ae}
}

\iclrfinalcopy
\begin{document}

\maketitle
\lhead{Preprint}

\begin{abstract}
Multimodal reasoning requires models to combine multiple pieces of evidence into decisions that satisfy all relevant constraints. Evaluations centered on final-answer accuracy leave unclear whether failures arise from reading individual facts or using them together. 
We study this question through the multi-evidence
integration gap, linking local fact recovery to global decision validity on controlled subset-selection and path-finding tasks. Our evaluation combines conditional performance given correct local answers with same-grid local queries, text-oracle inputs, retrieval controls, and constraint ablations.
We further examine hidden-state probes and activation replacement across network depth, token positions, and model families. 
Across the evaluated models, accurate local recovery coexists with substantial global failure. For example, Gemma-3-27B recovers all evaluated isolated item facts correctly but achieves only 19.75\% global validity
on eight-item visual instances. Probes decode information about solution membership from global-prompt states, while corresponding clean-state replacement recovers failed decisions over model-dependent depth ranges. 
These findings show that recovering facts and representing solution-related information are insufficient for reliable joint decisions. They motivate evaluations and model improvements that explicitly target how available evidence is combined, alongside advances in perception and model scale. See our implementation details at: https://github.com/YUYANG0915/mech.
\end{abstract}

\section{Introduction}
\label{sec:introduction}

Understanding visual evidence requires more than recognizing its individual components. For example, a model may accurately identify each item's cost and compatibility relations but choose items that exceed the total budget or cannot be used together. We call an answer \emph{globally valid} when the selected combination satisfies all task constraints simultaneously. Producing such an answer requires both recovering individual facts and using them jointly. This raises a fundamental question for multimodal reasoning: \emph{when a model can correctly recover each piece of visual evidence, how reliably can it combine those pieces into a globally valid answer?}

Existing research provide complementary perspectives on this question. Broad multimodal benchmarks assess knowledge and reasoning across diverse domains \citep{yue2024mmmu}, while multi-image benchmarks examine comparison, relational understanding, and reasoning over image collections and sequences \citep{wang2025muirbench,liu2024mibench,meng2025mmiu,wang2024mementos}. Research on visual composition further shows that recognizing familiar concepts does not ensure sensitivity to their relations or combinations \citep{yuksekgonul2022and,ma2023crepe}. Hierarchical evaluations such as ConvBench explicitly distinguish perception from higher-level capabilities \citep{liu2024convbench}. Together, these studies motivate separating evidence recovery from evidence use. Nevertheless, for joint constraint satisfaction explaining an incorrect answer requires connecting three observations: whether the constituent facts are recoverable, whether the resulting decision is globally valid, and whether interventions on internal states can change that decision.

To fully address this, we propose \emph{multi-evidence integration gap}: the discrepancy between successful recovery of constituent evidence and successful formation of a globally valid decision. Alongside unconditional performance, we evaluate global validity on instances for which the same model correctly answers every constituent local question. This conditional evaluation identifies global failures that coexist with demonstrated local competence. Crucially, local questions and global decisions are separate computations; successful local recovery does not ensure that all facts are correctly represented or jointly used during global reasoning. This gap motivates a question: \emph{what separates recoverable evidence from a globally valid decision?}

To investigate this, we construct a controlled visual selection task with specified facts and verified solutions. Models select subsets of evidence cards subject to numerical, categorical, and compatibility constraints. We first compare global decisions with local fact recovery, including a control that presents the complete visual grid during local questioning. We then examine alternative explanations through matched-retrieval tasks, constraint ablations, and text-oracle inputs. Comparisons across eight checkpoints spanning 3B to 32B parameters test whether the phenomenon extends across model families and scales, while a separate path-finding task examines its scope beyond subset selection. These findings motivate evaluating multimodal reasoning through the relationship between recoverable evidence, represented information, and globally consistent decisions.
Our contributions are threefold:
\begin{itemize}
    \item We develop a framework that links local evidence recovery to exactly verifiable global decisions, including conditional evaluation under matched visual presentation.
    \item We characterize the multi-evidence integration gap across model families and scales, testing its scope through retrieval controls, constraint ablations, modality comparisons, and a second structured reasoning task.
    \item We combine representation probes with controlled activation interventions to distinguish solution-related information from successful decisions and characterize recovery effects across network depth, token scopes, and evaluated checkpoints.
\end{itemize}

\begin{figure*}[!t]
\centering
\includegraphics[width=\textwidth]
{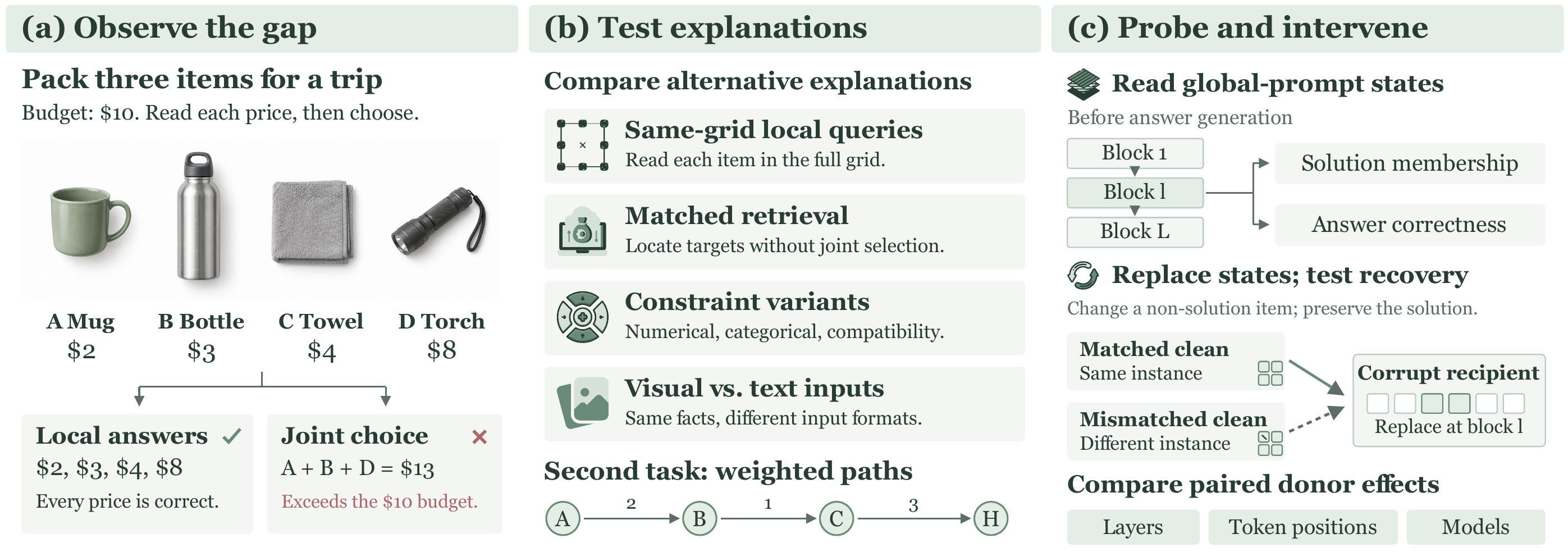}
\caption{\textbf{Overview of our study.}
(a) A packing example illustrates the local-to-global gap: every price is read correctly, yet the selected combination exceeds the budget.
(b) Same-grid local queries, matched retrieval, constraint variants, and visual/text inputs examine possible explanations for the gap; a weighted-path task tests whether it extends beyond subset selection.
(c) Probes assess solution membership and answer correctness in global-prompt hidden states.}
\vspace{-4mm}
\label{fig: overview}
\end{figure*}
\section{Related Work}
\label{sec:related_work}

\paragraph{Multimodal Evaluation.}
Benchmarks such as MMMU assess broad multimodal understanding and reasoning \citep{yue2024mmmu}. Multi-image evaluations extend this scope to relational understanding, comparison, and reasoning across visual inputs \citep{wang2025muirbench,liu2024mibench,meng2025mmiu}, including image sequences \citep{wang2024mementos}. ConvBench further separates perception from higher-level capabilities through hierarchical evaluation \citep{liu2024convbench}. Our work complements these approaches by examining global constraint satisfaction conditional on successful constituent fact recovery, and connecting this behavioral distinction to internal interventions.

\paragraph{Compositional Reasoning and Binding.}
Prior studies expose weaknesses in vision-language models' sensitivity to compositional structure, object relations, and spatial configurations \citep{yuksekgonul2022and,ma2023crepe,kamath2023s}. These findings motivate distinguishing recognition of individual concepts from their coordinated use. Research on contextual entity binding and propositional probes additionally investigates how language models represent associations and world states \citep{feng2024language,feng2025monitoring}. We examine whether individually recoverable visual facts jointly constrain a valid decision, rather than assuming that local recognition establishes successful integration.

\paragraph{Long-Context Evidence Use.}
Long-context evaluations identify difficulties in retrieving and combining distributed information, including sensitivity to evidence position \citep{liu2024lost,bai2024longbench,wang2024leave}. NoLiMa examines retrieval beyond literal matching \citep{modarressi2025nolima}, while MileBench studies multimodal long-context capabilities \citep{song2024milebench}. These directions highlight that increasing context does not guarantee effective evidence use. Our conditional evaluation focuses on the relationship between demonstrated local competence and joint validity; the observed gap alone does not exclude retrieval difficulties during global reasoning.

\paragraph{Mechanistic Analysis and Interventions.}
Mechanistic studies investigate factual recall and latent reasoning in language models \citep{geva2023dissecting,yang2024large}, and visual information processing in vision-language models \citep{neo2025towards}. Particularly relevant, delimiter-token scaling targets cross-image information leakage through hidden-state interventions \citep{lee2026enhancing}. We study exactly verifiable joint decisions using matched and mismatched activation donors, comparing recovery across layers and token scopes. Following work on patching methodology and interpretability pitfalls \citep{zhang2024towards,makelov2024subspace}, we distinguish probe decodability from causal use and interpret the repair horizon as an intervention-dependent recovery window, not the location where an error necessarily originates.
\section{Exploratory Study}
\label{sec:exploratory}

We investigate whether accurate recovery of individual visual facts is accompanied by successful joint decision-making. We first establish a local--global performance gap, examine whether it persists under matched visual presentation, and test its sensitivity to input modality and model scale. These comparisons motivate further diagnosis without attributing the observed gap to a specific internal mechanism.

\paragraph{Dataset Construction and Evaluation Setup}
\label{sec:controlled_task}

We construct a synthetic selection task in which each candidate is represented by a visual card containing numerical attributes, categorical attributes, and compatibility relations. Given a grid of cards, the model must select a subset satisfying a prescribed selection size, cost and weight budgets, a color requirement, type-count limits, and mutual compatibility. Ground-truth facts and valid subsets are obtained from the structured generator and exhaustive enumeration. An answer is considered globally valid if it satisfies all constraints.

The study contains 1,600 instances, with 400 at each candidate count $k\in\{2,4,6,8\}$. Each instance has a visual-grid version and a text-oracle version that presents its underlying facts as text while preserving the decision requirements. We evaluate eight checkpoints: Qwen2.5-VL-3B/7B/32B, Gemma-3-4B/12B/27B, InternVL3-8B, and LLaVA-OneVision-7B. The full-grid local-recovery control covers Qwen2.5-VL-3B, Gemma-3-4B, InternVL3-8B, and LLaVA-OneVision-7B. Generation and prompting details appear in the appendix. Increasing $k$ changes both the amount of evidence and the combinatorial decision problem; the requested subset size is two at $k=2$ and three otherwise. These comparisons therefore do not isolate a pure context-length effect.
Construction and scoring details are provided in Appendix~\ref{app:selection_instances}, with prompt examples in Appendix~\ref{app:prompts}.

\begin{figure*}[t]
    \centering
    \begin{subfigure}[t]{0.245\textwidth}
        \centering
        \includegraphics[width=\linewidth]{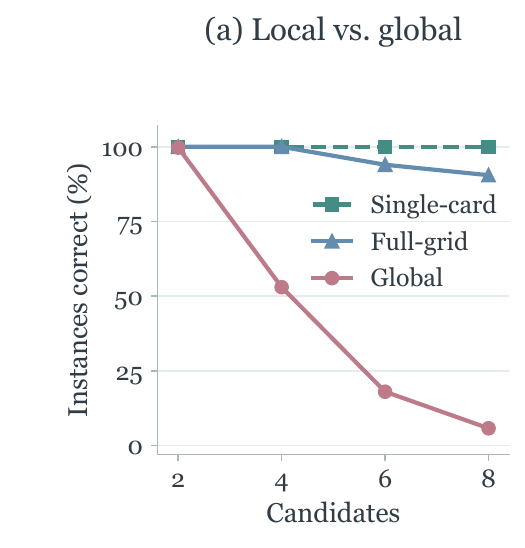}
    \end{subfigure}\hfill
    \begin{subfigure}[t]{0.245\textwidth}
        \centering
        \includegraphics[width=\linewidth]{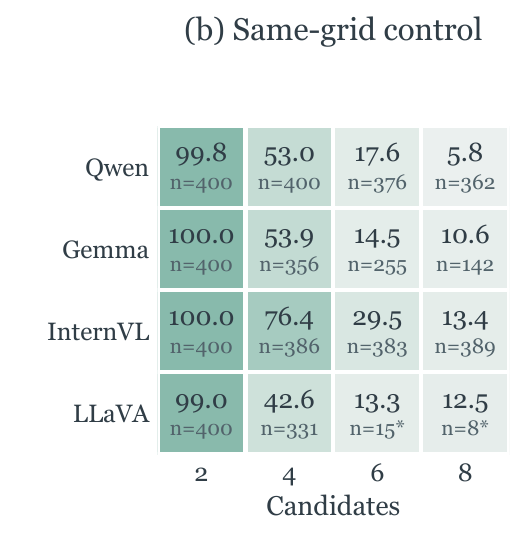}
    \end{subfigure}\hfill
    \begin{subfigure}[t]{0.245\textwidth}
        \centering
        \includegraphics[width=\linewidth]{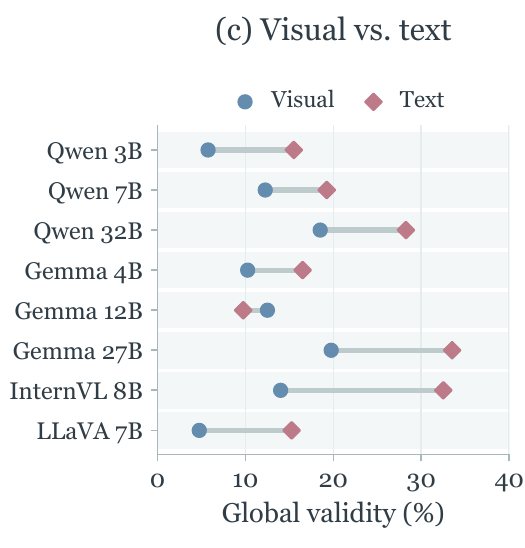}
    \end{subfigure}\hfill
    \begin{subfigure}[t]{0.245\textwidth}
        \centering
        \includegraphics[width=\linewidth]{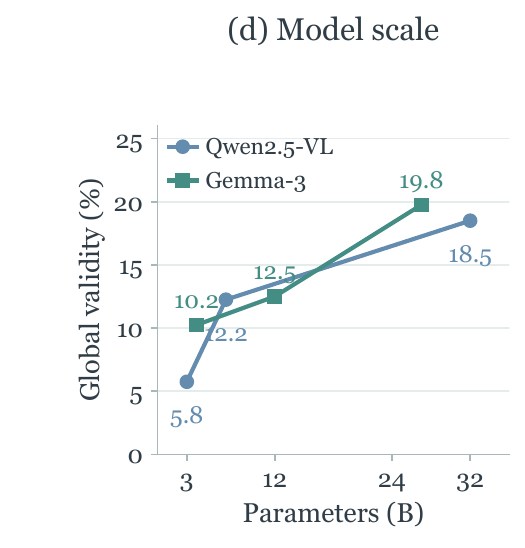}
    \end{subfigure}
    \caption{\textbf{Recoverable evidence does not guarantee globally valid decisions.}
    (a) Qwen2.5-VL-3B: instance-level local recovery under single-card and full-grid queries, compared with unconditional visual global validity.
    (b) Global validity conditioned on every full-grid local query being correct. Cells report percentages and retained sample counts; asterisks indicate $n<20$. Rows correspond to Qwen2.5-VL-3B, Gemma-3-4B, InternVL3-8B, and LLaVA-OneVision-7B.
    (c) Visual and text-oracle validity across eight checkpoints at $k=8$.
    (d) Within-family visual validity across model sizes at $k=8$; the vertical axis spans 0--26\%.
    Panels (a), (c), and (d) use 400 instances per condition. All rates are point estimates.}
    \vspace{-4mm}
    \label{fig:exploratory}
\end{figure*}

\paragraph{Local Recovery and Global Validity}
We distinguish local evidence recovery from global decision-making. In the single-card condition, the model extracts attributes from one card at a time. In the full-grid condition, it answers the same local queries while viewing the complete grid. Local recovery is scored at the instance level: every evaluated field on every constituent card must be correct. Global validity instead requires the selected subset to satisfy all task constraints.
Figure~\ref{fig:exploratory}(a) shows a substantial separation between these capabilities. Qwen2.5-VL-3B correctly recovers all 8,000 cards in isolation. Full-grid instance-level recovery also remains high, reaching 90.5\% at $k=8$. Nevertheless, visual global validity falls from 99.75\% at $k=2$ to 5.75\% at $k=8$. Thus, the ability to answer local questions about the evidence does not guarantee a valid joint decision. However, aggregate recovery rates alone, do not reveal whether global failures are concentrated among instances with local errors.
Appendix~\ref{app:local_recovery} reports field-level recovery and single-card conditional results (Tables~\ref{tab:app_local} and~\ref{tab:app_singlecond}).

\paragraph{Same-Grid Control}
To examine this possibility, we align local and global predictions by instance identity within each model and measure
\begin{equation}
    \Pr\!\left(
        \text{global valid}
        \mid
        \text{all full-grid local facts correct}
    \right).
\end{equation}
This comparison retains only instances for which every local query is answered correctly from the same grid used for the global task.
The gap persists within these subsets (Figure~\ref{fig:exploratory}b). At $k=8$, Qwen2.5-VL-3B achieves only 5.80\% global validity among 362 locally correct instances; InternVL3-8B achieves 13.37\% among 389. These results show that isolated-card presentation cannot fully explain the observed separation. The retained subsets nevertheless differ substantially across models: Gemma-3-4B retains 142 instances, whereas LLaVA-OneVision-7B retains only eight, making the latter estimate particularly uncertain. Moreover, local and global prompts elicit separate computations. This control establishes successful local recovery under matched visual presentation, not that every fact is correctly represented throughout the global decision process.
Full-grid attribute-level results are provided in Appendix~\ref{app:local_recovery}, Table~\ref{tab:app_grid}.

\paragraph{Modality and Scale}
We next examine whether changing the evidence format or increasing model scale resolves the difficulty. Figure~\ref{fig:exploratory}(c) compares visual and text-oracle validity at $k=8$. Textual presentation improves performance for most evaluated checkpoints, but substantial errors remain, and the advantage is not universal: Gemma-3-12B performs worse with text than with visual input. The results suggest that visual processing contributes to some failures without providing a complete explanation of the decision gap.
Larger checkpoints also improve visual validity at $k=8$ within both model families (Figure~\ref{fig:exploratory}d), but do not close the gap. Qwen2.5-VL-32B and Gemma-3-27B both achieve 100\% single-card fact recovery, yet their visual global validity is only 18.50\% and 19.75\%, respectively. Their text-oracle validity reaches 28.25\% and 33.50\%. These comparisons show that the phenomenon extends beyond the smallest checkpoints, while leaving room for further gains from scale.
Appendix~\ref{app:behavior_scale} provides the remaining candidate counts and generation-setting breakdowns (Tables~\ref{tab:app_global} and~\ref{tab:app_strata}).

Together, these observations establish a persistent gap between local evidence recovery and globally valid decisions in our controlled tasks. The gap remains on instances for which every local query is answered correctly from the full grid, and neither textual presentation nor increased model scale eliminates it. Its source remains unresolved: global decisions may require more difficult retrieval, joint constraint satisfaction, or preservation of an appropriate decision during generation.

\section{Behavioral Explanation}
\label{sec:behavioral}

Section~\ref{sec:exploratory} shows that the local--global gap survives a same-grid control and remains in larger checkpoints, yet these observations do not explain which demands of the global task contribute to failure. We now examine evidence localization, constraint satisfaction, and task specificity. We first ask whether models can retrieve the target items when joint selection is removed. We then restore the selection requirement while varying the active constraints, before testing whether the behavioral separation extends to path finding. These experiments cover Qwen2.5-VL-3B/7B and Gemma-3-4B/12B under visual and text-oracle inputs.

\begin{figure*}[t]
    \centering
    \begin{minipage}[t]{0.328\textwidth}
        \centering
        \includegraphics[width=\linewidth]
        {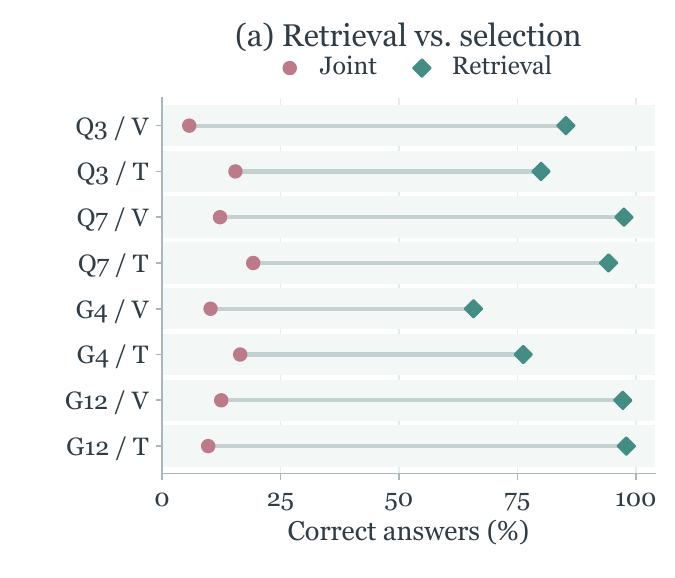}
    \end{minipage}\hfill
    \begin{minipage}[t]{0.328\textwidth}
        \centering
        \includegraphics[width=\linewidth]
        {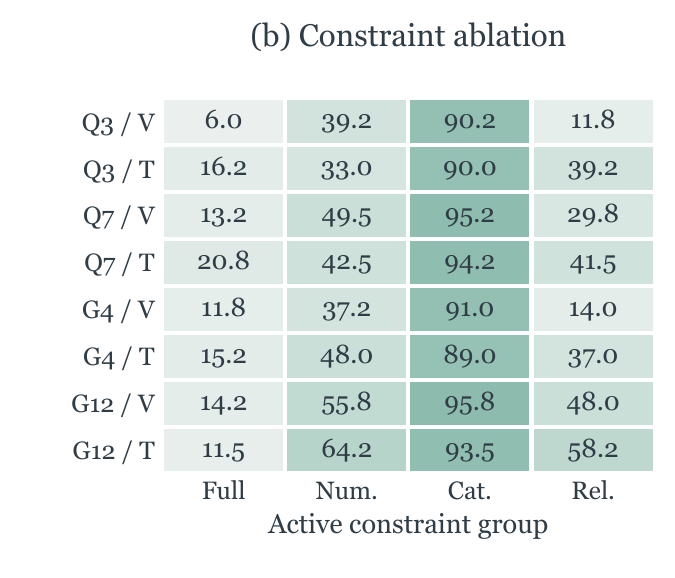}
    \end{minipage}\hfill
    \begin{minipage}[t]{0.328\textwidth}
        \centering
        \includegraphics[width=\linewidth]
        {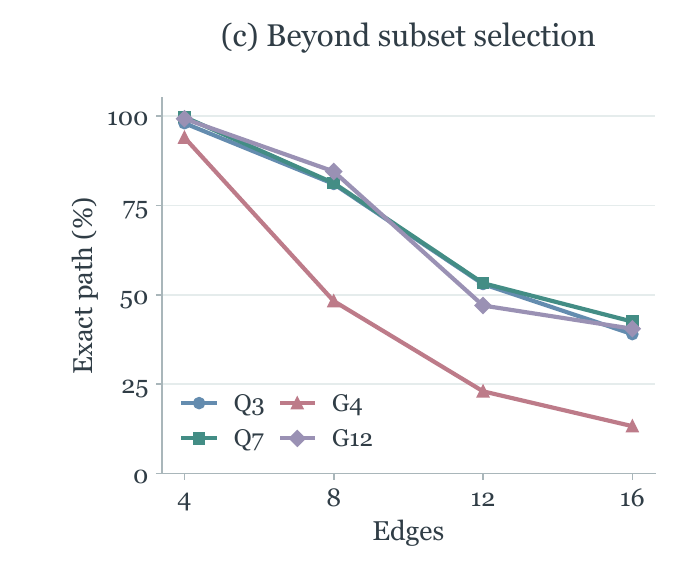}
    \end{minipage}
    \caption{\textbf{Behavioral diagnostics of the multi-evidence integration gap.}
    Q3 and Q7 denote Qwen2.5-VL-3B and Qwen2.5-VL-7B;
    G4 and G12 denote Gemma-3-4B and Gemma-3-12B.
    V and T indicate visual-grid and text-oracle inputs, respectively.
    \textit{(a) Retrieval versus joint selection.}
    At $k=8$, models more reliably retrieve target items from identifying attribute descriptions than select a combination satisfying all constraints in the original task.
    \textit{(b) Constraint ablation.}
    Cells report valid-answer rates (\%) at $k=8$;
    darker green indicates higher performance.
    Full retains all constraint groups, whereas Num., Cat., and Rel. retain only numerical, categorical, and compatibility constraints, respectively.
    \textit{(c) Beyond subset selection.}
    Visual-input exact-path accuracy decreases as the number of graph edges increases, although the target path remains three edges long.
    Each plotted condition contains 400 instances.
    }
    \vspace{-4mm}
    \label{fig:behavioral_explanation}
\end{figure*}

\paragraph{Retrieving Evidence versus Selecting a Valid Combination}
To distinguish evidence retrieval from joint selection, we construct a matched-retrieval variant for each of the 1,600 selection instances. The evidence, target item set, and required output size are preserved. We provide attribute-based descriptors that uniquely identify the target items and ask it to return their IDs. Accuracy measures exact recovery of this target set. Thus, the control retains evidence localization and the item-ID output format while removing the need to determine which combination satisfies the constraints.
Retrieval is substantially easier than joint selection (Figure~\ref{fig:behavioral_explanation}a). At eight candidates, visual retrieval accuracy ranges from 65.75\% to 97.50\%, whereas joint-selection validity ranges from 5.75\% to 12.50\%. For Qwen2.5-VL-7B, the corresponding rates are 97.50\% and 12.25\%; for Gemma-3-12B, they are 97.25\% and 12.50\%. The separation also occurs with text input. Consequently, difficulty locating items and returning their IDs is insufficient to account for the full decision gap.
The retrieval construction and example prompt appear in Appendices~\ref{app:diagnostic_tasks} and~\ref{app:prompts}; smaller-candidate results appear in Appendix~\ref{app:behavior_diagnostics}, Table~\ref{tab:app_retrieval}. 

\paragraph{Decomposing the Constraint Requirements}
Having established that evidence can often be located without successfully solving the selection task, we next examine its constraint requirements. For each source instance, we generate four variants: the full constraint set, numerical budgets only, categorical requirements only, and mutual compatibility only. All variants preserve the evidence and required subset size. We exhaustively enumerate the valid subsets under each active constraint set and accept any satisfying subset. The four variants yield 6,400 instances per modality and model. Comparisons use the full-constraint variant within this diagnostic suite, whose wording differs slightly from the original prompt.
At eight candidates, visual validity is 90.25--95.75\% for categorical constraints, 37.25--55.75\% for numerical constraints, and 11.75--48.00\% for compatibility constraints, compared with 6.00--14.25\% under the full set (Figure~\ref{fig:behavioral_explanation}b). For example, Qwen2.5-VL-7B achieves 95.25\%, 49.50\%, and 29.75\% on the three individual groups, but only 13.25\% when they are combined. Under text-oracle inputs, categorical constraints likewise yield the highest accuracy and the full set the lowest for all four checkpoints at this candidate count; the relative difficulty of numerical and compatibility constraints varies. Models perform well when checking categorical constraints alone, but often fail when adding up costs or weights or checking whether selected items are compatible. However, removing constraints also makes more answers valid, so higher accuracy may reflect an easier answer criterion instead of better reasoning 
(See Appendix~\ref{app:diagnostic_tasks}, Table~\ref{tab:app_solutions}, Appendix~\ref{app:behavior_diagnostics}, Table~\ref{tab:app_ablation} for acceptance-set sizes and remaining candidates.)

\begin{wraptable}{r}{0.60\textwidth}
    \centering
    \begingroup
    \normalsize
    \captionsetup{font=normalsize}
    \definecolor{meiSage}{HTML}{E5EFEB}
    \definecolor{meiBlue}{HTML}{648CAF}
    \definecolor{meiRose}{HTML}{BD7B89}
    \definecolor{meiRule}{HTML}{A9B6BA}
    \arrayrulecolor{meiRule}
    \setlength{\tabcolsep}{4pt}
    \renewcommand{\arraystretch}{1.05}
    \vspace{-4mm}
    \caption{\textbf{Correct edge facts, incorrect paths.}
    Local: all-field accuracy over 16,000 isolated edge queries.
    Visual/Text: exact-path accuracy on 400 sixteen-edge
    instances. All values are percentages.}
    \label{tab:path_diagnostic}

    \begin{tabular*}{\linewidth}{@{\extracolsep{\fill}}lrrr@{}}
        \toprule
        \rowcolor{meiSage}
        Model & Local &
        \textcolor{meiBlue}{Visual} &
        \textcolor{meiRose}{Text} \\
        \midrule
        Qwen2.5-VL-3B & 100.00 & 39.00 & 27.75 \\
        Qwen2.5-VL-7B & 100.00 & 42.50 & 45.50 \\
        Gemma-3-4B    & 100.00 & 13.25 & 43.00 \\
        Gemma-3-12B   & 100.00 & 40.50 & 71.25 \\
        \bottomrule
    \end{tabular*}
    \endgroup
\end{wraptable}

\paragraph{Beyond Subset Selection}
These results characterize failures within subset selection.
We next ask whether models also recover individual facts correctly but fail to combine them in a different task: finding a minimum-cost path in a directed, weighted graph.
We generate 1,600 directed, weighted graph instances, with 400 each at 4, 8, 12, and 16 edge cards. Local queries recover individual edge attributes; the global task asks for the minimum-cost path between specified endpoints. We evaluate exact agreement with the generated target path and compare visual-grid and text-oracle inputs. Consequently, the target path always contains three edges, so larger graphs add distractors without requiring a longer path.
The edge-cost and identifier construction is specified in Appendix~\ref{app:diagnostic_tasks} (Weighted paths), with task prompts in Appendix~\ref{app:prompts}.

All four models recover every evaluated local field across 16,000 isolated edge cards. Nevertheless, visual exact-path accuracy declines from 94.00--99.75\% at four edges to 13.25--42.50\% at sixteen edges (Figure~\ref{fig:behavioral_explanation}c). Table~\ref{tab:path_diagnostic} compares modalities at the largest graph size. Text helps some checkpoints substantially, including Gemma-3-12B, but reduces performance for Qwen2.5-VL-3B. Thus, the separation extends beyond selecting item subsets, while its magnitude and modality dependence vary with the model and task. Because local edge recovery is evaluated on isolated cards, this experiment does not independently exclude difficulties caused by presenting the full graph.
Appendix~\ref{app:behavior_diagnostics} provides the remaining text-path results, parsing rates, and per-field local accuracy. A stricter construction with randomized edge identifiers, overlapping cost ranges, and same-grid local queries yields the same separation in the supplementary materials.

\paragraph{Implications for Internal Analysis}
These controls narrow the behavioral explanation of the gap. Models often succeed when the target evidence is identified for them, their performance depends strongly on the active constraint requirements, and failures of global decision-making also occur in path finding despite perfect isolated evidence recovery. None of these comparisons establishes whether the global computation has represented a valid solution or where that computation fails. Thus, we turn to hidden-state probes and activation interventions to distinguish solution-related information from successful decisions and test when state replacement can change the outcome.

\section{Internal Representations and Causal Interventions}
\label{sec:internal}

In this section, we separate two questions: what information can be decoded from hidden states before generation, and whether replacing internal states can recover a correct decision.

\subsection{What Can Be Decoded from Hidden States?}

\paragraph{Two Readouts of the Same Prompt States}
We extract the hidden state at the final prompt token across layers, before any answer tokens are generated. We train linear logistic-regression probes for two targets. A correctness probe predicts whether the model's subsequent answer is globally valid. A membership probe predicts whether a specified candidate ID belongs to the unique valid subset. We train a separate membership classifier for each candidate ID and average AUC across candidates. Thus, the membership analysis tests information about individual solution components, not whether one decoder reconstructs a complete valid answer. Both analyses use global-task prompts, with visual-grid and text-oracle inputs evaluated separately.

\paragraph{Separating Difficulty from Decodability}
Because success rates differ substantially across candidate counts, we evaluate probes separately at $k=4,6,8$, with 400 instances per group. Standardization and classifier fitting use training data only. Five outer folds estimate held-out AUC; four inner folds select the layer without consulting the outer test fold. Membership classifiers split instances separately for each candidate ID, rather than pooling candidates from the same instance into independent examples. To assess sensitivity to the train--test split, we repeat nested cross-validation with five random seeds and report the mean AUC and its standard deviation across runs.
Appendix~\ref{app:probe_intervention_protocols}, Appendix~\ref{app:probe_sensitivity} provides probe protocol, the available per-seed scores and selected-layer records.

\begin{figure*}[!t]
\centering
\centering
\includegraphics[width=.248\textwidth]
{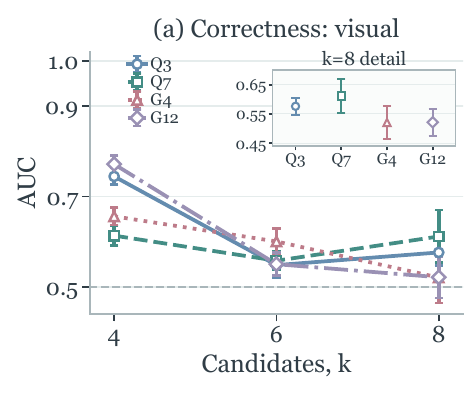}\hfill%
\includegraphics[width=.248\textwidth]
{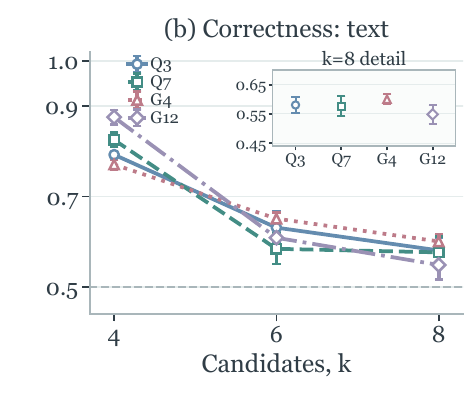}\hfill%
\includegraphics[width=.248\textwidth]
{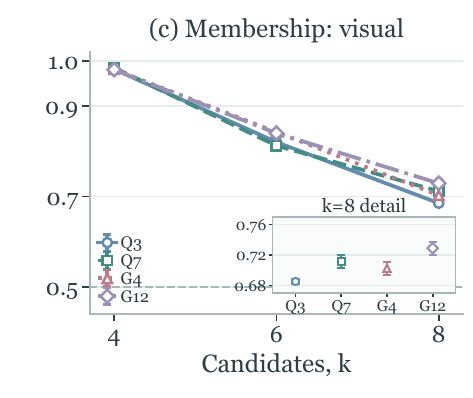}\hfill%
\includegraphics[width=.248\textwidth]
{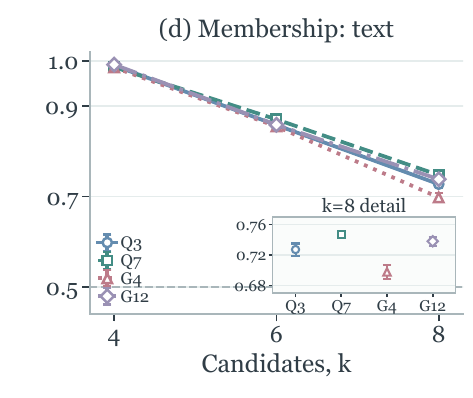}
\caption{\textbf{Decoding solution components and predicting answer correctness are different tests.}
Q3/Q7 denote Qwen2.5-VL-3B/7B; G4/G12 denote Gemma-3-4B/12B.
(a,b) Correctness AUC for visual-grid and text-oracle prompts.
(c,d) Candidate-membership macro-AUC for the same input conditions.
Points show means over five nested-cross-validation seeds; error bars show the standard deviation across seeds. Each candidate-count group contains 400 instances. Layers are selected within the training folds. The dashed line marks AUC $0.5$.}
\label{fig:internal_probes}
\end{figure*}

\paragraph{Solution Components Remain Decodable}
Figure~\ref{fig:internal_probes} shows that membership is more readily decoded than final correctness under these readouts. At $k=8$, visual membership macro-AUC ranges from $0.685$ to $0.729$, whereas visual correctness AUC ranges from $0.521$ to $0.611$. Text membership macro-AUC ranges from $0.698$ to $0.747$, compared with correctness AUC of $0.549$--$0.600$. Membership is much more decodable at $k=4$ and weakens as the candidate set grows. Larger checkpoints do not uniformly improve correctness prediction: for example, the visual correctness probe is weaker for Qwen-7B than Qwen-3B at $k=4$. In conclusion, the probes can identify which items belong to the correct solution more accurately than they can predict whether the model will answer correctly.
Candidate-specific membership results appear in Appendix~\ref{app:probe_sensitivity}; additional-model results and label controls appear in Appendix~\ref{app:probe_controls}. Appendix~\ref{app:layer_profiles} supplies the fixed-layer profiles.

\begin{wraptable}{r}{0.60\textwidth}
\centering
\begingroup
\normalsize
\definecolor{meiSage}{HTML}{E5EFEB}
\definecolor{meiRule}{HTML}{A9B6BA}
\arrayrulecolor{meiRule}
\setlength{\tabcolsep}{4pt}
\renewcommand{\arraystretch}{1.05}
\vspace{-4mm}
\caption{\textbf{Activation-patching coverage.}
Counts report evaluated clean-correct/corrupt-wrong pairs.
Layers gives the number of tested decoder blocks for
full-prompt replacement. Qwen2.5-VL-3B also receives image-
and text-token interventions at eight shared blocks on
the same pairs.}
\vspace{-2mm}
\label{tab:patch_coverage}
\begin{tabular*}{\linewidth}{@{\extracolsep{\fill}}lrrrr@{}}
\toprule
\rowcolor{meiSage}
Model & $k=4$ & $k=6$ & $k=8$ & Layers \\
\midrule
Qwen2.5-VL-3B & 99 & 100 & 100 & 13 \\
Qwen2.5-VL-7B & 37 & 15 & 10 & 9 \\
Gemma-3-4B & 34 & 8 & 6 & 14 \\
Gemma-3-12B & 47 & 15 & 13 & 9 \\
InternVL3-8B & 60 & 16 & 11 & 13 \\
\bottomrule
\end{tabular*}

\endgroup
\end{wraptable}

\paragraph{From Readout to Intervention}
A probe can exploit information that the model's own generation process does not use successfully. Moreover, independent membership predictions need not combine into a subset satisfying every constraint. Given this, we do not interpret these AUCs as evidence that the model has already solved the task internally. Instead, we ask whether states obtained from a corresponding correctly answered input can alter an incorrect decision.

\subsection{Can State Replacement Recover Correct Decisions?}

\paragraph{Solution-Preserving Input Pairs}
We modify one non-solution item to make it superficially attractive, with low cost and weight, high value, and the required color, while leaving it incompatible with the other items. Exhaustive enumeration verifies that the valid solution set is unchanged. We retain clean/corrupt pairs for which the model answers the clean input correctly and the corrupted input incorrectly. Table~\ref{tab:patch_coverage} reports the pairs actually represented in the intervention outputs. Recovery is measured on a selected, perturbation-sensitive subset.
Appendix~\ref{app:cohort_selection}, Table~\ref{tab:app_outcomes}, reports the clean/corrupt outcome counts before intervention subsampling.

\paragraph{Matched and Mismatched State Replacement}
For each recipient, a matched donor is its corresponding clean input. A mismatched donor is a different clean instance with the same candidate count and compatible prompt length; the evaluated donors have a different target subset. We replace decoder-block output states at the chosen prompt positions, keeping the corrupted recipient input and decoding procedure otherwise fixed. Generation is greedy with caching disabled: the selected prompt positions are replaced on every full-sequence forward pass, rather than only during initial prefill. Generated-token positions are not directly replaced. Full-prompt replacement patches every prompt position; restricted interventions patch only image-token positions or their non-image complement, which we call text tokens.

For a layer $\ell$ and position set $S$, we summarize the donor contrast as
\begin{equation}
\Delta_{\ell,S}=\frac{1}{N}\sum_{i=1}^{N}
\left(v_i^{\mathrm{matched}}(\ell,S)-v_i^{\mathrm{mismatched}}(\ell,S)\right),
\label{eq:donor_contrast}
\end{equation}
where $v_i$ indicates whether the patched answer is globally valid. Positive values mean that corresponding clean states recover valid answers more often than unrelated clean states. We use paired bootstrap resampling of recipient pairs for pointwise 95\% intervals. This contrast measures donor-specific recovery under the intervention.
Appendix~\ref{app:probe_intervention_protocols} provides the uncertainty protocol.

\paragraph{Recovery Depends on Intervention Depth}
For Qwen-3B at $k=4$, full-prompt matched replacement recovers 95.96\% of the selected failures at block 0, compared with 0\% for mismatched replacement (Figure~\ref{fig:internal_recovery}a). Recovery remains high through much of the tested depth range. At block 24, matched recovery is 87.88\%, but it falls to 3.03\% at block 27, where mismatched recovery is 2.02\%. Thus, the advantage of matched over mismatched replacement drops
sharply between blocks 24 and 27. Corresponding clean states can recover correct answers when introduced earlier, but provide little benefit at the later tested block.

\subsection{Which Token Positions Support Recovery?}

\begin{figure*}[!t]
\centering
\includegraphics[width=.328\textwidth]{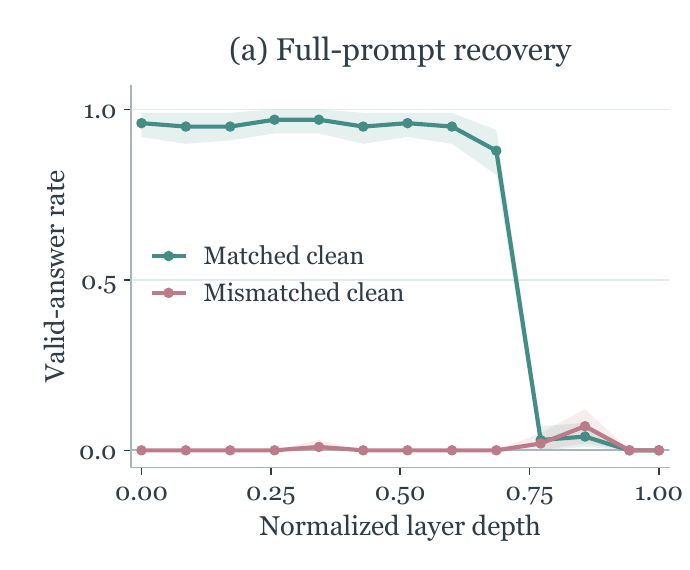}\hfill
\includegraphics[width=.328\textwidth]{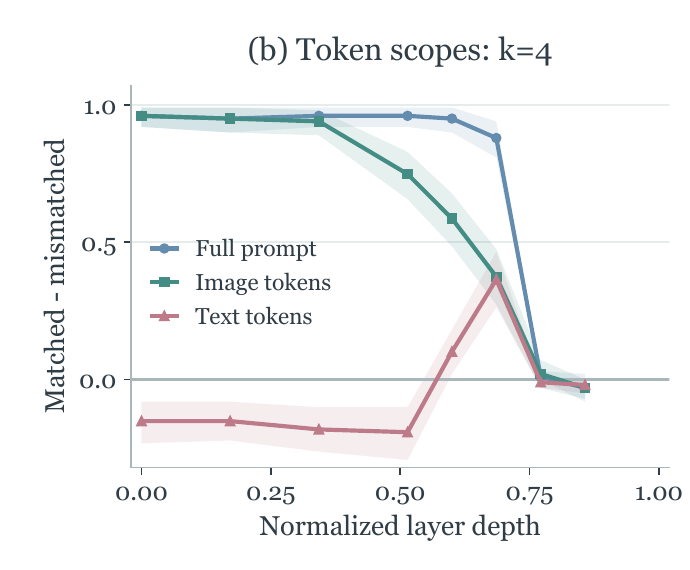}\hfill
\includegraphics[width=.328\textwidth]{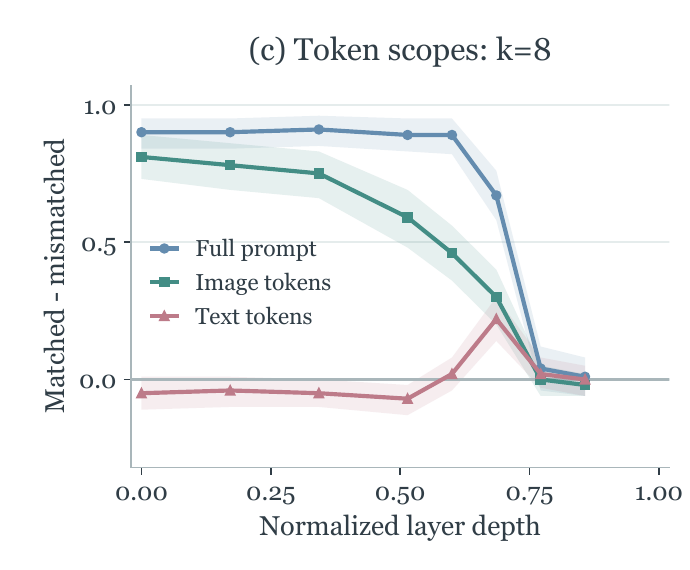}
\caption{\textbf{Recovery depends on both depth and replaced token positions.}
All panels use Qwen2.5-VL-3B with visual-grid inputs.
(a) Valid-answer rates after full-prompt matched or mismatched replacement at $k=4$.
(b,c) Matched-minus-mismatched valid-answer rates for three replacement scopes at $k=4$ and $k=8$, restricted to their shared tested blocks. Each scope uses the same 99 and 100 recipient pairs, respectively.
Depth is the zero-based decoder-block index divided by the final block index. Shading shows pointwise paired-bootstrap 95\% intervals; intervals in (a) are bootstrap intervals for each recovery rate.}
\vspace{-4mm}
\label{fig:internal_recovery}
\end{figure*}

\paragraph{Restricting the Replaced Positions}
Since full-prompt replacement does not distinguish contributions from different prompt positions, we then compare all-prompt, image-token, and text-token replacement on the same Qwen-3B pairs and shared tested blocks (Figure~\ref{fig:internal_recovery}b,c). At $k=4$, image-token replacement has a donor contrast of $0.960$ at block 0 and $0.374$ at block 24. Text-token replacement has no positive contrast at the earliest tested blocks, but reaches $0.364$ at block 24. At that block, full-prompt replacement yields a larger contrast of $0.879$. At $k=8$, image- and text-token contrasts at block 24 are $0.300$ and $0.220$, compared with $0.670$ for full-prompt replacement.

\paragraph{Position Is Not Information Modality}
The restricted interventions show that replacing image positions and replacing text positions have different depth profiles. Through attention, text-token states can also carry visual information. The later recovery from text-token replacement is consistent with image evidence becoming accessible at text positions. In Qwen-3B, image-token replacement improves recovery at early layers, whereas text-token replacement becomes beneficial later.
At block 24, replacing all prompt positions yields a larger
matched-donor advantage than replacing either group alone.
Appendix~\ref{app:token_scopes} completes the scope comparison at $k=6$ and reports changed-item avoidance.

\subsection{Do These Patterns Persist Across Models?}

\begin{wrapfigure}{r}{0.66\textwidth}
    \centering
    \includegraphics[width=.49\linewidth]
        {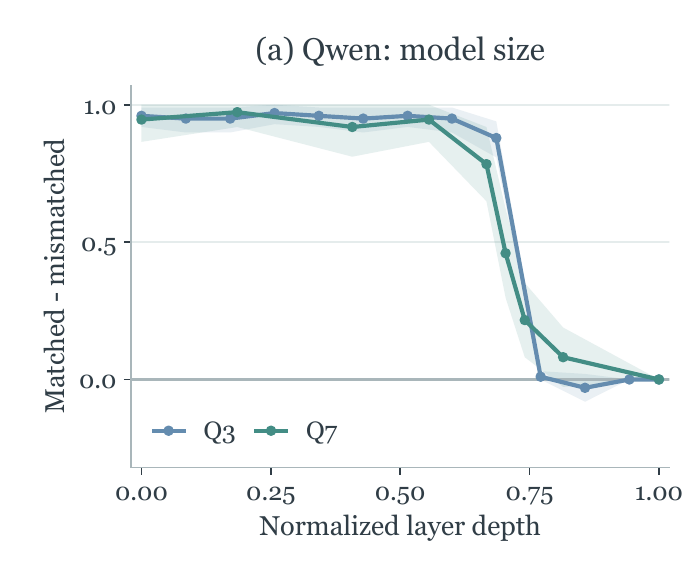}\hfill
    \includegraphics[width=.49\linewidth]
        {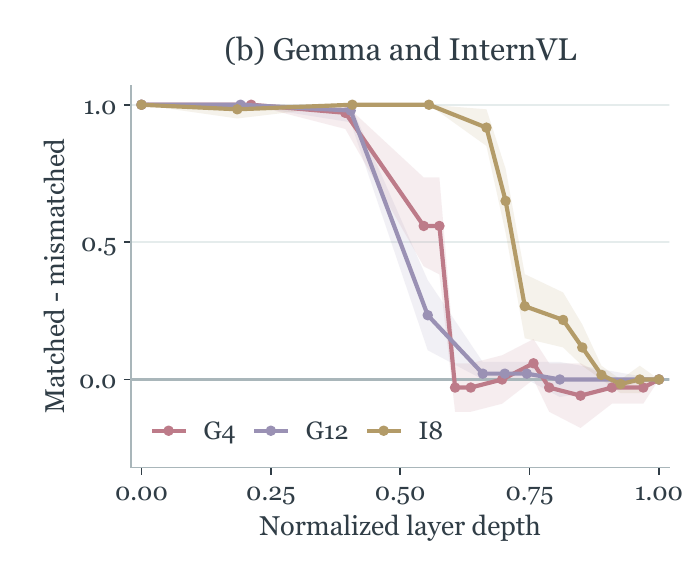}
    \vspace{-4mm}
    \caption{Full-prompt recovery profiles vary across model families and sizes. Both panels show matched-minus-mismatched valid-answer rates at $k=4$ with visual-grid inputs.
    (a) Q3/Q7: Qwen2.5-VL-3B/7B.
    (b) G4/G12: Gemma-3-4B/12B; I8: InternVL3-8B.
    Shading shows pointwise paired-bootstrap 95\% intervals.
    Normalized depth aligns relative block indices.}
    \vspace{-4mm}
    \label{fig:internal_models}
\end{wrapfigure}

\paragraph{Architecture and Scale}
We next compare full-prompt donor contrasts across model families and within Qwen and Gemma (Figure~\ref{fig:internal_models}). The $k=4$ comparison has the largest intervention cohorts for the additional models. Qwen-7B retains a positive point estimate of $0.216$ at normalized depth $0.741$, while the Qwen-3B contrast is near zero by its next tested depth, $0.771$. Gemma-4B changes from $0.559$ at depth $0.576$ to $-0.029$ at $0.606$; Gemma-12B has a contrast of $0.234$ at $0.553$ and $0.021$ at $0.660$. InternVL3-8B retains a positive lower pointwise interval bound through depth $0.852$. The recovery advantage persists into later layers in InternVL
than in Gemma, while increasing model size produces different changes in the Qwen and Gemma families.

\paragraph{Comparing Recovery Profiles}
We then compare how the matched-donor advantage changes across depth. For each model, the tested layers indicate the range over which replacement can recover correct answers. We focus the cross-model comparison on $k=4$, where the larger cohorts provide more stable estimates. At $k=6,8$, smaller cohorts make the estimated recovery
range more sensitive to individual examples.
Appendix~\ref{app:recovery_counts} presents the $k=6,8$ recovery profiles for all five models and their paired donor-contrast intervals.

\vspace{-0.2cm}
\paragraph{From Available Information to Successful Decisions}
Together, the analyses distinguish three observations: individual facts can be recovered, solution membership can be decoded from global-prompt states, and corresponding clean-state replacement can recover some failed decisions. The integration gap persists even when individual facts
are recoverable and solution-related information is accessible to probes. Probes describe information accessible to an external readout; interventions show that decisions are sensitive to particular state replacements. Their combination sharpens the integration gap without claiming that a complete solution was already represented or that one layer universally causes the failure.

\vspace{-0.1cm}
\section{Conclusion}
\label{sec:conclusion}

This study presents a multi-evidence integration gap in
vision-language models: accurate recovery of individual facts can coexist with decisions that violate their joint constraints. Experiments on subset selection and path finding reveal this separation, while comparisons across input modalities and model sizes show that its magnitude depends on both the task and the model. Hidden-state probes reveal decodable solution-membership information, and activation interventions show that corresponding clean states can recover failed decisions across model-dependent depths and token positions.
Extract--Solve further shows that structured evidence paired with deterministic execution can recover difficult global decisions. Together, these findings motivate multimodal evaluations and systems that connect evidence recovery to globally consistent decision-making.

\section*{AI Use Statement}

AI-assisted tools were used during manuscript preparation for language editing. All citations were selected and checked by the authors. The authors take full responsibility for the accuracy and integrity of the work.

\section*{Ethics Statement}

Our experiments use synthetic visual reasoning tasks and publicly available model checkpoints; they involve no human participants or personal data. The central risk highlighted by our results is practical: a model can read every relevant fact correctly and still reach a decision that violates their joint constraints. In settings where several pieces of evidence must be combined, strong perception scores alone may give a misleading picture of reliability. Such systems should be evaluated on the validity of their final decisions and paired with independent checks and human oversight when errors carry meaningful consequences. We release our code to make the evidence behind these findings auditable and to support further work on reliable multi-evidence reasoning.

\section*{Reproducibility Statement}

The anonymous repo contains the dataset generators, prompts, inference pipelines, exact-match evaluators, behavioral controls, hidden-state probes, activation-replacement procedures, Extract--Solve analysis, statistical summaries, and figure-generation scripts used in this study. The release also provides model identifiers, loader configurations, random seeds, sample counts, software requirements, portable Slurm templates, and unit tests.

\bibliography{iclr2027_conference}
\bibliographystyle{iclr2027_conference}

\clearpage
\appendix

% Appendix figure and table numbering
\setcounter{figure}{0}
\setcounter{table}{0}
\renewcommand{\thefigure}{A\arabic{figure}}
\renewcommand{\thetable}{A\arabic{table}}

% Appendix table styling
\arrayrulecolor{meiRule}
\setlength{\LTcapwidth}{\textwidth}

\clearpage
\appendix

\section*{Appendix Contents}
\label{app:contents}

\begingroup
\setlength{\parindent}{0pt}
\setlength{\parskip}{3pt}

\textbf{Appendix A: Experimental Construction and Evaluation}
\dotfill \pageref*{app:construction}\par
\hspace*{1.5em}A.1 Selection Instances
\dotfill \pageref*{app:selection_instances}\par
\hspace*{1.5em}A.2 Diagnostic Tasks
\dotfill \pageref*{app:diagnostic_tasks}\par
\hspace*{1.5em}A.3 Probe and Intervention Protocols
\dotfill \pageref*{app:probe_intervention_protocols}\par
\hspace*{1.5em}A.4 Representative Prompts
\dotfill \pageref*{app:prompts}\par

\textbf{Appendix B: Extended Behavioral Results}
\dotfill \pageref*{app:behavior}\par
\hspace*{1.5em}B.1 Candidate Count and Model Scale
\dotfill \pageref*{app:behavior_scale}\par
\hspace*{1.5em}B.2 Local Recovery and Conditional Decisions
\dotfill \pageref*{app:local_recovery}\par
\hspace*{1.5em}B.3 Retrieval, Constraint Requirements, and Paths
\dotfill \pageref*{app:behavior_diagnostics}\par

\textbf{Appendix C: Extended Probing Results}
\dotfill \pageref*{app:probes}\par
\hspace*{1.5em}C.1 Information Across Layers
\dotfill \pageref*{app:layer_profiles}\par
\hspace*{1.5em}C.2 Additional Models and Label Controls
\dotfill \pageref*{app:probe_controls}\par
\hspace*{1.5em}C.3 Split Sensitivity and Candidate Membership
\dotfill \pageref*{app:probe_sensitivity}\par

\textbf{Appendix D: Extended Activation-Patching Results}
\dotfill \pageref*{app:patching}\par
\hspace*{1.5em}D.1 Recovery at Larger Candidate Counts
\dotfill \pageref*{app:recovery_counts}\par
\hspace*{1.5em}D.2 Token Positions and Changed-Item Selection
\dotfill \pageref*{app:token_scopes}\par
\hspace*{1.5em}D.3 Selection of Intervention Cohorts
\dotfill \pageref*{app:cohort_selection}\par

\textbf{Appendix E: Uncertainty in Behavioral Results}
\dotfill \pageref*{app:behavior_ci}\par

\textbf{Appendix F: Relation to the Main Conclusions}
\dotfill \pageref*{app:relation_main}\par

\textbf{Appendix G: Prompting Controls}
\dotfill \pageref*{app:prompting_controls}\par
\hspace*{1.5em}G.1 Explicit Reasoning Budget
\dotfill \pageref*{app:reasoning_control}\par

\textbf{Appendix H: A Constructive Direction: Extract--Solve}
\dotfill \pageref*{app:extract_solve}\par

\textbf{Appendix I: Discussion}
\dotfill \pageref*{sec:discussion}\par

\textbf{Appendix J: Extended Related Work}
\dotfill \pageref*{app:extended_related_work}\par
\endgroup

\clearpage

\section{Experimental Construction and Evaluation}
\label{app:construction}
This appendix supplements the behavioral and internal analyses with experimental specifications, disaggregated results, and uncertainty estimates. The tables and figures use the completed benchmark runs and final intervention cohorts. Candidate counts and conditions already shown in the main figures are referenced there; additional conditions and finer-grained breakdowns are presented below.

\paragraph{Models and notation}
Q3, Q7, and Q32 denote Qwen2.5-VL-3B, -7B, and -32B; G4, G12, and G27 denote Gemma-3-4B, -12B, and -27B; I8 denotes InternVL3-8B; and L7 denotes LLaVA-OneVision-7B. These abbreviations identify checkpoints throughout the supplementary tables. V and T denote visual-grid and text-oracle global inputs. All eight models have selection and isolated local evaluations. The diagnostic experiments use Q3, Q7, G4, and G12. Same-grid local controls additionally cover I8 and L7, while full-prompt interventions cover Q3, Q7, G4, G12, and I8. The largest checkpoints contribute behavioral scaling evidence.

\subsection{Selection Instances}
\label{app:selection_instances}
The selection benchmark contains 1,600 instances, with 400 each at two, four, six, and eight candidate items. Each item has cost, weight, value, color, type, and a compatibility list. The requested subset contains two items when there are two candidates and three items otherwise. A valid answer contains the requested number of distinct items, respects the cost and weight budgets, contains the required color, satisfies the type-count restriction, and satisfies mutual compatibility. The rendered value field is queried locally but is not an optimization objective in this task.

The generator enumerates candidate subsets and retains instances with a unique valid solution. Within each candidate count, the nominal conflict settings zero and one are crossed with distractor ratios zero and 0.25, yielding 100 instances per cell. Realized modifications are limited by the number of items outside the solution. In particular, the two-candidate condition requires selecting both candidates and serves as a simple endpoint of the benchmark.

\paragraph{Inputs and scoring}
Visual inputs render the evidence cards together in a grid. Text-oracle inputs provide their attributes symbolically with the selection question. Global scoring compares the normalized predicted item set with the enumerated valid solutions. Local questions request all six attributes of a specified item. Field accuracy measures individual attribute recovery; all-field accuracy requires all attributes to be correct. Instance-level local success requires every constituent item query to be correct. Conditional global validity is computed on that retained instance subset, with its denominator reported explicitly.

Isolated-card queries and same-grid queries measure recovery under different input conditions. The same-grid control preserves the local question and labels while replacing the isolated image with the full evidence grid. This provides the closer behavioral comparison with global selection. Answers to separate local queries establish recoverability under those queries; internal probes subsequently examine states produced by the global prompt itself.

\subsection{Diagnostic Tasks}
\label{app:diagnostic_tasks}
\paragraph{Retrieval and constraint requirements}
The retrieval diagnostic contains 1,600 matched instances. Descriptors identify the target cards uniquely, so the model can recover the requested items without solving the joint constraints. The constraint diagnostic contains 6,400 instances: full, numerical-only, categorical-only, and compatibility-only variants of each source instance. Each variant retains the required subset size. Scoring enumerates the solutions permitted by its active constraints. Relaxing requirements can increase the number of acceptable answers, as quantified below; both the computational requirements and the acceptance set therefore contribute to the observed differences.
\begingroup\normalsize\setlength{\tabcolsep}{4pt}\renewcommand{\arraystretch}{1.08}
\begin{longtable}{lrrrr}
\caption{Mean number of valid subsets after constraint ablation. Enumerated on the final 400 instances per candidate count. Required subset size remains fixed; this quantifies the change in the answer acceptance criterion.}\label{tab:app_solutions}\\
\toprule\rowcolor{meiSage} $k$ & Full & Numerical & Categorical & Compatibility \\
\midrule\endfirsthead
\toprule\rowcolor{meiSage} $k$ & Full & Numerical & Categorical & Compatibility \\
\midrule\endhead\bottomrule\endfoot
2 & 1.00 & 1.00 & 1.00 & 1.00 \\
4 & 1.00 & 1.96 & 3.38 & 1.19 \\
6 & 1.00 & 5.84 & 14.75 & 2.01 \\
8 & 1.00 & 14.87 & 36.31 & 3.36 \\
\end{longtable}\endgroup

\paragraph{Weighted paths}
The second task contains 1,600 directed weighted graph instances, with 400 each at four, eight, twelve, and sixteen edge cards. The question asks for the minimum-cost path from A to H among nodes A--H. Each target path has three edges and two intermediate nodes. Target-edge costs range from one to four; distractor-edge costs exceed the target path's total cost by five to fifteen. Graph size therefore increases the surrounding evidence while the target path length remains fixed. Local questions recover source, destination, and cost from individual edge cards; global scoring compares the predicted ordered node sequence with the generated target path.

Target edges receive identifiers Edge 01--03 before presentation order is shuffled. Consequently, edge identifiers and individual edge costs provide cues to target membership. The experiment establishes a behavioral separation within this controlled construction. Generalization to arbitrary graph search would require additional constructions that remove these cues. The 16,000 local edge queries use isolated cards, so the same-grid control for selection remains the evidence addressing full-grid local recovery.

\subsection{Probe and Intervention Protocols}
\label{app:probe_intervention_protocols}
\paragraph{Layer-wise decoding}
Probes use the final prompt-position hidden state before answer generation. Hidden-state index zero denotes the embedding output; subsequent indices denote decoder-block outputs. Standardization is fitted on training data, followed by a class-balanced logistic regression classifier. Correctness labels describe the model's global decision. Membership probes instead predict whether a specified candidate belongs to the valid solution, fitting a separate classifier for each candidate identifier. These targets measure different properties of the same global-prompt representation.

Nested evaluation uses five outer folds and four inner folds. The inner folds select the layer; outer folds evaluate the selected probe on held-out instances. Repeated evaluations use seeds 13, 23, 37, 51, and 79. Membership scores average candidate-specific AUCs. Constant-label groups are omitted, and class counts constrain the available folds. Single-layer curves separately describe the distribution of decodable information across depth; their highest points are layer-search summaries rather than nested estimates.

\paragraph{Clean-state replacement}
Intervention recipients are instances answered correctly under the clean input and incorrectly after modification of a non-solution card. The modified evidence preserves the enumerated valid solution. A corresponding donor supplies states from the same clean instance; a mismatched donor supplies states from another clean instance with the required input alignment. The evaluated mismatched donors have different target solutions. Each comparison holds the recipient and intervention location fixed.

Replacement acts on decoder-block outputs at prompt positions during generation. Image-token interventions replace image positions; text-token interventions replace the complementary prompt positions; full-prompt interventions replace both. These position sets can carry shared information after attention. Their comparison measures which positions support recovery when replaced. Generated-token positions are not directly replaced.

\paragraph{Uncertainty and depth}
The paired donor contrast is the mean difference in recipient validity between corresponding and mismatched replacement. Supplementary intervals use 2,000 bootstrap resamples of recipient pairs and pointwise 95\% bounds. Repeated-seed probe variation describes sensitivity to the evaluation split. Training-label shuffles are reported separately because they hold the previously selected layer fixed. The deepest tested block with a positive lower bound summarizes the tested recovery range. Its resolution depends on the sampled blocks, and its uncertainty depends on cohort size. Comparisons across models describe their respective selected cohorts.

\subsection{Representative Prompts}
\label{app:prompts}
The following examples are taken from the final task manifests. They show the requested outputs and the distinction between local recovery, retrieval, selection, and path queries.
\paragraph{Selection query}
\begin{Verbatim}[breaklines=true,fontsize=\small,frame=single]
You are given 4 item cards. Choose exactly 3 items satisfying all constraints:

Available item IDs: Item A, Item B, Item C, Item D

- total cost <= 26
- total weight <= 16
- at least one selected item is yellow
- no more than 1 selected item may be type alpha
- all selected items must be mutually compatible

Return only a JSON object with one key named items. The value must be a list of exactly 3 selected item IDs from the available item IDs.
\end{Verbatim}
\paragraph{Single-item query}
\begin{Verbatim}[breaklines=true,fontsize=\small,frame=single]
Look at the image of Item A. What are its cost, weight, value, color, type, and compatible item list? Return JSON only.
\end{Verbatim}
\paragraph{Symbolic evidence}
\begin{Verbatim}[breaklines=true,fontsize=\small,frame=single]
Item A: cost=12, weight=4, value=6, color=green, type=gamma, compatible_with=['Item B', 'Item D']
Item B: cost=13, weight=8, value=11, color=yellow, type=gamma, compatible_with=['Item A', 'Item C', 'Item D']
Item C: cost=3, weight=1, value=8, color=green, type=gamma, compatible_with=['Item B', 'Item D']
Item D: cost=8, weight=6, value=24, color=yellow, type=alpha, compatible_with=['Item A', 'Item B', 'Item C']
\end{Verbatim}
\paragraph{Matched retrieval query}
\begin{Verbatim}[breaklines=true,fontsize=\small,frame=single]
You are given 4 item cards. Locate the one card matching each descriptor below. Each descriptor identifies exactly one card; do not apply the joint selection constraints.

- Descriptor 1: cost=13
- Descriptor 2: cost=3
- Descriptor 3: cost=8

Return only a JSON object with one key named items. The value must contain exactly 3 item IDs, one for each descriptor.
\end{Verbatim}
\paragraph{Compatibility-only query}
\begin{Verbatim}[breaklines=true,fontsize=\small,frame=single]
You are given 4 item cards. Choose exactly 3 items satisfying the compatibility constraint set.

Available item IDs: Item A, Item B, Item C, Item D

- all selected items must be mutually compatible

Return only a JSON object with one key named items. The value must be a list of exactly 3 selected item IDs. 
\end{Verbatim}
\paragraph{Path query}
\begin{Verbatim}[breaklines=true,fontsize=\small,frame=single]
The cards describe directed weighted edges. Find the minimum-cost path from A to H. Combine edge evidence as needed. Return only a JSON object with key path, whose value is the ordered list of node labels from start to goal.
\end{Verbatim}
\paragraph{Edge query}
\begin{Verbatim}[breaklines=true,fontsize=\small,frame=single]
Read Edge 03. Return JSON with source, target, and cost only.
\end{Verbatim}

\section{Extended Behavioral Results}
\label{app:behavior}
\subsection{Candidate Count and Model Scale}
\label{app:behavior_scale}
The main cross-model panel emphasizes the largest candidate count. Table~\ref{tab:app_global} supplies the remaining counts, while Table~\ref{tab:app_strata} separates the nominal generation settings. Together these breakdowns expose variation hidden by aggregate accuracy. The subgroups reuse the final predictions and retain the same evaluation rule.
\begingroup\normalsize\setlength{\tabcolsep}{4pt}\renewcommand{\arraystretch}{1.08}
\begin{longtable}{lrrrr}
\caption{\textbf{Global selection accuracy across candidate counts.} Entries report the percentage of answers satisfying all selection constraints on 400 instances per candidate count. V and T denote visual-grid and text-oracle inputs, respectively. The $k=8$ results appear in the main cross-model figure.}\label{tab:app_global}\\
\toprule\rowcolor{meiSage} Model & Input & $k=2$ & $k=4$ & $k=6$ \\
\midrule\endfirsthead
\toprule\rowcolor{meiSage} Model & Input & $k=2$ & $k=4$ & $k=6$ \\
\midrule\endhead\bottomrule\endfoot
Q3 & V & 99.75 & 53.00 & 18.00 \\
Q3 & T & 100.00 & 67.75 & 29.50 \\
Q7 & V & 100.00 & 76.00 & 29.50 \\
Q7 & T & 100.00 & 83.00 & 36.25 \\
Q32 & V & 100.00 & 90.50 & 39.25 \\
Q32 & T & 100.00 & 86.50 & 49.25 \\
G4 & V & 100.00 & 52.50 & 15.75 \\
G4 & T & 100.00 & 73.50 & 32.75 \\
G12 & V & 100.00 & 71.75 & 23.50 \\
G12 & T & 100.00 & 60.75 & 17.75 \\
G27 & V & 100.00 & 70.50 & 33.75 \\
G27 & T & 100.00 & 85.25 & 48.00 \\
I8 & V & 100.00 & 76.00 & 29.50 \\
I8 & T & 100.00 & 84.75 & 49.00 \\
L7 & V & 99.00 & 43.00 & 11.00 \\
L7 & T & 100.00 & 80.00 & 34.00 \\
\end{longtable}\endgroup

\begingroup\normalsize\setlength{\tabcolsep}{4pt}\renewcommand{\arraystretch}{1.08}
\begin{longtable}{lrrrrrr}
\caption{Final selection predictions stratified by nominal conflict $c$ and distractor ratio $d$. Each cell is validity (\%) over 100 instances. These are disjoint subgroups of the main benchmark. Parameter labels describe generation settings; the realized number of distractors is capped by the number of non-solution items.}\label{tab:app_strata}\\
\toprule\rowcolor{meiSage} Model & Input & $k$ & $c=0,d=0$ & $0,.25$ & $1,0$ & $1,.25$ \\
\midrule\endfirsthead
\toprule\rowcolor{meiSage} Model & Input & $k$ & $c=0,d=0$ & $0,.25$ & $1,0$ & $1,.25$ \\
\midrule\endhead\bottomrule\endfoot
Q3 & V & 4 & 41.00 & 71.00 & 39.00 & 61.00 \\
Q3 & V & 6 & 17.00 & 30.00 & 10.00 & 15.00 \\
Q3 & V & 8 & 10.00 & 4.00 & 3.00 & 6.00 \\
Q3 & T & 4 & 63.00 & 74.00 & 64.00 & 70.00 \\
Q3 & T & 6 & 30.00 & 36.00 & 22.00 & 30.00 \\
Q3 & T & 8 & 14.00 & 15.00 & 11.00 & 22.00 \\
Q7 & V & 4 & 69.00 & 82.00 & 70.00 & 83.00 \\
Q7 & V & 6 & 26.00 & 49.00 & 17.00 & 26.00 \\
Q7 & V & 8 & 16.00 & 13.00 & 10.00 & 10.00 \\
Q7 & T & 4 & 74.00 & 89.00 & 87.00 & 82.00 \\
Q7 & T & 6 & 31.00 & 51.00 & 28.00 & 35.00 \\
Q7 & T & 8 & 18.00 & 24.00 & 15.00 & 20.00 \\
Q32 & V & 4 & 81.00 & 99.00 & 85.00 & 97.00 \\
Q32 & V & 6 & 32.00 & 62.00 & 26.00 & 37.00 \\
Q32 & V & 8 & 20.00 & 25.00 & 11.00 & 18.00 \\
Q32 & T & 4 & 77.00 & 94.00 & 81.00 & 94.00 \\
Q32 & T & 6 & 46.00 & 58.00 & 34.00 & 59.00 \\
Q32 & T & 8 & 27.00 & 29.00 & 20.00 & 37.00 \\
G4 & V & 4 & 62.00 & 63.00 & 33.00 & 52.00 \\
G4 & V & 6 & 16.00 & 19.00 & 15.00 & 13.00 \\
G4 & V & 8 & 11.00 & 14.00 & 9.00 & 7.00 \\
G4 & T & 4 & 72.00 & 82.00 & 64.00 & 76.00 \\
G4 & T & 6 & 30.00 & 42.00 & 26.00 & 33.00 \\
G4 & T & 8 & 25.00 & 19.00 & 5.00 & 17.00 \\
G12 & V & 4 & 67.00 & 84.00 & 47.00 & 89.00 \\
G12 & V & 6 & 23.00 & 33.00 & 20.00 & 18.00 \\
G12 & V & 8 & 19.00 & 17.00 & 8.00 & 6.00 \\
G12 & T & 4 & 55.00 & 90.00 & 15.00 & 83.00 \\
G12 & T & 6 & 14.00 & 37.00 & 11.00 & 9.00 \\
G12 & T & 8 & 15.00 & 14.00 & 4.00 & 6.00 \\
G27 & V & 4 & 70.00 & 81.00 & 53.00 & 78.00 \\
G27 & V & 6 & 32.00 & 42.00 & 32.00 & 29.00 \\
G27 & V & 8 & 20.00 & 22.00 & 18.00 & 19.00 \\
G27 & T & 4 & 79.00 & 87.00 & 90.00 & 85.00 \\
G27 & T & 6 & 43.00 & 51.00 & 46.00 & 52.00 \\
G27 & T & 8 & 31.00 & 32.00 & 29.00 & 42.00 \\
I8 & V & 4 & 78.00 & 90.00 & 50.00 & 86.00 \\
I8 & V & 6 & 22.00 & 53.00 & 17.00 & 26.00 \\
I8 & V & 8 & 17.00 & 20.00 & 9.00 & 10.00 \\
I8 & T & 4 & 78.00 & 86.00 & 92.00 & 83.00 \\
I8 & T & 6 & 44.00 & 52.00 & 50.00 & 50.00 \\
I8 & T & 8 & 32.00 & 40.00 & 21.00 & 37.00 \\
L7 & V & 4 & 51.00 & 47.00 & 30.00 & 44.00 \\
L7 & V & 6 & 12.00 & 21.00 & 6.00 & 5.00 \\
L7 & V & 8 & 6.00 & 6.00 & 4.00 & 3.00 \\
L7 & T & 4 & 75.00 & 83.00 & 81.00 & 81.00 \\
L7 & T & 6 & 28.00 & 38.00 & 33.00 & 37.00 \\
L7 & T & 8 & 17.00 & 17.00 & 13.00 & 14.00 \\
\end{longtable}\endgroup

\subsection{Local Recovery and Conditional Decisions}
\label{app:local_recovery}
Field-level scores identify which attributes contribute to local errors. Conditional results then restrict global evaluation to instances with complete local recovery. Denominators matter: models with different local accuracy can be evaluated on different retained subsets. Same-grid field breakdowns complement the main conditional-grid figure and make its local-recovery condition explicit.
\begingroup\normalsize\setlength{\tabcolsep}{4pt}\renewcommand{\arraystretch}{1.08}
\begin{longtable}{lrrrrrrr}
\caption{Single-card field recovery (\%) over 8,000 item queries per model. All denotes that all six fields of an item are correct. This is an item-level metric; conditional selection uses all items in an instance.}\label{tab:app_local}\\
\toprule\rowcolor{meiSage} Model & Cost & Weight & Value & Color & Type & Compat. & All \\
\midrule\endfirsthead
\toprule\rowcolor{meiSage} Model & Cost & Weight & Value & Color & Type & Compat. & All \\
\midrule\endhead\bottomrule\endfoot
Q3 & 100.00 & 100.00 & 100.00 & 100.00 & 100.00 & 100.00 & 100.00 \\
Q7 & 100.00 & 100.00 & 100.00 & 100.00 & 100.00 & 100.00 & 100.00 \\
Q32 & 100.00 & 100.00 & 100.00 & 100.00 & 100.00 & 100.00 & 100.00 \\
G4 & 100.00 & 100.00 & 100.00 & 100.00 & 100.00 & 100.00 & 100.00 \\
G12 & 100.00 & 100.00 & 100.00 & 100.00 & 100.00 & 100.00 & 100.00 \\
G27 & 100.00 & 100.00 & 100.00 & 100.00 & 100.00 & 100.00 & 100.00 \\
I8 & 99.99 & 99.99 & 99.99 & 99.99 & 99.99 & 99.99 & 99.99 \\
L7 & 100.00 & 100.00 & 100.00 & 100.00 & 100.00 & 100.00 & 100.00 \\
\end{longtable}\endgroup
\begingroup\normalsize\setlength{\tabcolsep}{4pt}\renewcommand{\arraystretch}{1.08}
\begin{longtable}{lrrrr}
\caption{Selection validity conditioned on all single-card local questions being correct. $n$ is the retained number of instances out of 400. V/T are separate global runs conditioned on the same local-query subset. Unlike the main same-grid control, these local questions show one card at a time.}\label{tab:app_singlecond}\\
\toprule\rowcolor{meiSage} Model & $k$ & $n$ & V (\%) & T (\%) \\
\midrule\endfirsthead
\toprule\rowcolor{meiSage} Model & $k$ & $n$ & V (\%) & T (\%) \\
\midrule\endhead\bottomrule\endfoot
Q3 & 2 & 400 & 99.75 & 100.00 \\
Q3 & 4 & 400 & 53.00 & 67.75 \\
Q3 & 6 & 400 & 18.00 & 29.50 \\
Q3 & 8 & 400 & 5.75 & 15.50 \\
Q7 & 2 & 400 & 100.00 & 100.00 \\
Q7 & 4 & 400 & 76.00 & 83.00 \\
Q7 & 6 & 400 & 29.50 & 36.25 \\
Q7 & 8 & 400 & 12.25 & 19.25 \\
Q32 & 2 & 400 & 100.00 & 100.00 \\
Q32 & 4 & 400 & 90.50 & 86.50 \\
Q32 & 6 & 400 & 39.25 & 49.25 \\
Q32 & 8 & 400 & 18.50 & 28.25 \\
G4 & 2 & 400 & 100.00 & 100.00 \\
G4 & 4 & 400 & 52.50 & 73.50 \\
G4 & 6 & 400 & 15.75 & 32.75 \\
G4 & 8 & 400 & 10.25 & 16.50 \\
G12 & 2 & 400 & 100.00 & 100.00 \\
G12 & 4 & 400 & 71.75 & 60.75 \\
G12 & 6 & 400 & 23.50 & 17.75 \\
G12 & 8 & 400 & 12.50 & 9.75 \\
G27 & 2 & 400 & 100.00 & 100.00 \\
G27 & 4 & 400 & 70.50 & 85.25 \\
G27 & 6 & 400 & 33.75 & 48.00 \\
G27 & 8 & 400 & 19.75 & 33.50 \\
I8 & 2 & 400 & 100.00 & 100.00 \\
I8 & 4 & 400 & 76.00 & 84.75 \\
I8 & 6 & 400 & 29.50 & 49.00 \\
I8 & 8 & 399 & 14.04 & 32.58 \\
L7 & 2 & 400 & 99.00 & 100.00 \\
L7 & 4 & 400 & 43.00 & 80.00 \\
L7 & 6 & 400 & 11.00 & 34.00 \\
L7 & 8 & 400 & 4.75 & 15.25 \\
\end{longtable}\endgroup

\begingroup\normalsize\setlength{\tabcolsep}{4pt}\renewcommand{\arraystretch}{1.08}
\begin{longtable}{lrrrrrrrr}
\caption{Same-grid local field recovery (\%). All denotes all fields of one queried item. Each row contains $400k$ local queries. The main-text heatmap already reports instance-level retention and conditional global validity.}\label{tab:app_grid}\\
\toprule\rowcolor{meiSage} Model & $k$ & Cost & Weight & Value & Color & Type & Compat. & All \\
\midrule\endfirsthead
\toprule\rowcolor{meiSage} Model & $k$ & Cost & Weight & Value & Color & Type & Compat. & All \\
\midrule\endhead\bottomrule\endfoot
Q3 & 2 & 100.00 & 100.00 & 100.00 & 100.00 & 100.00 & 100.00 & 100.00 \\
Q3 & 4 & 100.00 & 100.00 & 100.00 & 100.00 & 100.00 & 100.00 & 100.00 \\
Q3 & 6 & 100.00 & 99.96 & 99.96 & 99.96 & 100.00 & 98.96 & 98.96 \\
Q3 & 8 & 100.00 & 99.91 & 100.00 & 100.00 & 100.00 & 98.88 & 98.78 \\
G4 & 2 & 100.00 & 100.00 & 100.00 & 100.00 & 100.00 & 100.00 & 100.00 \\
G4 & 4 & 98.81 & 99.69 & 99.81 & 100.00 & 100.00 & 98.44 & 97.19 \\
G4 & 6 & 99.96 & 99.83 & 98.42 & 100.00 & 100.00 & 94.33 & 92.83 \\
G4 & 8 & 99.94 & 95.75 & 99.12 & 99.91 & 98.44 & 93.12 & 87.16 \\
I8 & 2 & 100.00 & 100.00 & 100.00 & 100.00 & 100.00 & 100.00 & 100.00 \\
I8 & 4 & 100.00 & 100.00 & 100.00 & 100.00 & 100.00 & 99.12 & 99.12 \\
I8 & 6 & 100.00 & 100.00 & 100.00 & 100.00 & 100.00 & 99.29 & 99.29 \\
I8 & 8 & 100.00 & 99.94 & 100.00 & 100.00 & 99.97 & 99.75 & 99.66 \\
L7 & 2 & 100.00 & 100.00 & 100.00 & 100.00 & 100.00 & 100.00 & 100.00 \\
L7 & 4 & 100.00 & 99.94 & 100.00 & 100.00 & 100.00 & 95.75 & 95.69 \\
L7 & 6 & 100.00 & 99.96 & 99.62 & 99.88 & 99.54 & 73.92 & 73.33 \\
L7 & 8 & 100.00 & 99.47 & 99.34 & 99.19 & 98.81 & 81.16 & 79.59 \\
\end{longtable}\endgroup

\subsection{Retrieval, Constraint Requirements, and Paths}
\label{app:behavior_diagnostics}
The main diagnostic panels focus on eight candidates. The following results supply smaller candidate counts for both modalities. Each relaxed-constraint score uses its own enumerated acceptance set, allowing readers to interpret the performance changes alongside the solution-count table. Path supplements provide the text condition at smaller graph sizes and separate response parsing from exact-path correctness. High parsing rates locate the remaining failures in the predicted answer rather than an absent response.
\begingroup\normalsize\setlength{\tabcolsep}{4pt}\renewcommand{\arraystretch}{1.08}
\begin{longtable}{lrrrr}
\caption{Exact target-set retrieval (\%) for the remaining candidate counts. Each cell uses 400 instances; the main text reports $k=8$.}\label{tab:app_retrieval}\\
\toprule\rowcolor{meiSage} Model & Input & $k=2$ & $k=4$ & $k=6$ \\
\midrule\endfirsthead
\toprule\rowcolor{meiSage} Model & Input & $k=2$ & $k=4$ & $k=6$ \\
\midrule\endhead\bottomrule\endfoot
Q3 & V & 100.00 & 87.25 & 89.75 \\
Q3 & T & 100.00 & 83.00 & 83.50 \\
Q7 & V & 100.00 & 100.00 & 99.75 \\
Q7 & T & 100.00 & 100.00 & 98.25 \\
G4 & V & 100.00 & 87.75 & 82.75 \\
G4 & T & 100.00 & 98.50 & 90.75 \\
G12 & V & 100.00 & 99.75 & 98.50 \\
G12 & T & 100.00 & 100.00 & 99.50 \\
\end{longtable}\endgroup
\begingroup\normalsize\setlength{\tabcolsep}{4pt}\renewcommand{\arraystretch}{1.08}
\begin{longtable}{lrrrrrr}
\caption{Constraint-ablation validity (\%) at the remaining candidate counts, 400 instances per cell. Full is the within-suite prompt control; Num., Cat., and Rel. retain numerical, categorical, and compatibility requirements, respectively. Required subset size is retained throughout.}\label{tab:app_ablation}\\
\toprule\rowcolor{meiSage} Model & Input & $k$ & Full & Num. & Cat. & Rel. \\
\midrule\endfirsthead
\toprule\rowcolor{meiSage} Model & Input & $k$ & Full & Num. & Cat. & Rel. \\
\midrule\endhead\bottomrule\endfoot
Q3 & V & 2 & 99.50 & 100.00 & 99.00 & 100.00 \\
Q3 & V & 4 & 52.00 & 72.00 & 97.25 & 61.75 \\
Q3 & V & 6 & 18.25 & 46.25 & 96.75 & 22.25 \\
Q3 & T & 2 & 100.00 & 100.00 & 100.00 & 100.00 \\
Q3 & T & 4 & 69.50 & 79.75 & 96.50 & 81.50 \\
Q3 & T & 6 & 31.50 & 46.50 & 95.75 & 50.50 \\
Q7 & V & 2 & 100.00 & 100.00 & 100.00 & 100.00 \\
Q7 & V & 4 & 74.50 & 86.25 & 99.00 & 88.50 \\
Q7 & V & 6 & 30.25 & 64.50 & 97.50 & 49.50 \\
Q7 & T & 2 & 100.00 & 100.00 & 100.00 & 100.00 \\
Q7 & T & 4 & 82.50 & 82.25 & 99.00 & 87.75 \\
Q7 & T & 6 & 40.00 & 50.00 & 94.75 & 52.75 \\
G4 & V & 2 & 100.00 & 100.00 & 100.00 & 100.00 \\
G4 & V & 4 & 56.50 & 64.75 & 98.50 & 82.75 \\
G4 & V & 6 & 16.00 & 45.75 & 94.00 & 33.00 \\
G4 & T & 2 & 100.00 & 100.00 & 100.00 & 100.00 \\
G4 & T & 4 & 72.00 & 77.75 & 98.25 & 80.25 \\
G4 & T & 6 & 33.50 & 51.75 & 94.75 & 49.50 \\
G12 & V & 2 & 100.00 & 100.00 & 100.00 & 100.00 \\
G12 & V & 4 & 71.75 & 80.00 & 98.75 & 86.75 \\
G12 & V & 6 & 26.25 & 52.25 & 97.25 & 58.75 \\
G12 & T & 2 & 100.00 & 100.00 & 100.00 & 100.00 \\
G12 & T & 4 & 60.50 & 77.25 & 96.75 & 80.50 \\
G12 & T & 6 & 19.25 & 62.50 & 95.75 & 64.75 \\
\end{longtable}\endgroup

\begingroup\normalsize\setlength{\tabcolsep}{4pt}\renewcommand{\arraystretch}{1.08}
\begin{longtable}{lrrr}
\caption{Text-oracle exact-path accuracy (\%) at the remaining graph sizes, 400 instances per cell. The main text provides the full visual curve and the sixteen-edge text result.}\label{tab:app_pathtext}\\
\toprule\rowcolor{meiSage} Model & 4 edges & 8 edges & 12 edges \\
\midrule\endfirsthead
\toprule\rowcolor{meiSage} Model & 4 edges & 8 edges & 12 edges \\
\midrule\endhead\bottomrule\endfoot
Q3 & 96.25 & 56.00 & 41.50 \\
Q7 & 99.00 & 83.00 & 60.75 \\
G4 & 96.00 & 77.00 & 56.75 \\
G12 & 99.25 & 88.25 & 80.75 \\
\end{longtable}\endgroup
\begingroup\normalsize\setlength{\tabcolsep}{4pt}\renewcommand{\arraystretch}{1.08}
\begin{longtable}{lrrrrr}
\caption{Nonempty parsed-path rates (\%).}\label{tab:app_pathformat}\\
\toprule\rowcolor{meiSage} Model & Input & 4 edges & 8 edges & 12 edges & 16 edges \\
\midrule\endfirsthead
\toprule\rowcolor{meiSage} Model & Input & 4 edges & 8 edges & 12 edges & 16 edges \\
\midrule\endhead\bottomrule\endfoot
Q3 & V & 100.00 & 100.00 & 100.00 & 100.00 \\
Q3 & T & 100.00 & 100.00 & 100.00 & 100.00 \\
Q7 & V & 100.00 & 100.00 & 100.00 & 100.00 \\
Q7 & T & 100.00 & 100.00 & 100.00 & 100.00 \\
G4 & V & 100.00 & 100.00 & 100.00 & 100.00 \\
G4 & T & 100.00 & 100.00 & 100.00 & 100.00 \\
G12 & V & 100.00 & 100.00 & 100.00 & 100.00 \\
G12 & T & 100.00 & 100.00 & 100.00 & 100.00 \\
\end{longtable}\endgroup
\begingroup\normalsize\setlength{\tabcolsep}{4pt}\renewcommand{\arraystretch}{1.08}
\begin{longtable}{lrrr}
\caption{Local edge-field accuracy (\%) over 16,000 isolated queries. The main-text all-field score is not repeated.}\label{tab:app_pathlocal}\\
\toprule\rowcolor{meiSage} Model & Source & Target & Cost \\
\midrule\endfirsthead
\toprule\rowcolor{meiSage} Model & Source & Target & Cost \\
\midrule\endhead\bottomrule\endfoot
Q3 & 100.00 & 100.00 & 100.00 \\
Q7 & 100.00 & 100.00 & 100.00 \\
G4 & 100.00 & 100.00 & 100.00 \\
G12 & 100.00 & 100.00 & 100.00 \\
\end{longtable}\endgroup

\section{Extended Probing Results}
\label{app:probes}
\subsection{Information Across Layers}
\label{app:layer_profiles}
The following curves distinguish pooled correctness prediction from prediction within a fixed candidate count. Pooled labels covary with task difficulty, so the within-count curves provide the more focused view of decision-related information. Both are included to make that contrast inspectable. The membership analyses below address solution content directly.
\begin{figure}[!htbp]\centering
\includegraphics[width=\linewidth]{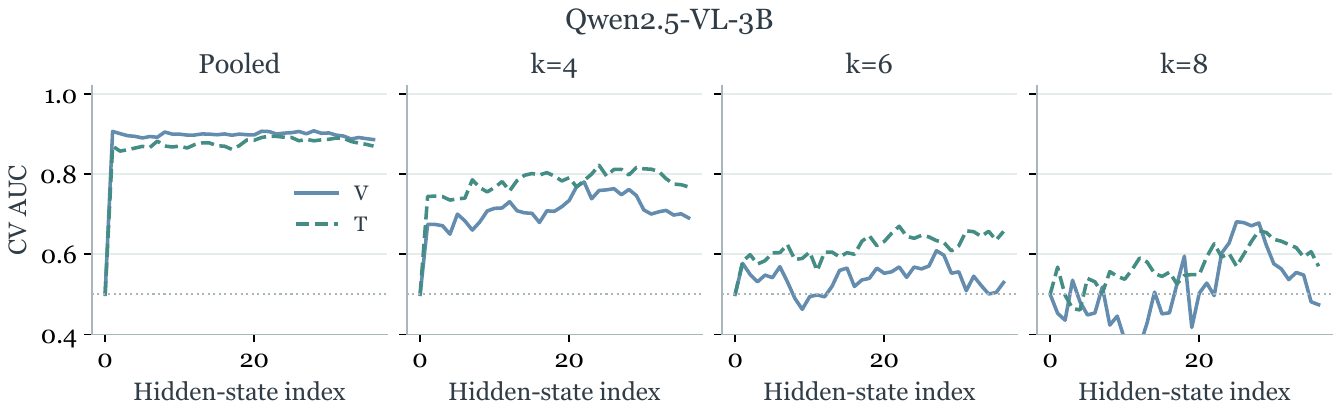}
\caption{Qwen2.5-VL-3B: fixed-layer correctness AUC, pooled and grouped by candidate count. V/T denote visual/text input. Each layer is independently evaluated by cross-validation; choosing a peak from these curves is exploratory, unlike nested layer selection. Hidden-state index 0 denotes the embedding output. }
\label{fig:app_layer_Q3}
\end{figure}
\begin{figure}[!htbp]\centering
\includegraphics[width=\linewidth]{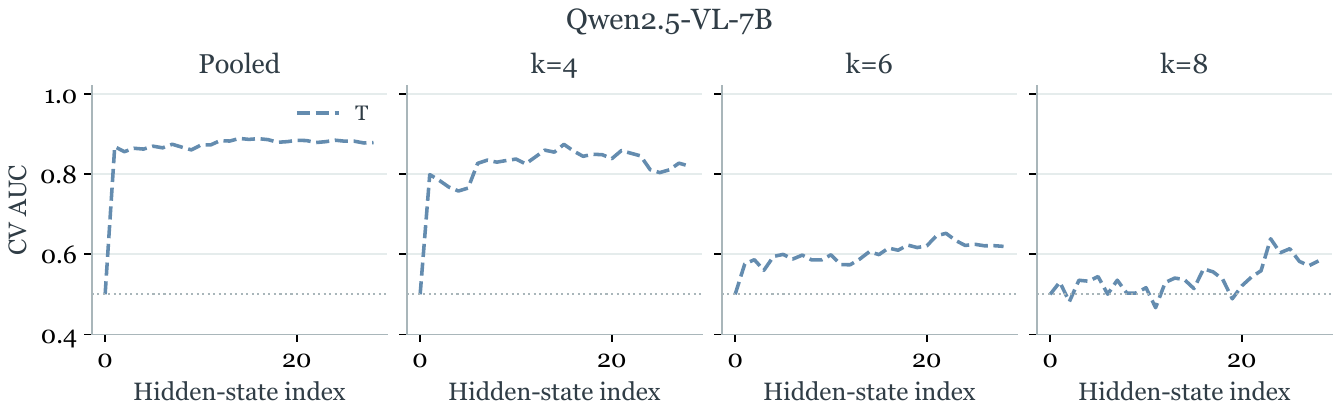}
\caption{Qwen2.5-VL-7B: fixed-layer correctness AUC, pooled and grouped by candidate count. V/T denote visual/text input. Each layer is independently evaluated by cross-validation; choosing a peak from these curves is exploratory, unlike nested layer selection. Hidden-state index 0 denotes the embedding output. Only the reconciled text condition is included.}
\label{fig:app_layer_Q7}
\end{figure}
\begin{figure}[!htbp]\centering
\includegraphics[width=\linewidth]{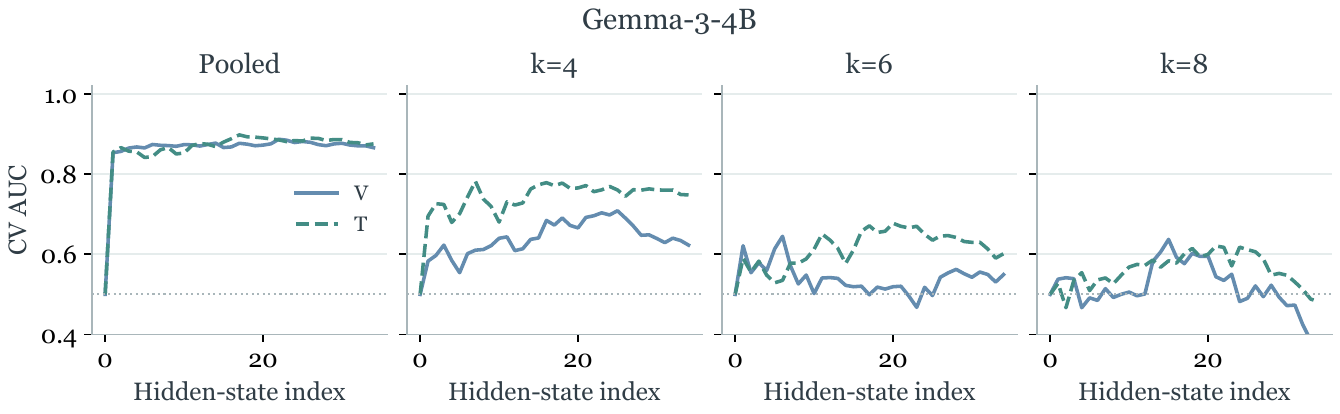}
\caption{Gemma-3-4B: fixed-layer correctness AUC, pooled and grouped by candidate count. V/T denote visual/text input. Each layer is independently evaluated by cross-validation; choosing a peak from these curves is exploratory, unlike nested layer selection. Hidden-state index 0 denotes the embedding output. }
\label{fig:app_layer_G4}
\end{figure}
\begin{figure}[!htbp]\centering
\includegraphics[width=\linewidth]{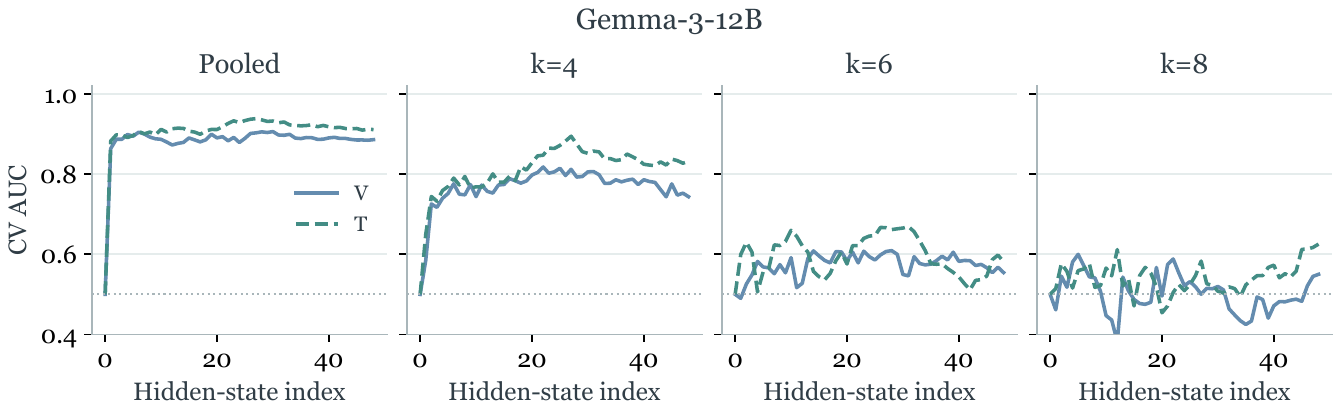}
\caption{Gemma-3-12B: fixed-layer correctness AUC, pooled and grouped by candidate count. V/T denote visual/text input. Each layer is independently evaluated by cross-validation; choosing a peak from these curves is exploratory, unlike nested layer selection. Hidden-state index 0 denotes the embedding output. }
\label{fig:app_layer_G12}
\end{figure}
\begin{figure}[!htbp]\centering
\includegraphics[width=\linewidth]{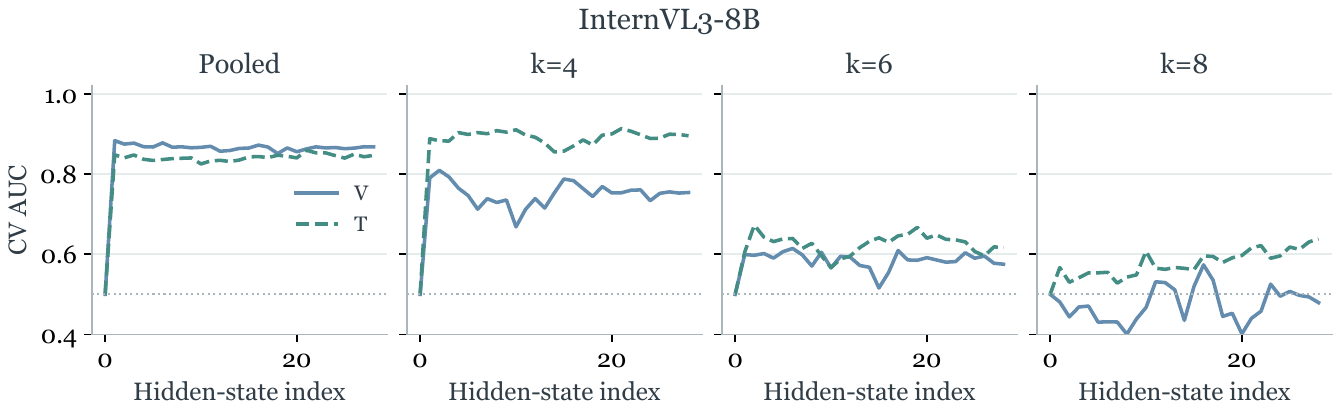}
\caption{InternVL3-8B: fixed-layer correctness AUC, pooled and grouped by candidate count. V/T denote visual/text input. Each layer is independently evaluated by cross-validation; choosing a peak from these curves is exploratory, unlike nested layer selection. Hidden-state index 0 denotes the embedding output. }
\label{fig:app_layer_I8}
\end{figure}
\begin{figure}[!htbp]\centering
\includegraphics[width=\linewidth]{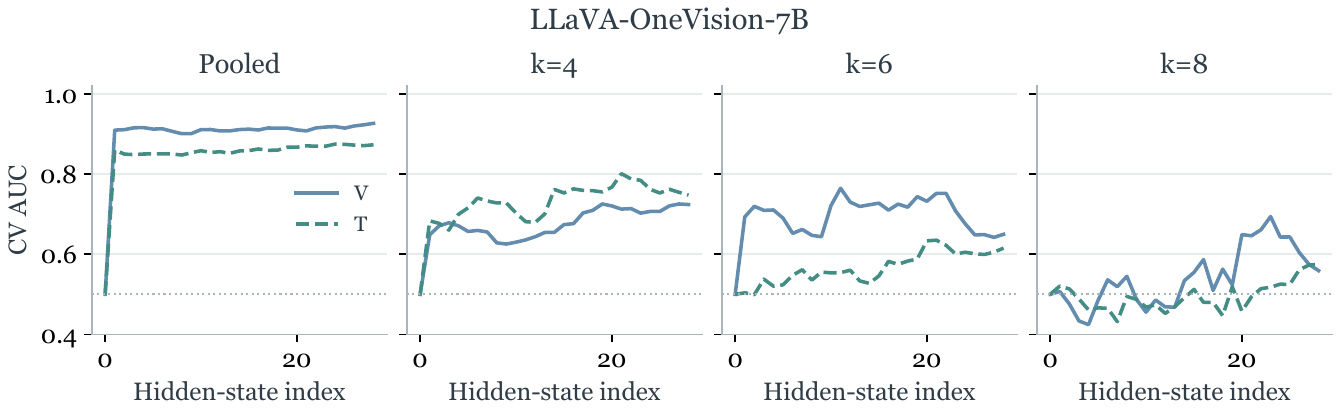}
\caption{LLaVA-OneVision-7B: fixed-layer correctness AUC, pooled and grouped by candidate count. V/T denote visual/text input. Each layer is independently evaluated by cross-validation; choosing a peak from these curves is exploratory, unlike nested layer selection. Hidden-state index 0 denotes the embedding output. }
\label{fig:app_layer_L7}
\end{figure}

\FloatBarrier

\subsection{Additional Models and Label Controls}
\label{app:probe_controls}
The additional-model nested results extend the main repeated-probe comparison to InternVL3 and LLaVA-OneVision. Label-shuffle scores characterize the reference behavior under disrupted training labels at the selected layer. Class-count tables explain groups omitted from the probe outputs. These controls are reported with their actual evaluation procedure rather than converted into permutation significance claims.
\begingroup\normalsize\setlength{\tabcolsep}{4pt}\renewcommand{\arraystretch}{1.08}
\begin{longtable}{lrrrrrr}
\caption{Additional-model nested correctness probes. AUC and accuracy are means across outer folds. Positive rate describes class balance; depth is the mean selected hidden-state index normalized by the final index. These models are absent from the main repeated-probe figure.}\label{tab:app_nested}\\
\toprule\rowcolor{meiSage} Model & Input & $k$ & AUC & Acc. & Positive & Depth \\
\midrule\endfirsthead
\toprule\rowcolor{meiSage} Model & Input & $k$ & AUC & Acc. & Positive & Depth \\
\midrule\endhead\bottomrule\endfoot
I8 & V & 4 & 0.790 & 0.720 & 0.760 & 0.157 \\
I8 & V & 6 & 0.561 & 0.610 & 0.295 & 0.443 \\
I8 & V & 8 & 0.470 & 0.735 & 0.140 & 0.557 \\
I8 & T & 4 & 0.873 & 0.850 & 0.848 & 0.457 \\
I8 & T & 6 & 0.621 & 0.588 & 0.490 & 0.386 \\
I8 & T & 8 & 0.616 & 0.620 & 0.325 & 0.836 \\
L7 & V & 4 & 0.703 & 0.645 & 0.430 & 0.629 \\
L7 & V & 6 & 0.713 & 0.785 & 0.110 & 0.479 \\
L7 & V & 8 & 0.596 & 0.903 & 0.048 & 0.664 \\
L7 & T & 4 & 0.792 & 0.775 & 0.800 & 0.700 \\
L7 & T & 6 & 0.567 & 0.557 & 0.340 & 0.564 \\
L7 & T & 8 & 0.565 & 0.728 & 0.152 & 0.800 \\
\end{longtable}\endgroup
\begingroup\normalsize\setlength{\tabcolsep}{4pt}\renewcommand{\arraystretch}{1.08}
\begin{longtable}{lrrrrr}
\caption{Training-label shuffle controls at the already selected layer. SD is across pooled shuffle evaluations, not a confidence interval or a nested permutation test. Each included run uses five outer folds and ten shuffles per fold. The main repeated-seed runs use zero shuffles.}\label{tab:app_shuffle}\\
\toprule\rowcolor{meiSage} Model & Input & $k$ & Shuffle AUC & SD & Shuffles/fold \\
\midrule\endfirsthead
\toprule\rowcolor{meiSage} Model & Input & $k$ & Shuffle AUC & SD & Shuffles/fold \\
\midrule\endhead\bottomrule\endfoot
Q3 & V & 4 & 0.505 & 0.059 & 10 \\
Q3 & V & 6 & 0.486 & 0.081 & 10 \\
Q3 & V & 8 & 0.515 & 0.134 & 10 \\
Q3 & T & 4 & 0.516 & 0.089 & 10 \\
Q3 & T & 6 & 0.522 & 0.083 & 10 \\
Q3 & T & 8 & 0.501 & 0.102 & 10 \\
Q7 & T & 4 & 0.511 & 0.095 & 10 \\
Q7 & T & 6 & 0.509 & 0.061 & 10 \\
Q7 & T & 8 & 0.508 & 0.100 & 10 \\
G4 & V & 4 & 0.495 & 0.067 & 10 \\
G4 & V & 6 & 0.519 & 0.085 & 10 \\
G4 & V & 8 & 0.492 & 0.101 & 10 \\
G4 & T & 4 & 0.494 & 0.073 & 10 \\
G4 & T & 6 & 0.494 & 0.074 & 10 \\
G4 & T & 8 & 0.507 & 0.103 & 10 \\
G12 & V & 4 & 0.527 & 0.077 & 10 \\
G12 & V & 6 & 0.506 & 0.085 & 10 \\
G12 & V & 8 & 0.512 & 0.105 & 10 \\
G12 & T & 4 & 0.526 & 0.075 & 10 \\
G12 & T & 6 & 0.496 & 0.079 & 10 \\
G12 & T & 8 & 0.505 & 0.133 & 10 \\
I8 & V & 4 & 0.504 & 0.074 & 10 \\
I8 & V & 6 & 0.507 & 0.070 & 10 \\
I8 & V & 8 & 0.505 & 0.086 & 10 \\
I8 & T & 4 & 0.529 & 0.096 & 10 \\
I8 & T & 6 & 0.509 & 0.081 & 10 \\
I8 & T & 8 & 0.510 & 0.062 & 10 \\
L7 & V & 4 & 0.515 & 0.075 & 10 \\
L7 & V & 6 & 0.479 & 0.110 & 10 \\
L7 & V & 8 & 0.505 & 0.148 & 10 \\
L7 & T & 4 & 0.494 & 0.086 & 10 \\
L7 & T & 6 & 0.485 & 0.067 & 10 \\
L7 & T & 8 & 0.501 & 0.098 & 10 \\
\end{longtable}\endgroup
\begingroup\normalsize\setlength{\tabcolsep}{4pt}\renewcommand{\arraystretch}{1.08}
\begin{longtable}{lrrrr}
\caption{Candidate-count groups absent from the fixed-layer five-fold probe outputs. Positive counts refer to the final behavioral predictions; all remaining instances are negatives.}\label{tab:app_skips}\\
\toprule\rowcolor{meiSage} Model & Input & $k$ & Positive & $n$ \\
\midrule\endfirsthead
\toprule\rowcolor{meiSage} Model & Input & $k$ & Positive & $n$ \\
\midrule\endhead\bottomrule\endfoot
Q3 & V & 2 & 399 & 400 \\
Q3 & T & 2 & 400 & 400 \\
Q7 & V & 2 & 400 & 400 \\
Q7 & T & 2 & 400 & 400 \\
G4 & V & 2 & 400 & 400 \\
G4 & T & 2 & 400 & 400 \\
G12 & V & 2 & 400 & 400 \\
G12 & T & 2 & 400 & 400 \\
I8 & V & 2 & 400 & 400 \\
I8 & T & 2 & 400 & 400 \\
L7 & T & 2 & 400 & 400 \\
\end{longtable}\endgroup

\begingroup\normalsize\setlength{\tabcolsep}{4pt}\renewcommand{\arraystretch}{1.08}
\begin{longtable}{lrrr}
\caption{LLaVA-OneVision visual correctness at $k=2$: all recorded fixed-layer scores. There are 396 positive and four negative instances. These small-minority-class results are reported separately from the five-outer-fold nested analysis. SD describes the recorded cross-validation folds.}\label{tab:app_rareclass}\\
\toprule\rowcolor{meiSage} Index & AUC & AUC SD & Accuracy \\
\midrule\endfirsthead
\toprule\rowcolor{meiSage} Index & AUC & AUC SD & Accuracy \\
\midrule\endhead\bottomrule\endfoot
0 & 0.500 & 0.000 & 0.010 \\
1 & 0.785 & 0.159 & 0.985 \\
2 & 0.854 & 0.025 & 0.985 \\
3 & 0.881 & 0.031 & 0.983 \\
4 & 0.937 & 0.023 & 0.982 \\
5 & 0.970 & 0.010 & 0.982 \\
6 & 0.967 & 0.015 & 0.980 \\
7 & 0.967 & 0.023 & 0.980 \\
8 & 0.962 & 0.027 & 0.990 \\
9 & 0.962 & 0.015 & 0.982 \\
10 & 0.939 & 0.042 & 0.985 \\
11 & 0.922 & 0.050 & 0.982 \\
12 & 0.924 & 0.027 & 0.982 \\
13 & 0.917 & 0.019 & 0.983 \\
14 & 0.949 & 0.033 & 0.985 \\
15 & 0.967 & 0.018 & 0.980 \\
16 & 0.980 & 0.012 & 0.982 \\
17 & 0.972 & 0.004 & 0.985 \\
18 & 0.985 & 0.005 & 0.988 \\
19 & 0.992 & 0.008 & 0.988 \\
20 & 0.997 & 0.004 & 0.988 \\
21 & 0.997 & 0.004 & 0.995 \\
22 & 0.997 & 0.004 & 0.998 \\
23 & 0.997 & 0.004 & 0.995 \\
24 & 0.997 & 0.004 & 0.992 \\
25 & 0.997 & 0.004 & 0.992 \\
26 & 0.997 & 0.004 & 0.992 \\
27 & 0.997 & 0.004 & 0.992 \\
28 & 0.997 & 0.004 & 0.995 \\
\end{longtable}\endgroup

\subsection{Split Sensitivity and Candidate Membership}
\label{app:probe_sensitivity}
The main probe figure reports seed-level means and standard deviations. The tables below expose individual seed scores, candidate-specific membership results, and outer-fold layer choices. They show whether an aggregate pattern is shared across candidates and whether the selected layer is stable across splits. Membership decoding answers whether solution membership can be recovered from the state; correctness decoding answers whether that state predicts the model's eventual success.
\begingroup\normalsize\setlength{\tabcolsep}{4pt}\renewcommand{\arraystretch}{1.08}
\begin{longtable}{lrrrrrrr}
\caption{Qwen2.5-VL-3B: per-seed held-out correctness AUC. The main text presents mean and SD; this table exposes sensitivity to the split without reprinting that summary.}\label{tab:app_seeds_Q3}\\
\toprule\rowcolor{meiSage} Model & Input & $k$ & 13 & 23 & 37 & 51 & 79 \\
\midrule\endfirsthead
\toprule\rowcolor{meiSage} Model & Input & $k$ & 13 & 23 & 37 & 51 & 79 \\
\midrule\endhead\bottomrule\endfoot
Q3 & V & 4 & 0.754 & 0.754 & 0.714 & 0.740 & 0.759 \\
Q3 & V & 6 & 0.569 & 0.542 & 0.508 & 0.541 & 0.581 \\
Q3 & V & 8 & 0.594 & 0.598 & 0.547 & 0.543 & 0.598 \\
Q3 & T & 4 & 0.792 & 0.801 & 0.787 & 0.789 & 0.792 \\
Q3 & T & 6 & 0.606 & 0.628 & 0.664 & 0.669 & 0.588 \\
Q3 & T & 8 & 0.615 & 0.544 & 0.588 & 0.597 & 0.559 \\
\end{longtable}\endgroup
\begingroup\normalsize\setlength{\tabcolsep}{4pt}\renewcommand{\arraystretch}{1.08}
\begin{longtable}{lrrrrrrr}
\caption{Qwen2.5-VL-7B: per-seed held-out correctness AUC. The main text presents mean and SD; this table exposes sensitivity to the split without reprinting that summary.}\label{tab:app_seeds_Q7}\\
\toprule\rowcolor{meiSage} Model & Input & $k$ & 13 & 23 & 37 & 51 & 79 \\
\midrule\endfirsthead
\toprule\rowcolor{meiSage} Model & Input & $k$ & 13 & 23 & 37 & 51 & 79 \\
\midrule\endhead\bottomrule\endfoot
Q7 & T & 4 & 0.821 & 0.815 & 0.851 & 0.829 & 0.810 \\
Q7 & T & 6 & 0.599 & 0.607 & 0.588 & 0.525 & 0.603 \\
Q7 & T & 8 & 0.565 & 0.557 & 0.635 & 0.576 & 0.548 \\
\end{longtable}\endgroup
\begingroup\normalsize\setlength{\tabcolsep}{4pt}\renewcommand{\arraystretch}{1.08}
\begin{longtable}{lrrrrrrr}
\caption{Gemma-3-4B: per-seed held-out correctness AUC. The main text presents mean and SD; this table exposes sensitivity to the split without reprinting that summary.}\label{tab:app_seeds_G4}\\
\toprule\rowcolor{meiSage} Model & Input & $k$ & 13 & 23 & 37 & 51 & 79 \\
\midrule\endfirsthead
\toprule\rowcolor{meiSage} Model & Input & $k$ & 13 & 23 & 37 & 51 & 79 \\
\midrule\endhead\bottomrule\endfoot
G4 & V & 4 & 0.645 & 0.642 & 0.682 & 0.640 & 0.670 \\
G4 & V & 6 & 0.585 & 0.575 & 0.644 & 0.613 & 0.585 \\
G4 & V & 8 & 0.458 & 0.502 & 0.588 & 0.571 & 0.484 \\
G4 & T & 4 & 0.765 & 0.753 & 0.778 & 0.775 & 0.782 \\
G4 & T & 6 & 0.657 & 0.646 & 0.653 & 0.668 & 0.630 \\
G4 & T & 8 & 0.583 & 0.593 & 0.612 & 0.623 & 0.592 \\
\end{longtable}\endgroup
\begingroup\normalsize\setlength{\tabcolsep}{4pt}\renewcommand{\arraystretch}{1.08}
\begin{longtable}{lrrrrrrr}
\caption{Gemma-3-12B: per-seed held-out correctness AUC. The main text presents mean and SD; this table exposes sensitivity to the split without reprinting that summary.}\label{tab:app_seeds_G12}\\
\toprule\rowcolor{meiSage} Model & Input & $k$ & 13 & 23 & 37 & 51 & 79 \\
\midrule\endfirsthead
\toprule\rowcolor{meiSage} Model & Input & $k$ & 13 & 23 & 37 & 51 & 79 \\
\midrule\endhead\bottomrule\endfoot
G12 & V & 4 & 0.798 & 0.770 & 0.778 & 0.766 & 0.744 \\
G12 & V & 6 & 0.524 & 0.531 & 0.587 & 0.564 & 0.545 \\
G12 & V & 8 & 0.451 & 0.549 & 0.525 & 0.509 & 0.573 \\
G12 & T & 4 & 0.893 & 0.883 & 0.870 & 0.879 & 0.851 \\
G12 & T & 6 & 0.644 & 0.620 & 0.558 & 0.610 & 0.611 \\
G12 & T & 8 & 0.537 & 0.587 & 0.502 & 0.561 & 0.556 \\
\end{longtable}\endgroup

\begingroup\normalsize\setlength{\tabcolsep}{4pt}\renewcommand{\arraystretch}{1.08}
\begin{longtable}{lrrrrrrrr}
\caption{Qwen2.5-VL-3B: candidate-specific membership AUC for each seed. A--H denote candidate IDs. Positive is the fraction of instances whose valid subset contains that candidate. The main figure averages these candidate-level results.}\label{tab:app_member_Q3}\\
\toprule\rowcolor{meiSage} Input & $k$ & ID & 13 & 23 & 37 & 51 & 79 & Positive \\
\midrule\endfirsthead
\toprule\rowcolor{meiSage} Input & $k$ & ID & 13 & 23 & 37 & 51 & 79 & Positive \\
\midrule\endhead\bottomrule\endfoot
V & 4 & A & 0.993 & 0.998 & 0.995 & 0.998 & 0.998 & 0.792 \\
V & 4 & B & 0.967 & 0.978 & 0.980 & 0.969 & 0.973 & 0.745 \\
V & 4 & C & 0.967 & 0.970 & 0.960 & 0.967 & 0.979 & 0.755 \\
V & 4 & D & 0.976 & 0.991 & 0.994 & 0.990 & 0.985 & 0.708 \\
V & 6 & A & 0.949 & 0.959 & 0.945 & 0.961 & 0.955 & 0.500 \\
V & 6 & B & 0.845 & 0.838 & 0.862 & 0.846 & 0.829 & 0.510 \\
V & 6 & C & 0.849 & 0.778 & 0.803 & 0.817 & 0.806 & 0.495 \\
V & 6 & D & 0.772 & 0.787 & 0.792 & 0.761 & 0.740 & 0.515 \\
V & 6 & E & 0.763 & 0.747 & 0.747 & 0.745 & 0.747 & 0.470 \\
V & 6 & F & 0.785 & 0.794 & 0.792 & 0.780 & 0.808 & 0.510 \\
V & 8 & A & 0.851 & 0.828 & 0.844 & 0.871 & 0.850 & 0.420 \\
V & 8 & B & 0.732 & 0.740 & 0.729 & 0.720 & 0.719 & 0.345 \\
V & 8 & C & 0.677 & 0.679 & 0.702 & 0.677 & 0.684 & 0.375 \\
V & 8 & D & 0.664 & 0.654 & 0.677 & 0.615 & 0.646 & 0.338 \\
V & 8 & E & 0.611 & 0.609 & 0.651 & 0.611 & 0.629 & 0.388 \\
V & 8 & F & 0.609 & 0.621 & 0.608 & 0.618 & 0.651 & 0.357 \\
V & 8 & G & 0.717 & 0.689 & 0.684 & 0.694 & 0.649 & 0.412 \\
V & 8 & H & 0.651 & 0.633 & 0.606 & 0.656 & 0.650 & 0.365 \\
T & 4 & A & 0.998 & 0.995 & 0.995 & 0.998 & 0.998 & 0.792 \\
T & 4 & B & 0.975 & 0.985 & 0.985 & 0.982 & 0.988 & 0.745 \\
T & 4 & C & 0.975 & 0.993 & 0.986 & 0.990 & 0.990 & 0.755 \\
T & 4 & D & 0.991 & 0.993 & 0.993 & 0.993 & 0.988 & 0.708 \\
T & 6 & A & 0.954 & 0.955 & 0.958 & 0.962 & 0.956 & 0.500 \\
T & 6 & B & 0.887 & 0.895 & 0.897 & 0.915 & 0.893 & 0.510 \\
T & 6 & C & 0.824 & 0.847 & 0.816 & 0.827 & 0.843 & 0.495 \\
T & 6 & D & 0.764 & 0.788 & 0.763 & 0.787 & 0.774 & 0.515 \\
T & 6 & E & 0.822 & 0.816 & 0.813 & 0.821 & 0.844 & 0.470 \\
T & 6 & F & 0.861 & 0.865 & 0.861 & 0.857 & 0.867 & 0.510 \\
T & 8 & A & 0.856 & 0.847 & 0.869 & 0.871 & 0.887 & 0.420 \\
T & 8 & B & 0.732 & 0.736 & 0.725 & 0.742 & 0.731 & 0.345 \\
T & 8 & C & 0.687 & 0.640 & 0.614 & 0.652 & 0.698 & 0.375 \\
T & 8 & D & 0.643 & 0.668 & 0.663 & 0.623 & 0.674 & 0.338 \\
T & 8 & E & 0.686 & 0.627 & 0.661 & 0.665 & 0.686 & 0.388 \\
T & 8 & F & 0.712 & 0.714 & 0.708 & 0.677 & 0.681 & 0.357 \\
T & 8 & G & 0.758 & 0.740 & 0.752 & 0.742 & 0.752 & 0.412 \\
T & 8 & H & 0.801 & 0.798 & 0.771 & 0.810 & 0.789 & 0.365 \\
\end{longtable}\endgroup
\begingroup\normalsize\setlength{\tabcolsep}{4pt}\renewcommand{\arraystretch}{1.08}
\begin{longtable}{lrrrrrrrr}
\caption{Qwen2.5-VL-7B: candidate-specific membership AUC for each seed. A--H denote candidate IDs. Positive is the fraction of instances whose valid subset contains that candidate. The main figure averages these candidate-level results.}\label{tab:app_member_Q7}\\
\toprule\rowcolor{meiSage} Input & $k$ & ID & 13 & 23 & 37 & 51 & 79 & Positive \\
\midrule\endfirsthead
\toprule\rowcolor{meiSage} Input & $k$ & ID & 13 & 23 & 37 & 51 & 79 & Positive \\
\midrule\endhead\bottomrule\endfoot
V & 4 & A & 0.989 & 0.990 & 0.988 & 0.984 & 0.987 & 0.792 \\
V & 4 & B & 0.982 & 0.984 & 0.983 & 0.983 & 0.989 & 0.745 \\
V & 4 & C & 0.980 & 0.987 & 0.992 & 0.987 & 0.983 & 0.755 \\
V & 4 & D & 0.974 & 0.982 & 0.985 & 0.985 & 0.975 & 0.708 \\
V & 6 & A & 0.970 & 0.968 & 0.965 & 0.957 & 0.964 & 0.500 \\
V & 6 & B & 0.827 & 0.857 & 0.847 & 0.859 & 0.851 & 0.510 \\
V & 6 & C & 0.753 & 0.699 & 0.755 & 0.763 & 0.692 & 0.495 \\
V & 6 & D & 0.797 & 0.788 & 0.764 & 0.790 & 0.799 & 0.515 \\
V & 6 & E & 0.757 & 0.754 & 0.755 & 0.743 & 0.753 & 0.470 \\
V & 6 & F & 0.798 & 0.806 & 0.774 & 0.785 & 0.777 & 0.510 \\
V & 8 & A & 0.903 & 0.905 & 0.912 & 0.916 & 0.909 & 0.420 \\
V & 8 & B & 0.828 & 0.823 & 0.841 & 0.813 & 0.800 & 0.345 \\
V & 8 & C & 0.683 & 0.676 & 0.663 & 0.618 & 0.663 & 0.375 \\
V & 8 & D & 0.643 & 0.623 & 0.638 & 0.654 & 0.658 & 0.338 \\
V & 8 & E & 0.618 & 0.628 & 0.631 & 0.627 & 0.595 & 0.388 \\
V & 8 & F & 0.660 & 0.676 & 0.709 & 0.643 & 0.619 & 0.357 \\
V & 8 & G & 0.685 & 0.652 & 0.626 & 0.664 & 0.618 & 0.412 \\
V & 8 & H & 0.724 & 0.731 & 0.718 & 0.746 & 0.720 & 0.365 \\
T & 4 & A & 0.994 & 0.995 & 0.994 & 0.991 & 0.995 & 0.792 \\
T & 4 & B & 0.979 & 0.986 & 0.983 & 0.991 & 0.980 & 0.745 \\
T & 4 & C & 0.975 & 0.974 & 0.986 & 0.977 & 0.971 & 0.755 \\
T & 4 & D & 0.990 & 0.989 & 0.990 & 0.990 & 0.984 & 0.708 \\
T & 6 & A & 0.975 & 0.975 & 0.981 & 0.977 & 0.973 & 0.500 \\
T & 6 & B & 0.891 & 0.915 & 0.897 & 0.905 & 0.899 & 0.510 \\
T & 6 & C & 0.843 & 0.845 & 0.852 & 0.843 & 0.852 & 0.495 \\
T & 6 & D & 0.776 & 0.784 & 0.796 & 0.760 & 0.778 & 0.515 \\
T & 6 & E & 0.826 & 0.847 & 0.846 & 0.854 & 0.826 & 0.470 \\
T & 6 & F & 0.884 & 0.883 & 0.879 & 0.875 & 0.873 & 0.510 \\
T & 8 & A & 0.899 & 0.916 & 0.910 & 0.887 & 0.899 & 0.420 \\
T & 8 & B & 0.759 & 0.797 & 0.772 & 0.768 & 0.761 & 0.345 \\
T & 8 & C & 0.693 & 0.712 & 0.664 & 0.728 & 0.666 & 0.375 \\
T & 8 & D & 0.649 & 0.682 & 0.656 & 0.636 & 0.683 & 0.338 \\
T & 8 & E & 0.701 & 0.674 & 0.715 & 0.754 & 0.713 & 0.388 \\
T & 8 & F & 0.722 & 0.686 & 0.709 & 0.701 & 0.765 & 0.357 \\
T & 8 & G & 0.716 & 0.719 & 0.705 & 0.713 & 0.712 & 0.412 \\
T & 8 & H & 0.791 & 0.815 & 0.823 & 0.813 & 0.806 & 0.365 \\
\end{longtable}\endgroup
\begingroup\normalsize\setlength{\tabcolsep}{4pt}\renewcommand{\arraystretch}{1.08}
\begin{longtable}{lrrrrrrrr}
\caption{Gemma-3-4B: candidate-specific membership AUC for each seed. A--H denote candidate IDs. Positive is the fraction of instances whose valid subset contains that candidate. The main figure averages these candidate-level results.}\label{tab:app_member_G4}\\
\toprule\rowcolor{meiSage} Input & $k$ & ID & 13 & 23 & 37 & 51 & 79 & Positive \\
\midrule\endfirsthead
\toprule\rowcolor{meiSage} Input & $k$ & ID & 13 & 23 & 37 & 51 & 79 & Positive \\
\midrule\endhead\bottomrule\endfoot
V & 4 & A & 0.992 & 0.991 & 0.992 & 0.981 & 0.995 & 0.792 \\
V & 4 & B & 0.976 & 0.972 & 0.966 & 0.978 & 0.971 & 0.745 \\
V & 4 & C & 0.975 & 0.986 & 0.982 & 0.982 & 0.981 & 0.755 \\
V & 4 & D & 0.973 & 0.982 & 0.986 & 0.980 & 0.978 & 0.708 \\
V & 6 & A & 0.927 & 0.916 & 0.928 & 0.916 & 0.920 & 0.500 \\
V & 6 & B & 0.859 & 0.870 & 0.862 & 0.845 & 0.872 & 0.510 \\
V & 6 & C & 0.851 & 0.865 & 0.855 & 0.844 & 0.857 & 0.495 \\
V & 6 & D & 0.815 & 0.795 & 0.785 & 0.816 & 0.804 & 0.515 \\
V & 6 & E & 0.832 & 0.825 & 0.829 & 0.819 & 0.830 & 0.470 \\
V & 6 & F & 0.784 & 0.778 & 0.807 & 0.777 & 0.787 & 0.510 \\
V & 8 & A & 0.812 & 0.805 & 0.816 & 0.806 & 0.818 & 0.420 \\
V & 8 & B & 0.676 & 0.677 & 0.684 & 0.645 & 0.619 & 0.345 \\
V & 8 & C & 0.675 & 0.649 & 0.661 & 0.692 & 0.684 & 0.375 \\
V & 8 & D & 0.695 & 0.595 & 0.647 & 0.673 & 0.637 & 0.338 \\
V & 8 & E & 0.756 & 0.695 & 0.729 & 0.716 & 0.740 & 0.388 \\
V & 8 & F & 0.667 & 0.681 & 0.669 & 0.667 & 0.671 & 0.357 \\
V & 8 & G & 0.745 & 0.725 & 0.736 & 0.743 & 0.729 & 0.412 \\
V & 8 & H & 0.697 & 0.714 & 0.692 & 0.686 & 0.677 & 0.365 \\
T & 4 & A & 0.995 & 0.998 & 0.999 & 0.996 & 0.998 & 0.792 \\
T & 4 & B & 0.979 & 0.971 & 0.978 & 0.983 & 0.983 & 0.745 \\
T & 4 & C & 0.969 & 0.964 & 0.984 & 0.982 & 0.985 & 0.755 \\
T & 4 & D & 0.985 & 0.990 & 0.983 & 0.991 & 0.989 & 0.708 \\
T & 6 & A & 0.956 & 0.961 & 0.972 & 0.965 & 0.963 & 0.500 \\
T & 6 & B & 0.913 & 0.935 & 0.939 & 0.924 & 0.917 & 0.510 \\
T & 6 & C & 0.780 & 0.795 & 0.768 & 0.795 & 0.784 & 0.495 \\
T & 6 & D & 0.805 & 0.767 & 0.802 & 0.773 & 0.789 & 0.515 \\
T & 6 & E & 0.799 & 0.800 & 0.803 & 0.831 & 0.814 & 0.470 \\
T & 6 & F & 0.855 & 0.857 & 0.864 & 0.851 & 0.863 & 0.510 \\
T & 8 & A & 0.925 & 0.931 & 0.913 & 0.933 & 0.928 & 0.420 \\
T & 8 & B & 0.760 & 0.795 & 0.716 & 0.764 & 0.724 & 0.345 \\
T & 8 & C & 0.638 & 0.620 & 0.615 & 0.656 & 0.663 & 0.375 \\
T & 8 & D & 0.591 & 0.607 & 0.613 & 0.578 & 0.603 & 0.338 \\
T & 8 & E & 0.604 & 0.619 & 0.627 & 0.592 & 0.636 & 0.388 \\
T & 8 & F & 0.637 & 0.632 & 0.568 & 0.577 & 0.642 & 0.357 \\
T & 8 & G & 0.686 & 0.676 & 0.668 & 0.646 & 0.690 & 0.412 \\
T & 8 & H & 0.763 & 0.772 & 0.764 & 0.776 & 0.761 & 0.365 \\
\end{longtable}\endgroup
\begingroup\normalsize\setlength{\tabcolsep}{4pt}\renewcommand{\arraystretch}{1.08}
\begin{longtable}{lrrrrrrrr}
\caption{Gemma-3-12B: candidate-specific membership AUC for each seed. A--H denote candidate IDs. Positive is the fraction of instances whose valid subset contains that candidate. The main figure averages these candidate-level results.}\label{tab:app_member_G12}\\
\toprule\rowcolor{meiSage} Input & $k$ & ID & 13 & 23 & 37 & 51 & 79 & Positive \\
\midrule\endfirsthead
\toprule\rowcolor{meiSage} Input & $k$ & ID & 13 & 23 & 37 & 51 & 79 & Positive \\
\midrule\endhead\bottomrule\endfoot
V & 4 & A & 0.997 & 0.998 & 0.996 & 0.989 & 0.992 & 0.792 \\
V & 4 & B & 0.975 & 0.980 & 0.979 & 0.980 & 0.976 & 0.745 \\
V & 4 & C & 0.967 & 0.968 & 0.970 & 0.972 & 0.965 & 0.755 \\
V & 4 & D & 0.979 & 0.986 & 0.976 & 0.980 & 0.981 & 0.708 \\
V & 6 & A & 0.972 & 0.968 & 0.970 & 0.962 & 0.977 & 0.500 \\
V & 6 & B & 0.884 & 0.905 & 0.891 & 0.890 & 0.883 & 0.510 \\
V & 6 & C & 0.790 & 0.809 & 0.761 & 0.791 & 0.774 & 0.495 \\
V & 6 & D & 0.791 & 0.777 & 0.781 & 0.778 & 0.783 & 0.515 \\
V & 6 & E & 0.799 & 0.820 & 0.811 & 0.821 & 0.793 & 0.470 \\
V & 6 & F & 0.815 & 0.788 & 0.791 & 0.827 & 0.791 & 0.510 \\
V & 8 & A & 0.919 & 0.917 & 0.925 & 0.921 & 0.927 & 0.420 \\
V & 8 & B & 0.838 & 0.827 & 0.802 & 0.776 & 0.799 & 0.345 \\
V & 8 & C & 0.678 & 0.675 & 0.671 & 0.661 & 0.674 & 0.375 \\
V & 8 & D & 0.724 & 0.684 & 0.673 & 0.701 & 0.697 & 0.338 \\
V & 8 & E & 0.707 & 0.673 & 0.701 & 0.666 & 0.688 & 0.388 \\
V & 8 & F & 0.636 & 0.624 & 0.631 & 0.660 & 0.640 & 0.357 \\
V & 8 & G & 0.704 & 0.667 & 0.666 & 0.652 & 0.702 & 0.412 \\
V & 8 & H & 0.732 & 0.714 & 0.722 & 0.751 & 0.735 & 0.365 \\
T & 4 & A & 0.999 & 0.990 & 0.999 & 0.997 & 0.997 & 0.792 \\
T & 4 & B & 0.991 & 0.987 & 0.986 & 0.989 & 0.993 & 0.745 \\
T & 4 & C & 0.979 & 0.986 & 0.981 & 0.985 & 0.989 & 0.755 \\
T & 4 & D & 0.997 & 0.995 & 0.997 & 0.996 & 0.996 & 0.708 \\
T & 6 & A & 0.983 & 0.973 & 0.981 & 0.979 & 0.987 & 0.500 \\
T & 6 & B & 0.918 & 0.920 & 0.925 & 0.916 & 0.916 & 0.510 \\
T & 6 & C & 0.802 & 0.815 & 0.790 & 0.794 & 0.809 & 0.495 \\
T & 6 & D & 0.749 & 0.741 & 0.728 & 0.764 & 0.745 & 0.515 \\
T & 6 & E & 0.851 & 0.820 & 0.831 & 0.831 & 0.848 & 0.470 \\
T & 6 & F & 0.862 & 0.835 & 0.875 & 0.874 & 0.891 & 0.510 \\
T & 8 & A & 0.954 & 0.954 & 0.953 & 0.955 & 0.944 & 0.420 \\
T & 8 & B & 0.785 & 0.791 & 0.784 & 0.780 & 0.756 & 0.345 \\
T & 8 & C & 0.664 & 0.677 & 0.669 & 0.668 & 0.679 & 0.375 \\
T & 8 & D & 0.673 & 0.648 & 0.640 & 0.650 & 0.657 & 0.338 \\
T & 8 & E & 0.655 & 0.666 & 0.667 & 0.651 & 0.678 & 0.388 \\
T & 8 & F & 0.656 & 0.587 & 0.592 & 0.590 & 0.592 & 0.357 \\
T & 8 & G & 0.799 & 0.793 & 0.776 & 0.795 & 0.782 & 0.412 \\
T & 8 & H & 0.788 & 0.805 & 0.778 & 0.820 & 0.767 & 0.365 \\
\end{longtable}\endgroup

\begingroup\normalsize\setlength{\tabcolsep}{4pt}\renewcommand{\arraystretch}{1.08}
\begin{longtable}{lrrrr}
\caption{Selected hidden-state indices for the five outer folds in each correctness run. These support interpretation of split sensitivity.}\label{tab:app_selection}\\
\toprule\rowcolor{meiSage} Model & Input & $k$ & Seed & Selected indices \\
\midrule\endfirsthead
\toprule\rowcolor{meiSage} Model & Input & $k$ & Seed & Selected indices \\
\midrule\endhead\bottomrule\endfoot
Q3 & V & 4 & 13 & 25,28,28,27,28 \\
Q3 & V & 4 & 23 & 28,26,26,29,22 \\
Q3 & V & 4 & 37 & 26,29,21,28,26 \\
Q3 & V & 4 & 51 & 26,28,22,21,21 \\
Q3 & V & 4 & 79 & 28,25,26,28,22 \\
Q3 & V & 6 & 13 & 2,28,28,27,5 \\
Q3 & V & 6 & 23 & 1,32,2,20,19 \\
Q3 & V & 6 & 37 & 28,25,19,14,24 \\
Q3 & V & 6 & 51 & 28,5,6,28,27 \\
Q3 & V & 6 & 79 & 28,28,28,28,23 \\
Q3 & V & 8 & 13 & 25,29,28,34,27 \\
Q3 & V & 8 & 23 & 28,23,27,28,15 \\
Q3 & V & 8 & 37 & 25,24,28,17,23 \\
Q3 & V & 8 & 51 & 23,28,25,27,25 \\
Q3 & V & 8 & 79 & 3,17,23,25,28 \\
Q3 & T & 4 & 13 & 17,17,28,24,28 \\
Q3 & T & 4 & 23 & 24,24,17,24,15 \\
Q3 & T & 4 & 37 & 17,17,34,31,24 \\
Q3 & T & 4 & 51 & 27,24,24,17,18 \\
Q3 & T & 4 & 79 & 17,24,27,17,14 \\
Q3 & T & 6 & 13 & 35,36,36,7,24 \\
Q3 & T & 6 & 23 & 31,26,22,36,20 \\
Q3 & T & 6 & 37 & 27,22,31,25,21 \\
Q3 & T & 6 & 51 & 34,32,28,25,32 \\
Q3 & T & 6 & 79 & 25,31,23,22,34 \\
Q3 & T & 8 & 13 & 29,22,23,22,27 \\
Q3 & T & 8 & 23 & 24,31,33,21,28 \\
Q3 & T & 8 & 37 & 22,9,30,23,22 \\
Q3 & T & 8 & 51 & 22,24,22,32,35 \\
Q3 & T & 8 & 79 & 22,22,31,23,35 \\
Q7 & T & 4 & 13 & 22,19,27,18,20 \\
Q7 & T & 4 & 23 & 14,12,19,9,9 \\
Q7 & T & 4 & 37 & 15,16,21,9,13 \\
Q7 & T & 4 & 51 & 15,19,19,8,15 \\
Q7 & T & 4 & 79 & 14,10,18,8,16 \\
Q7 & T & 6 & 13 & 28,26,26,1,22 \\
Q7 & T & 6 & 23 & 17,1,28,10,22 \\
Q7 & T & 6 & 37 & 4,28,20,4,19 \\
Q7 & T & 6 & 51 & 1,13,19,23,18 \\
Q7 & T & 6 & 79 & 22,26,7,10,21 \\
Q7 & T & 8 & 13 & 23,23,3,23,24 \\
Q7 & T & 8 & 23 & 23,23,23,23,6 \\
Q7 & T & 8 & 37 & 23,23,23,23,23 \\
Q7 & T & 8 & 51 & 23,23,24,23,6 \\
Q7 & T & 8 & 79 & 23,13,12,23,23 \\
G4 & V & 4 & 13 & 17,16,15,17,23 \\
G4 & V & 4 & 23 & 17,23,23,24,15 \\
G4 & V & 4 & 37 & 22,17,17,18,16 \\
G4 & V & 4 & 51 & 16,15,23,16,17 \\
G4 & V & 4 & 79 & 16,17,23,21,17 \\
G4 & V & 6 & 13 & 5,6,13,3,5 \\
G4 & V & 6 & 23 & 11,32,5,3,6 \\
G4 & V & 6 & 37 & 3,5,5,5,5 \\
G4 & V & 6 & 51 & 3,5,1,5,5 \\
G4 & V & 6 & 79 & 5,9,5,5,11 \\
G4 & V & 8 & 13 & 23,20,2,5,14 \\
G4 & V & 8 & 23 & 20,20,20,7,18 \\
G4 & V & 8 & 37 & 20,16,18,19,17 \\
G4 & V & 8 & 51 & 16,19,14,16,18 \\
G4 & V & 8 & 79 & 17,3,19,20,15 \\
G4 & T & 4 & 13 & 20,21,18,16,28 \\
G4 & T & 4 & 23 & 32,7,29,20,16 \\
G4 & T & 4 & 37 & 28,20,15,16,19 \\
G4 & T & 4 & 51 & 15,21,20,20,16 \\
G4 & T & 4 & 79 & 17,17,29,20,15 \\
G4 & T & 6 & 13 & 21,17,20,17,12 \\
G4 & T & 6 & 23 & 11,26,26,17,16 \\
G4 & T & 6 & 37 & 17,20,17,23,23 \\
G4 & T & 6 & 51 & 17,16,17,17,17 \\
G4 & T & 6 & 79 & 32,20,16,17,20 \\
G4 & T & 8 & 13 & 20,26,21,18,22 \\
G4 & T & 8 & 23 & 21,25,20,21,22 \\
G4 & T & 8 & 37 & 21,11,18,24,24 \\
G4 & T & 8 & 51 & 24,17,18,26,22 \\
G4 & T & 8 & 79 & 23,13,18,21,13 \\
G12 & V & 4 & 13 & 32,25,27,32,30 \\
G12 & V & 4 & 23 & 38,25,21,32,28 \\
G12 & V & 4 & 37 & 31,22,21,32,25 \\
G12 & V & 4 & 51 & 32,26,30,25,28 \\
G12 & V & 4 & 79 & 22,26,27,11,30 \\
G12 & V & 6 & 13 & 4,18,4,45,45 \\
G12 & V & 6 & 23 & 27,14,18,21,4 \\
G12 & V & 6 & 37 & 47,14,14,21,28 \\
G12 & V & 6 & 51 & 18,44,4,14,18 \\
G12 & V & 6 & 79 & 14,18,14,14,4 \\
G12 & V & 8 & 13 & 7,28,5,22,19 \\
G12 & V & 8 & 23 & 23,6,21,22,19 \\
G12 & V & 8 & 37 & 24,23,7,24,23 \\
G12 & V & 8 & 51 & 19,4,5,23,6 \\
G12 & V & 8 & 79 & 23,5,22,5,22 \\
G12 & T & 4 & 13 & 27,27,27,27,26 \\
G12 & T & 4 & 23 & 27,27,27,26,27 \\
G12 & T & 4 & 37 & 26,27,26,26,32 \\
G12 & T & 4 & 51 & 28,27,27,27,27 \\
G12 & T & 4 & 79 & 41,27,23,27,26 \\
G12 & T & 6 & 13 & 26,24,22,24,29 \\
G12 & T & 6 & 23 & 11,28,23,22,29 \\
G12 & T & 6 & 37 & 23,22,10,30,31 \\
G12 & T & 6 & 51 & 2,23,31,26,26 \\
G12 & T & 6 & 79 & 22,2,13,29,26 \\
G12 & T & 8 & 13 & 26,17,44,7,7 \\
G12 & T & 8 & 23 & 12,44,46,10,5 \\
G12 & T & 8 & 37 & 48,12,5,27,44 \\
G12 & T & 8 & 51 & 14,43,10,7,38 \\
G12 & T & 8 & 79 & 38,12,7,17,27 \\
\end{longtable}\endgroup

\section{Extended Activation-Patching Results}
\label{app:patching}
\subsection{Recovery at Larger Candidate Counts}
\label{app:recovery_counts}
The main cross-model figure compares four-candidate cohorts. The following panels supply six- and eight-candidate cohorts, displaying corresponding and mismatched donor rates separately. Their paired contrasts and intervals follow in a table. Separating the absolute rates from the contrast shows whether a change reflects corresponding-donor recovery, mismatched-donor success, or both.
\begin{figure}[!htbp]\centering
\includegraphics[width=\linewidth]{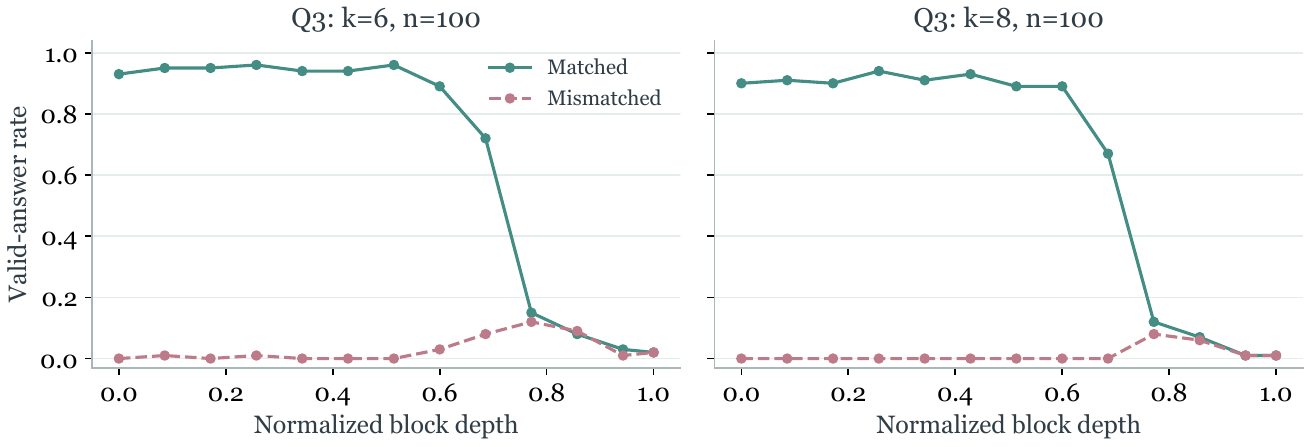}
\caption{Q3: full-prompt recovery at $k=6,8$, complementing the main $k=4$ comparisons. Both donor conditions use the same recipient pairs. Curves show absolute valid-answer rates. Points denote tested blocks.}
\label{fig:app_recovery_Q3}
\end{figure}
\begin{figure}[!htbp]\centering
\includegraphics[width=\linewidth]{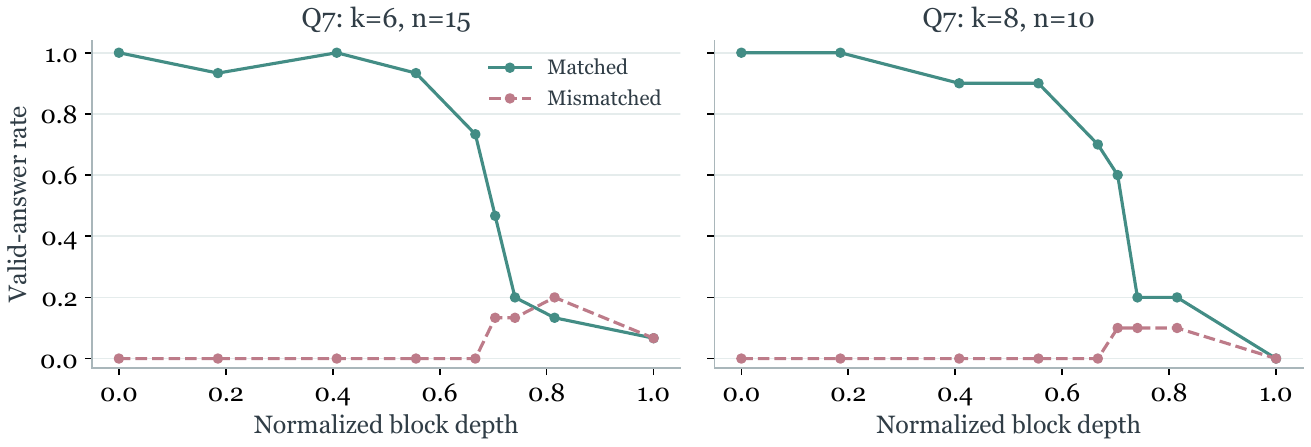}
\caption{Q7: full-prompt recovery at $k=6,8$, complementing the main $k=4$ comparisons. Both donor conditions use the same recipient pairs. Curves show absolute valid-answer rates. Points denote tested blocks.}
\label{fig:app_recovery_Q7}
\end{figure}
\begin{figure}[!htbp]\centering
\includegraphics[width=\linewidth]{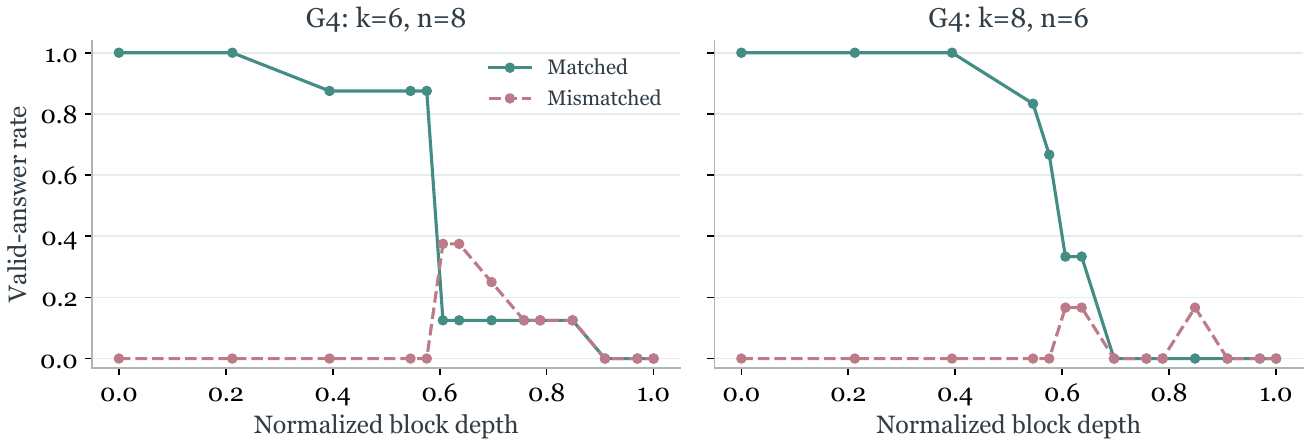}
\caption{G4: full-prompt recovery at $k=6,8$, complementing the main $k=4$ comparisons. Both donor conditions use the same recipient pairs. Curves show absolute valid-answer rates. Points denote tested blocks.}
\label{fig:app_recovery_G4}
\end{figure}
\begin{figure}[!htbp]\centering
\includegraphics[width=\linewidth]{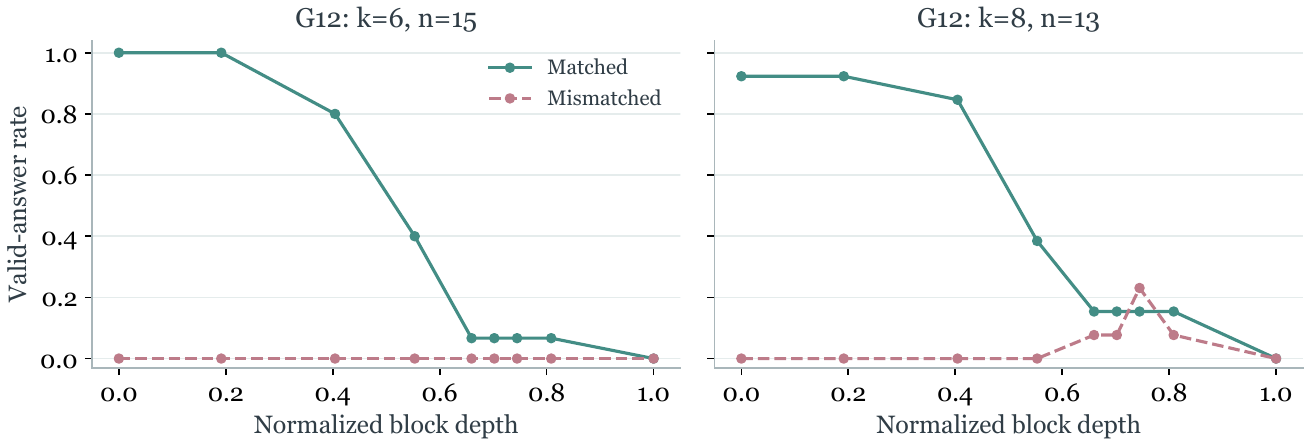}
\caption{G12: full-prompt recovery at $k=6,8$, complementing the main $k=4$ comparisons. Both donor conditions use the same recipient pairs. Curves show absolute valid-answer rates. Points denote tested blocks.}
\label{fig:app_recovery_G12}
\end{figure}
\begin{figure}[!htbp]\centering
\includegraphics[width=\linewidth]{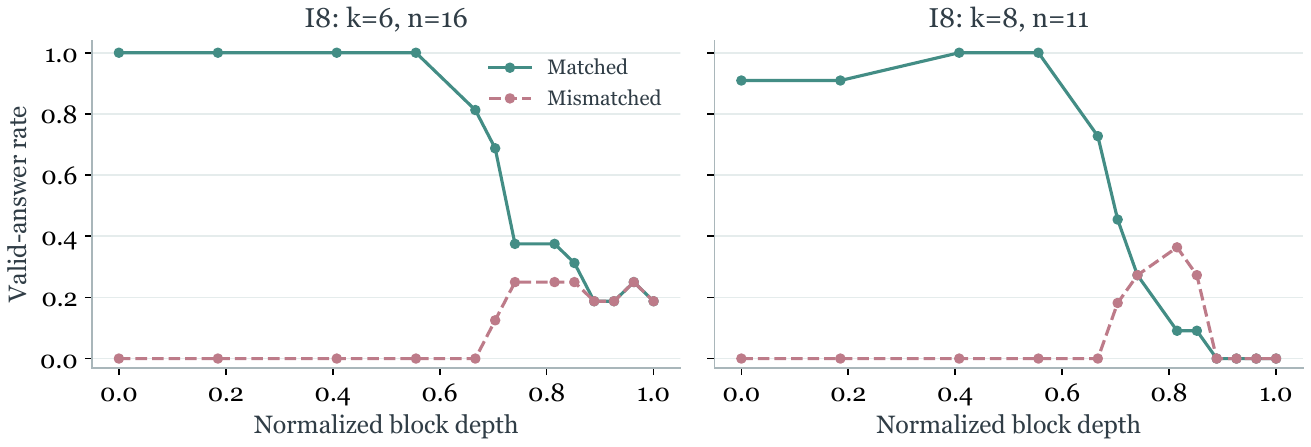}
\caption{I8: full-prompt recovery at $k=6,8$, complementing the main $k=4$ comparisons. Both donor conditions use the same recipient pairs. Curves show absolute valid-answer rates. Points denote tested blocks.}
\label{fig:app_recovery_I8}
\end{figure}

\FloatBarrier
\begingroup\normalsize\setlength{\tabcolsep}{4pt}\renewcommand{\arraystretch}{1.08}
\begin{longtable}{lrrrr}
\caption{Full-prompt paired donor contrasts at $k=6,8$. Intervals use 2,000 recipient-level bootstrap resamples and are pointwise. Cohort sizes appear in the corresponding recovery panels and main coverage table. Boundary intervals can collapse when all observed differences are identical.}\label{tab:app_intervals}\\
\toprule\rowcolor{meiSage} Model & $k$ & Block & $\Delta$ & 95\% interval \\
\midrule\endfirsthead
\toprule\rowcolor{meiSage} Model & $k$ & Block & $\Delta$ & 95\% interval \\
\midrule\endhead\bottomrule\endfoot
G12 & 6 & 0 & 1.000 & [1.000, 1.000] \\
G12 & 6 & 9 & 1.000 & [1.000, 1.000] \\
G12 & 6 & 19 & 0.800 & [0.600, 1.000] \\
G12 & 6 & 26 & 0.400 & [0.133, 0.667] \\
G12 & 6 & 31 & 0.067 & [0.000, 0.200] \\
G12 & 6 & 33 & 0.067 & [0.000, 0.200] \\
G12 & 6 & 35 & 0.067 & [0.000, 0.200] \\
G12 & 6 & 38 & 0.067 & [0.000, 0.200] \\
G12 & 6 & 47 & 0.000 & [0.000, 0.000] \\
G12 & 8 & 0 & 0.923 & [0.769, 1.000] \\
G12 & 8 & 9 & 0.923 & [0.769, 1.000] \\
G12 & 8 & 19 & 0.846 & [0.615, 1.000] \\
G12 & 8 & 26 & 0.385 & [0.154, 0.615] \\
G12 & 8 & 31 & 0.077 & [0.000, 0.231] \\
G12 & 8 & 33 & 0.077 & [0.000, 0.231] \\
G12 & 8 & 35 & -0.077 & [-0.231, 0.000] \\
G12 & 8 & 38 & 0.077 & [0.000, 0.231] \\
G12 & 8 & 47 & 0.000 & [0.000, 0.000] \\
G4 & 6 & 0 & 1.000 & [1.000, 1.000] \\
G4 & 6 & 7 & 1.000 & [1.000, 1.000] \\
G4 & 6 & 13 & 0.875 & [0.625, 1.000] \\
G4 & 6 & 18 & 0.875 & [0.625, 1.000] \\
G4 & 6 & 19 & 0.875 & [0.625, 1.000] \\
G4 & 6 & 20 & -0.250 & [-0.500, 0.000] \\
G4 & 6 & 21 & -0.250 & [-0.503, 0.000] \\
G4 & 6 & 23 & -0.125 & [-0.375, 0.000] \\
G4 & 6 & 25 & 0.000 & [-0.375, 0.375] \\
G4 & 6 & 26 & 0.000 & [-0.375, 0.375] \\
G4 & 6 & 28 & 0.000 & [-0.375, 0.375] \\
G4 & 6 & 30 & 0.000 & [0.000, 0.000] \\
G4 & 6 & 32 & 0.000 & [0.000, 0.000] \\
G4 & 6 & 33 & 0.000 & [0.000, 0.000] \\
G4 & 8 & 0 & 1.000 & [1.000, 1.000] \\
G4 & 8 & 7 & 1.000 & [1.000, 1.000] \\
G4 & 8 & 13 & 1.000 & [1.000, 1.000] \\
G4 & 8 & 18 & 0.833 & [0.500, 1.000] \\
G4 & 8 & 19 & 0.667 & [0.333, 1.000] \\
G4 & 8 & 20 & 0.167 & [0.000, 0.500] \\
G4 & 8 & 21 & 0.167 & [0.000, 0.500] \\
G4 & 8 & 23 & 0.000 & [0.000, 0.000] \\
G4 & 8 & 25 & 0.000 & [0.000, 0.000] \\
G4 & 8 & 26 & 0.000 & [0.000, 0.000] \\
G4 & 8 & 28 & -0.167 & [-0.500, 0.000] \\
G4 & 8 & 30 & 0.000 & [0.000, 0.000] \\
G4 & 8 & 32 & 0.000 & [0.000, 0.000] \\
G4 & 8 & 33 & 0.000 & [0.000, 0.000] \\
I8 & 6 & 0 & 1.000 & [1.000, 1.000] \\
I8 & 6 & 5 & 1.000 & [1.000, 1.000] \\
I8 & 6 & 11 & 1.000 & [1.000, 1.000] \\
I8 & 6 & 15 & 1.000 & [1.000, 1.000] \\
I8 & 6 & 18 & 0.812 & [0.625, 1.000] \\
I8 & 6 & 19 & 0.562 & [0.312, 0.812] \\
I8 & 6 & 20 & 0.125 & [-0.125, 0.375] \\
I8 & 6 & 22 & 0.125 & [-0.125, 0.375] \\
I8 & 6 & 23 & 0.062 & [-0.125, 0.250] \\
I8 & 6 & 24 & 0.000 & [0.000, 0.000] \\
I8 & 6 & 25 & 0.000 & [0.000, 0.000] \\
I8 & 6 & 26 & 0.000 & [0.000, 0.000] \\
I8 & 6 & 27 & 0.000 & [0.000, 0.000] \\
I8 & 8 & 0 & 0.909 & [0.727, 1.000] \\
I8 & 8 & 5 & 0.909 & [0.727, 1.000] \\
I8 & 8 & 11 & 1.000 & [1.000, 1.000] \\
I8 & 8 & 15 & 1.000 & [1.000, 1.000] \\
I8 & 8 & 18 & 0.727 & [0.455, 1.000] \\
I8 & 8 & 19 & 0.273 & [-0.182, 0.727] \\
I8 & 8 & 20 & 0.000 & [-0.455, 0.455] \\
I8 & 8 & 22 & -0.273 & [-0.545, 0.000] \\
I8 & 8 & 23 & -0.182 & [-0.455, 0.000] \\
I8 & 8 & 24 & 0.000 & [0.000, 0.000] \\
I8 & 8 & 25 & 0.000 & [0.000, 0.000] \\
I8 & 8 & 26 & 0.000 & [0.000, 0.000] \\
I8 & 8 & 27 & 0.000 & [0.000, 0.000] \\
Q3 & 6 & 0 & 0.930 & [0.880, 0.970] \\
Q3 & 6 & 3 & 0.940 & [0.890, 0.980] \\
Q3 & 6 & 6 & 0.950 & [0.900, 0.990] \\
Q3 & 6 & 9 & 0.950 & [0.900, 0.990] \\
Q3 & 6 & 12 & 0.940 & [0.890, 0.980] \\
Q3 & 6 & 15 & 0.940 & [0.890, 0.980] \\
Q3 & 6 & 18 & 0.960 & [0.920, 0.990] \\
Q3 & 6 & 21 & 0.860 & [0.790, 0.920] \\
Q3 & 6 & 24 & 0.640 & [0.550, 0.730] \\
Q3 & 6 & 27 & 0.030 & [-0.040, 0.100] \\
Q3 & 6 & 30 & -0.010 & [-0.070, 0.050] \\
Q3 & 6 & 33 & 0.020 & [0.000, 0.050] \\
Q3 & 6 & 35 & 0.000 & [0.000, 0.000] \\
Q3 & 8 & 0 & 0.900 & [0.840, 0.950] \\
Q3 & 8 & 3 & 0.910 & [0.850, 0.960] \\
Q3 & 8 & 6 & 0.900 & [0.840, 0.950] \\
Q3 & 8 & 9 & 0.940 & [0.890, 0.980] \\
Q3 & 8 & 12 & 0.910 & [0.850, 0.960] \\
Q3 & 8 & 15 & 0.930 & [0.880, 0.970] \\
Q3 & 8 & 18 & 0.890 & [0.820, 0.950] \\
Q3 & 8 & 21 & 0.890 & [0.830, 0.950] \\
Q3 & 8 & 24 & 0.670 & [0.580, 0.760] \\
Q3 & 8 & 27 & 0.040 & [-0.040, 0.120] \\
Q3 & 8 & 30 & 0.010 & [-0.060, 0.080] \\
Q3 & 8 & 33 & 0.000 & [0.000, 0.000] \\
Q3 & 8 & 35 & 0.000 & [0.000, 0.000] \\
Q7 & 6 & 0 & 1.000 & [1.000, 1.000] \\
Q7 & 6 & 5 & 0.933 & [0.800, 1.000] \\
Q7 & 6 & 11 & 1.000 & [1.000, 1.000] \\
Q7 & 6 & 15 & 0.933 & [0.800, 1.000] \\
Q7 & 6 & 18 & 0.733 & [0.533, 0.933] \\
Q7 & 6 & 19 & 0.333 & [0.133, 0.600] \\
Q7 & 6 & 20 & 0.067 & [0.000, 0.200] \\
Q7 & 6 & 22 & -0.067 & [-0.333, 0.200] \\
Q7 & 6 & 27 & 0.000 & [0.000, 0.000] \\
Q7 & 8 & 0 & 1.000 & [1.000, 1.000] \\
Q7 & 8 & 5 & 1.000 & [1.000, 1.000] \\
Q7 & 8 & 11 & 0.900 & [0.700, 1.000] \\
Q7 & 8 & 15 & 0.900 & [0.700, 1.000] \\
Q7 & 8 & 18 & 0.700 & [0.400, 1.000] \\
Q7 & 8 & 19 & 0.500 & [0.200, 0.800] \\
Q7 & 8 & 20 & 0.100 & [0.000, 0.300] \\
Q7 & 8 & 22 & 0.100 & [0.000, 0.300] \\
Q7 & 8 & 27 & 0.000 & [0.000, 0.000] \\
\end{longtable}\endgroup

\subsection{Token Positions and Changed-Item Selection}
\label{app:token_scopes}
The six-candidate scope comparison completes the main four- and eight-candidate panels. Restricted scopes use their shared eight tested blocks; full-prompt replacement additionally supplies the denser depth grid. Changed-item avoidance provides a complementary outcome: rejecting the modified item can accompany recovery, but a valid answer must also satisfy the remaining constraints. The corresponding table therefore reports avoidance separately from validity.
\begin{figure}[!htbp]\centering
\includegraphics[width=\linewidth]{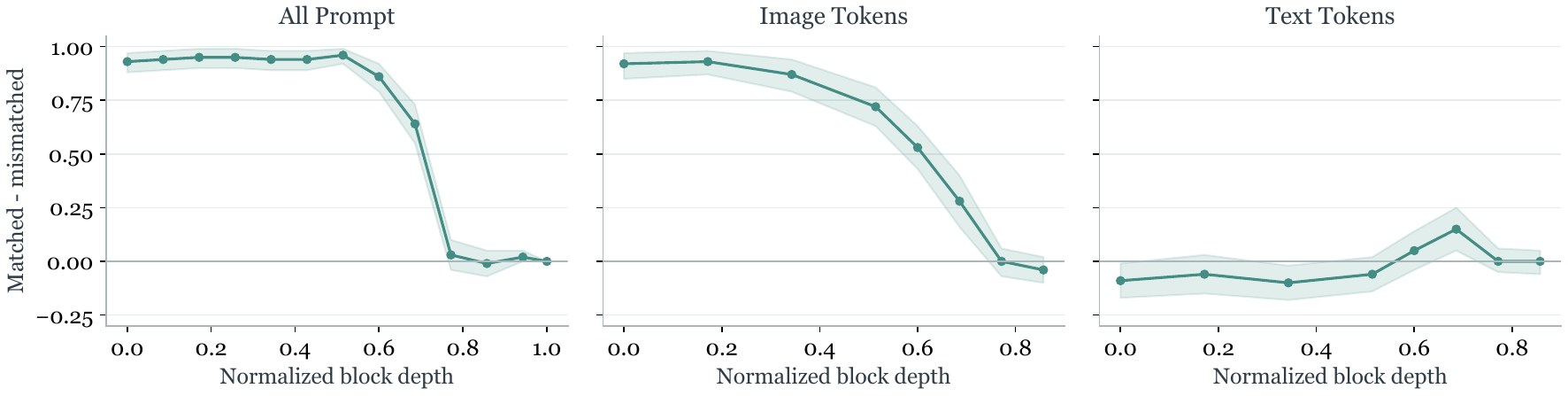}
\caption{Qwen-3B token-scope contrasts at $k=6$ on 100 matched recipient pairs. Shading shows pointwise 95\% paired-bootstrap intervals (2,000 resamples). Full prompt has 13 tested blocks; each restricted scope has eight. The main text reports the $k=4,8$ scope comparisons.}
\label{fig:app_scopes_k6}
\end{figure}

\FloatBarrier
\begingroup\normalsize\setlength{\tabcolsep}{4pt}\renewcommand{\arraystretch}{1.08}
\begin{longtable}{lrrr}
\caption{Decoy avoidance at Qwen-3B block 24. Each value is the fraction of patched outputs that omit the modified non-solution item. An output can avoid the decoy yet fail other constraints.}\label{tab:app_avoid}\\
\toprule\rowcolor{meiSage} Scope & $k$ & Matched & Mismatched \\
\midrule\endfirsthead
\toprule\rowcolor{meiSage} Scope & $k$ & Matched & Mismatched \\
\midrule\endhead\bottomrule\endfoot
Full & 4 & 0.879 & 0.000 \\
Full & 6 & 0.930 & 0.530 \\
Full & 8 & 0.920 & 0.630 \\
Image & 4 & 0.455 & 0.081 \\
Image & 6 & 0.790 & 0.500 \\
Image & 8 & 0.770 & 0.600 \\
Text & 4 & 0.444 & 0.081 \\
Text & 6 & 0.510 & 0.420 \\
Text & 8 & 0.620 & 0.490 \\
\end{longtable}\endgroup

\subsection{Selection of Intervention Cohorts}
\label{app:cohort_selection}
The clean/corrupt outcome counts show the pools from which recipients were selected. The main intervention-coverage table reports evaluated recipients and tested blocks; the table below instead reports selection outcomes before intervention. Recovery rates apply to these clean-correct/corrupt-wrong cohorts. Model differences in cohort composition and sample size are part of their interpretation, especially at larger candidate counts.
\begingroup\normalsize\setlength{\tabcolsep}{4pt}\renewcommand{\arraystretch}{1.08}
\begin{longtable}{lrrrrrr}
\caption{Pair-bank outcome counts before intervention subsampling. CC/CW/WC/WW refer to clean and corrupted correctness, respectively. Only CW pairs are eligible. Qwen-3B uses a larger base bank plus independent extensions; eligible counts can exceed the evaluated counts in the main table.}\label{tab:app_outcomes}\\
\toprule\rowcolor{meiSage} Model & Bank & $k$ & CC & CW & WC & WW \\
\midrule\endfirsthead
\toprule\rowcolor{meiSage} Model & Bank & $k$ & CC & CW & WC & WW \\
\midrule\endhead\bottomrule\endfoot
Q3 & base & 4 & 296 & 145 & 78 & 481 \\
Q3 & base & 6 & 87 & 54 & 46 & 1013 \\
Q3 & ext1 & 6 & 85 & 75 & 47 & 1193 \\
Q3 & base & 8 & 55 & 76 & 41 & 2828 \\
Q3 & ext1 & 8 & 35 & 42 & 17 & 1706 \\
Q7 & base & 4 & 27 & 37 & 6 & 30 \\
Q7 & base & 6 & 5 & 15 & 5 & 75 \\
Q7 & base & 8 & 6 & 10 & 6 & 78 \\
G4 & base & 4 & 28 & 34 & 2 & 36 \\
G4 & base & 6 & 8 & 8 & 4 & 80 \\
G4 & base & 8 & 5 & 6 & 1 & 88 \\
G12 & base & 4 & 20 & 47 & 0 & 33 \\
G12 & base & 6 & 7 & 16 & 3 & 74 \\
G12 & base & 8 & 6 & 13 & 0 & 81 \\
I8 & base & 4 & 18 & 60 & 1 & 21 \\
I8 & base & 6 & 6 & 16 & 7 & 71 \\
I8 & base & 8 & 6 & 11 & 2 & 81 \\
\end{longtable}\endgroup

\section{Uncertainty in Behavioral Results}
\label{app:behavior_ci}

We supplement the behavioral point estimates with pointwise 95\% Wilson score intervals. Each table reports the observed number of successes and its denominator; percentages and interval endpoints are rounded to two decimal places. These intervals quantify finite-instance uncertainty for the evaluated checkpoint and benchmark construction. They do not include variation across model training runs or alternative task generators. The binomial score intervals are marginal approximations for the benchmark's fixed mixture of generation settings.

For unconditional selection, retrieval, constraint ablation, and path finding, the unit is one generated instance, with 400 instances at each task size. For conditional selection, the denominator is the number of instances on which every constituent local answer is correct. The interval therefore describes validity within that retained cohort. For local recovery, we report the probability that all queried fields of all items in an instance are correct; grouping queries this way preserves the instance as the sampling unit. In particular, the graph-level intervals complement the perfect isolated-edge scores in Table~\ref{tab:path_diagnostic} without treating 16,000 correlated edge queries as independent trials.

The intervals cover individual rates, rather than paired differences between conditions, and are not adjusted for multiple comparisons. Comparisons involving shared instances require paired analysis; overlap or non-overlap of the marginal intervals alone is not a significance test. Probing variability across seeds and the paired-bootstrap intervals for activation interventions retain their separate definitions in Appendix~\ref{app:probe_intervention_protocols}.

Throughout these tables, Q3/Q7/Q32 denote Qwen2.5-VL-3B/7B/32B, G4/G12/G27 denote Gemma-3-4B/12B/27B, I8 denotes InternVL3-8B, and L7 denotes LLaVA-OneVision-7B. V/T denote visual-grid/text-oracle input. These are uncertainty summaries of the final experiments, not additional inference runs.

\begingroup\normalsize\setlength{\tabcolsep}{4pt}\renewcommand{\arraystretch}{1.08}
\begin{longtable}{lrrrr}
\caption{Selection validity with 95\% Wilson intervals. Each row uses 400 final benchmark instances.}\label{tab:app_ci_selection}\\
\toprule\rowcolor{meiSage} Model & Input & $k$ & Correct/$n$ & Validity [95\% CI] \\
\midrule\endfirsthead
\toprule\rowcolor{meiSage} Model & Input & $k$ & Correct/$n$ & Validity [95\% CI] \\
\midrule\endhead\bottomrule\endfoot
Q3 & V & 2 & 399/400 & 99.75 [98.60, 99.96] \\
Q3 & V & 4 & 212/400 & 53.00 [48.10, 57.84] \\
Q3 & V & 6 & 72/400 & 18.00 [14.55, 22.06] \\
Q3 & V & 8 & 23/400 & 5.75 [3.86, 8.48] \\
Q3 & T & 2 & 400/400 & 100.00 [99.05, 100.00] \\
Q3 & T & 4 & 271/400 & 67.75 [63.02, 72.14] \\
Q3 & T & 6 & 118/400 & 29.50 [25.24, 34.15] \\
Q3 & T & 8 & 62/400 & 15.50 [12.28, 19.37] \\
Q7 & V & 2 & 400/400 & 100.00 [99.05, 100.00] \\
Q7 & V & 4 & 304/400 & 76.00 [71.58, 79.93] \\
Q7 & V & 6 & 118/400 & 29.50 [25.24, 34.15] \\
Q7 & V & 8 & 49/400 & 12.25 [9.39, 15.83] \\
Q7 & T & 2 & 400/400 & 100.00 [99.05, 100.00] \\
Q7 & T & 4 & 332/400 & 83.00 [79.01, 86.36] \\
Q7 & T & 6 & 145/400 & 36.25 [31.69, 41.07] \\
Q7 & T & 8 & 77/400 & 19.25 [15.69, 23.40] \\
Q32 & V & 2 & 400/400 & 100.00 [99.05, 100.00] \\
Q32 & V & 4 & 362/400 & 90.50 [87.23, 93.00] \\
Q32 & V & 6 & 157/400 & 39.25 [34.59, 44.12] \\
Q32 & V & 8 & 74/400 & 18.50 [15.00, 22.60] \\
Q32 & T & 2 & 400/400 & 100.00 [99.05, 100.00] \\
Q32 & T & 4 & 346/400 & 86.50 [82.80, 89.50] \\
Q32 & T & 6 & 197/400 & 49.25 [44.38, 54.13] \\
Q32 & T & 8 & 113/400 & 28.25 [24.06, 32.85] \\
G4 & V & 2 & 400/400 & 100.00 [99.05, 100.00] \\
G4 & V & 4 & 210/400 & 52.50 [47.61, 57.35] \\
G4 & V & 6 & 63/400 & 15.75 [12.51, 19.64] \\
G4 & V & 8 & 41/400 & 10.25 [7.65, 13.61] \\
G4 & T & 2 & 400/400 & 100.00 [99.05, 100.00] \\
G4 & T & 4 & 294/400 & 73.50 [68.97, 77.59] \\
G4 & T & 6 & 131/400 & 32.75 [28.33, 37.49] \\
G4 & T & 8 & 66/400 & 16.50 [13.18, 20.45] \\
G12 & V & 2 & 400/400 & 100.00 [99.05, 100.00] \\
G12 & V & 4 & 287/400 & 71.75 [67.15, 75.94] \\
G12 & V & 6 & 94/400 & 23.50 [19.61, 27.90] \\
G12 & V & 8 & 50/400 & 12.50 [9.61, 16.10] \\
G12 & T & 2 & 400/400 & 100.00 [99.05, 100.00] \\
G12 & T & 4 & 243/400 & 60.75 [55.88, 65.41] \\
G12 & T & 6 & 71/400 & 17.75 [14.32, 21.80] \\
G12 & T & 8 & 39/400 & 9.75 [7.21, 13.05] \\
G27 & V & 2 & 400/400 & 100.00 [99.05, 100.00] \\
G27 & V & 4 & 282/400 & 70.50 [65.85, 74.76] \\
G27 & V & 6 & 135/400 & 33.75 [29.29, 38.52] \\
G27 & V & 8 & 79/400 & 19.75 [16.14, 23.93] \\
G27 & T & 2 & 400/400 & 100.00 [99.05, 100.00] \\
G27 & T & 4 & 341/400 & 85.25 [81.44, 88.39] \\
G27 & T & 6 & 192/400 & 48.00 [43.15, 52.89] \\
G27 & T & 8 & 134/400 & 33.50 [29.05, 38.26] \\
I8 & V & 2 & 400/400 & 100.00 [99.05, 100.00] \\
I8 & V & 4 & 304/400 & 76.00 [71.58, 79.93] \\
I8 & V & 6 & 118/400 & 29.50 [25.24, 34.15] \\
I8 & V & 8 & 56/400 & 14.00 [10.94, 17.74] \\
I8 & T & 2 & 400/400 & 100.00 [99.05, 100.00] \\
I8 & T & 4 & 339/400 & 84.75 [80.90, 87.94] \\
I8 & T & 6 & 196/400 & 49.00 [44.13, 53.89] \\
I8 & T & 8 & 130/400 & 32.50 [28.10, 37.24] \\
L7 & V & 2 & 396/400 & 99.00 [97.46, 99.61] \\
L7 & V & 4 & 172/400 & 43.00 [38.24, 47.90] \\
L7 & V & 6 & 44/400 & 11.00 [8.30, 14.45] \\
L7 & V & 8 & 19/400 & 4.75 [3.06, 7.30] \\
L7 & T & 2 & 400/400 & 100.00 [99.05, 100.00] \\
L7 & T & 4 & 320/400 & 80.00 [75.80, 83.63] \\
L7 & T & 6 & 136/400 & 34.00 [29.53, 38.77] \\
L7 & T & 8 & 61/400 & 15.25 [12.06, 19.10] \\
\end{longtable}\endgroup

\begingroup\normalsize\setlength{\tabcolsep}{4pt}\renewcommand{\arraystretch}{1.08}
\begin{longtable}{lrrr}
\caption{Isolated-card complete local recovery at the instance level: every field of every item must be correct. Denominator: 400 instances per row.}\label{tab:app_ci_isolated}\\
\toprule\rowcolor{meiSage} Model & $k$ & Retained/400 & Recovery [95\% CI] \\
\midrule\endfirsthead
\toprule\rowcolor{meiSage} Model & $k$ & Retained/400 & Recovery [95\% CI] \\
\midrule\endhead\bottomrule\endfoot
Q3 & 2 & 400/400 & 100.00 [99.05, 100.00] \\
Q3 & 4 & 400/400 & 100.00 [99.05, 100.00] \\
Q3 & 6 & 400/400 & 100.00 [99.05, 100.00] \\
Q3 & 8 & 400/400 & 100.00 [99.05, 100.00] \\
Q7 & 2 & 400/400 & 100.00 [99.05, 100.00] \\
Q7 & 4 & 400/400 & 100.00 [99.05, 100.00] \\
Q7 & 6 & 400/400 & 100.00 [99.05, 100.00] \\
Q7 & 8 & 400/400 & 100.00 [99.05, 100.00] \\
Q32 & 2 & 400/400 & 100.00 [99.05, 100.00] \\
Q32 & 4 & 400/400 & 100.00 [99.05, 100.00] \\
Q32 & 6 & 400/400 & 100.00 [99.05, 100.00] \\
Q32 & 8 & 400/400 & 100.00 [99.05, 100.00] \\
G4 & 2 & 400/400 & 100.00 [99.05, 100.00] \\
G4 & 4 & 400/400 & 100.00 [99.05, 100.00] \\
G4 & 6 & 400/400 & 100.00 [99.05, 100.00] \\
G4 & 8 & 400/400 & 100.00 [99.05, 100.00] \\
G12 & 2 & 400/400 & 100.00 [99.05, 100.00] \\
G12 & 4 & 400/400 & 100.00 [99.05, 100.00] \\
G12 & 6 & 400/400 & 100.00 [99.05, 100.00] \\
G12 & 8 & 400/400 & 100.00 [99.05, 100.00] \\
G27 & 2 & 400/400 & 100.00 [99.05, 100.00] \\
G27 & 4 & 400/400 & 100.00 [99.05, 100.00] \\
G27 & 6 & 400/400 & 100.00 [99.05, 100.00] \\
G27 & 8 & 400/400 & 100.00 [99.05, 100.00] \\
I8 & 2 & 400/400 & 100.00 [99.05, 100.00] \\
I8 & 4 & 400/400 & 100.00 [99.05, 100.00] \\
I8 & 6 & 400/400 & 100.00 [99.05, 100.00] \\
I8 & 8 & 399/400 & 99.75 [98.60, 99.96] \\
L7 & 2 & 400/400 & 100.00 [99.05, 100.00] \\
L7 & 4 & 400/400 & 100.00 [99.05, 100.00] \\
L7 & 6 & 400/400 & 100.00 [99.05, 100.00] \\
L7 & 8 & 400/400 & 100.00 [99.05, 100.00] \\
\end{longtable}\endgroup

\begingroup\normalsize\setlength{\tabcolsep}{4pt}\renewcommand{\arraystretch}{1.08}
\begin{longtable}{lrrrr}
\caption{Isolated-card conditional selection validity. The denominator is the retained cohort, not 400; these intervals describe that selected cohort.}\label{tab:app_ci_isolated_conditional}\\
\toprule\rowcolor{meiSage} Model & Input & $k$ & Correct/retained & Validity [95\% CI] \\
\midrule\endfirsthead
\toprule\rowcolor{meiSage} Model & Input & $k$ & Correct/retained & Validity [95\% CI] \\
\midrule\endhead\bottomrule\endfoot
Q3 & V & 2 & 399/400 & 99.75 [98.60, 99.96] \\
Q3 & T & 2 & 400/400 & 100.00 [99.05, 100.00] \\
Q3 & V & 4 & 212/400 & 53.00 [48.10, 57.84] \\
Q3 & T & 4 & 271/400 & 67.75 [63.02, 72.14] \\
Q3 & V & 6 & 72/400 & 18.00 [14.55, 22.06] \\
Q3 & T & 6 & 118/400 & 29.50 [25.24, 34.15] \\
Q3 & V & 8 & 23/400 & 5.75 [3.86, 8.48] \\
Q3 & T & 8 & 62/400 & 15.50 [12.28, 19.37] \\
Q7 & V & 2 & 400/400 & 100.00 [99.05, 100.00] \\
Q7 & T & 2 & 400/400 & 100.00 [99.05, 100.00] \\
Q7 & V & 4 & 304/400 & 76.00 [71.58, 79.93] \\
Q7 & T & 4 & 332/400 & 83.00 [79.01, 86.36] \\
Q7 & V & 6 & 118/400 & 29.50 [25.24, 34.15] \\
Q7 & T & 6 & 145/400 & 36.25 [31.69, 41.07] \\
Q7 & V & 8 & 49/400 & 12.25 [9.39, 15.83] \\
Q7 & T & 8 & 77/400 & 19.25 [15.69, 23.40] \\
Q32 & V & 2 & 400/400 & 100.00 [99.05, 100.00] \\
Q32 & T & 2 & 400/400 & 100.00 [99.05, 100.00] \\
Q32 & V & 4 & 362/400 & 90.50 [87.23, 93.00] \\
Q32 & T & 4 & 346/400 & 86.50 [82.80, 89.50] \\
Q32 & V & 6 & 157/400 & 39.25 [34.59, 44.12] \\
Q32 & T & 6 & 197/400 & 49.25 [44.38, 54.13] \\
Q32 & V & 8 & 74/400 & 18.50 [15.00, 22.60] \\
Q32 & T & 8 & 113/400 & 28.25 [24.06, 32.85] \\
G4 & V & 2 & 400/400 & 100.00 [99.05, 100.00] \\
G4 & T & 2 & 400/400 & 100.00 [99.05, 100.00] \\
G4 & V & 4 & 210/400 & 52.50 [47.61, 57.35] \\
G4 & T & 4 & 294/400 & 73.50 [68.97, 77.59] \\
G4 & V & 6 & 63/400 & 15.75 [12.51, 19.64] \\
G4 & T & 6 & 131/400 & 32.75 [28.33, 37.49] \\
G4 & V & 8 & 41/400 & 10.25 [7.65, 13.61] \\
G4 & T & 8 & 66/400 & 16.50 [13.18, 20.45] \\
G12 & V & 2 & 400/400 & 100.00 [99.05, 100.00] \\
G12 & T & 2 & 400/400 & 100.00 [99.05, 100.00] \\
G12 & V & 4 & 287/400 & 71.75 [67.15, 75.94] \\
G12 & T & 4 & 243/400 & 60.75 [55.88, 65.41] \\
G12 & V & 6 & 94/400 & 23.50 [19.61, 27.90] \\
G12 & T & 6 & 71/400 & 17.75 [14.32, 21.80] \\
G12 & V & 8 & 50/400 & 12.50 [9.61, 16.10] \\
G12 & T & 8 & 39/400 & 9.75 [7.21, 13.05] \\
G27 & V & 2 & 400/400 & 100.00 [99.05, 100.00] \\
G27 & T & 2 & 400/400 & 100.00 [99.05, 100.00] \\
G27 & V & 4 & 282/400 & 70.50 [65.85, 74.76] \\
G27 & T & 4 & 341/400 & 85.25 [81.44, 88.39] \\
G27 & V & 6 & 135/400 & 33.75 [29.29, 38.52] \\
G27 & T & 6 & 192/400 & 48.00 [43.15, 52.89] \\
G27 & V & 8 & 79/400 & 19.75 [16.14, 23.93] \\
G27 & T & 8 & 134/400 & 33.50 [29.05, 38.26] \\
I8 & V & 2 & 400/400 & 100.00 [99.05, 100.00] \\
I8 & T & 2 & 400/400 & 100.00 [99.05, 100.00] \\
I8 & V & 4 & 304/400 & 76.00 [71.58, 79.93] \\
I8 & T & 4 & 339/400 & 84.75 [80.90, 87.94] \\
I8 & V & 6 & 118/400 & 29.50 [25.24, 34.15] \\
I8 & T & 6 & 196/400 & 49.00 [44.13, 53.89] \\
I8 & V & 8 & 56/399 & 14.04 [10.97, 17.79] \\
I8 & T & 8 & 130/399 & 32.58 [28.17, 37.33] \\
L7 & V & 2 & 396/400 & 99.00 [97.46, 99.61] \\
L7 & T & 2 & 400/400 & 100.00 [99.05, 100.00] \\
L7 & V & 4 & 172/400 & 43.00 [38.24, 47.90] \\
L7 & T & 4 & 320/400 & 80.00 [75.80, 83.63] \\
L7 & V & 6 & 44/400 & 11.00 [8.30, 14.45] \\
L7 & T & 6 & 136/400 & 34.00 [29.53, 38.77] \\
L7 & V & 8 & 19/400 & 4.75 [3.06, 7.30] \\
L7 & T & 8 & 61/400 & 15.25 [12.06, 19.10] \\
\end{longtable}\endgroup

\begingroup\normalsize\setlength{\tabcolsep}{4pt}\renewcommand{\arraystretch}{1.08}
\begin{longtable}{lrrr}
\caption{Same-grid complete local recovery at the instance level: every field of every item must be correct. Denominator: 400 instances per row.}\label{tab:app_ci_grid}\\
\toprule\rowcolor{meiSage} Model & $k$ & Retained/400 & Recovery [95\% CI] \\
\midrule\endfirsthead
\toprule\rowcolor{meiSage} Model & $k$ & Retained/400 & Recovery [95\% CI] \\
\midrule\endhead\bottomrule\endfoot
Q3 & 2 & 400/400 & 100.00 [99.05, 100.00] \\
Q3 & 4 & 400/400 & 100.00 [99.05, 100.00] \\
Q3 & 6 & 376/400 & 94.00 [91.23, 95.94] \\
Q3 & 8 & 362/400 & 90.50 [87.23, 93.00] \\
G4 & 2 & 400/400 & 100.00 [99.05, 100.00] \\
G4 & 4 & 356/400 & 89.00 [85.55, 91.70] \\
G4 & 6 & 255/400 & 63.75 [58.93, 68.31] \\
G4 & 8 & 142/400 & 35.50 [30.97, 40.31] \\
I8 & 2 & 400/400 & 100.00 [99.05, 100.00] \\
I8 & 4 & 386/400 & 96.50 [94.21, 97.90] \\
I8 & 6 & 383/400 & 95.75 [93.30, 97.33] \\
I8 & 8 & 389/400 & 97.25 [95.14, 98.46] \\
L7 & 2 & 400/400 & 100.00 [99.05, 100.00] \\
L7 & 4 & 331/400 & 82.75 [78.74, 86.14] \\
L7 & 6 & 15/400 & 3.75 [2.29, 6.09] \\
L7 & 8 & 8/400 & 2.00 [1.02, 3.90] \\
\end{longtable}\endgroup

\begingroup\normalsize\setlength{\tabcolsep}{4pt}\renewcommand{\arraystretch}{1.08}
\begin{longtable}{lrrrr}
\caption{Same-grid conditional selection validity. The denominator is the retained cohort, not 400; these intervals describe that selected cohort.}\label{tab:app_ci_grid_conditional}\\
\toprule\rowcolor{meiSage} Model & Input & $k$ & Correct/retained & Validity [95\% CI] \\
\midrule\endfirsthead
\toprule\rowcolor{meiSage} Model & Input & $k$ & Correct/retained & Validity [95\% CI] \\
\midrule\endhead\bottomrule\endfoot
Q3 & V & 2 & 399/400 & 99.75 [98.60, 99.96] \\
Q3 & V & 4 & 212/400 & 53.00 [48.10, 57.84] \\
Q3 & V & 6 & 66/376 & 17.55 [14.04, 21.72] \\
Q3 & V & 8 & 21/362 & 5.80 [3.83, 8.71] \\
G4 & V & 2 & 400/400 & 100.00 [99.05, 100.00] \\
G4 & V & 4 & 192/356 & 53.93 [48.74, 59.04] \\
G4 & V & 6 & 37/255 & 14.51 [10.71, 19.36] \\
G4 & V & 8 & 15/142 & 10.56 [6.51, 16.70] \\
I8 & V & 2 & 400/400 & 100.00 [99.05, 100.00] \\
I8 & V & 4 & 295/386 & 76.42 [71.94, 80.39] \\
I8 & V & 6 & 113/383 & 29.50 [25.16, 34.26] \\
I8 & V & 8 & 52/389 & 13.37 [10.34, 17.11] \\
L7 & V & 2 & 396/400 & 99.00 [97.46, 99.61] \\
L7 & V & 4 & 141/331 & 42.60 [37.39, 47.98] \\
L7 & V & 6 & 2/15 & 13.33 [3.74, 37.88] \\
L7 & V & 8 & 1/8 & 12.50 [2.24, 47.09] \\
\end{longtable}\endgroup

\begingroup\normalsize\setlength{\tabcolsep}{4pt}\renewcommand{\arraystretch}{1.08}
\begin{longtable}{lrrrr}
\caption{Target-set retrieval with 95\% Wilson intervals; 400 instances per row.}\label{tab:app_ci_retrieval}\\
\toprule\rowcolor{meiSage} Model & Input & $k$ & Correct/$n$ & Rate [95\% CI] \\
\midrule\endfirsthead
\toprule\rowcolor{meiSage} Model & Input & $k$ & Correct/$n$ & Rate [95\% CI] \\
\midrule\endhead\bottomrule\endfoot
Q3 & V & 2 & 400/400 & 100.00 [99.05, 100.00] \\
Q3 & V & 4 & 349/400 & 87.25 [83.62, 90.17] \\
Q3 & V & 6 & 359/400 & 89.75 [86.39, 92.35] \\
Q3 & V & 8 & 341/400 & 85.25 [81.44, 88.39] \\
Q3 & T & 2 & 400/400 & 100.00 [99.05, 100.00] \\
Q3 & T & 4 & 332/400 & 83.00 [79.01, 86.36] \\
Q3 & T & 6 & 334/400 & 83.50 [79.55, 86.82] \\
Q3 & T & 8 & 320/400 & 80.00 [75.80, 83.63] \\
Q7 & V & 2 & 400/400 & 100.00 [99.05, 100.00] \\
Q7 & V & 4 & 400/400 & 100.00 [99.05, 100.00] \\
Q7 & V & 6 & 399/400 & 99.75 [98.60, 99.96] \\
Q7 & V & 8 & 390/400 & 97.50 [95.46, 98.64] \\
Q7 & T & 2 & 400/400 & 100.00 [99.05, 100.00] \\
Q7 & T & 4 & 400/400 & 100.00 [99.05, 100.00] \\
Q7 & T & 6 & 393/400 & 98.25 [96.43, 99.15] \\
Q7 & T & 8 & 377/400 & 94.25 [91.52, 96.14] \\
G4 & V & 2 & 400/400 & 100.00 [99.05, 100.00] \\
G4 & V & 4 & 351/400 & 87.75 [84.17, 90.61] \\
G4 & V & 6 & 331/400 & 82.75 [78.74, 86.14] \\
G4 & V & 8 & 263/400 & 65.75 [60.97, 70.23] \\
G4 & T & 2 & 400/400 & 100.00 [99.05, 100.00] \\
G4 & T & 4 & 394/400 & 98.50 [96.77, 99.31] \\
G4 & T & 6 & 363/400 & 90.75 [87.51, 93.21] \\
G4 & T & 8 & 305/400 & 76.25 [71.84, 80.16] \\
G12 & V & 2 & 400/400 & 100.00 [99.05, 100.00] \\
G12 & V & 4 & 399/400 & 99.75 [98.60, 99.96] \\
G12 & V & 6 & 394/400 & 98.50 [96.77, 99.31] \\
G12 & V & 8 & 389/400 & 97.25 [95.14, 98.46] \\
G12 & T & 2 & 400/400 & 100.00 [99.05, 100.00] \\
G12 & T & 4 & 400/400 & 100.00 [99.05, 100.00] \\
G12 & T & 6 & 398/400 & 99.50 [98.20, 99.86] \\
G12 & T & 8 & 392/400 & 98.00 [96.10, 98.98] \\
\end{longtable}\endgroup

\begingroup\normalsize\setlength{\tabcolsep}{4pt}\renewcommand{\arraystretch}{1.08}
\begin{longtable}{lrrrrr}
\caption{Constraint-ablation validity with 95\% Wilson intervals; 400 instances per row.}\label{tab:app_ci_ablation}\\
\toprule\rowcolor{meiSage} Model & Input & $k$ & Scope & Correct/$n$ & Rate [95\% CI] \\
\midrule\endfirsthead
\toprule\rowcolor{meiSage} Model & Input & $k$ & Scope & Correct/$n$ & Rate [95\% CI] \\
\midrule\endhead\bottomrule\endfoot
Q3 & V & 2 & Full & 398/400 & 99.50 [98.20, 99.86] \\
Q3 & V & 4 & Full & 208/400 & 52.00 [47.11, 56.85] \\
Q3 & V & 6 & Full & 73/400 & 18.25 [14.77, 22.33] \\
Q3 & V & 8 & Full & 24/400 & 6.00 [4.06, 8.77] \\
Q3 & V & 2 & Num. & 400/400 & 100.00 [99.05, 100.00] \\
Q3 & V & 4 & Num. & 288/400 & 72.00 [67.41, 76.17] \\
Q3 & V & 6 & Num. & 185/400 & 46.25 [41.42, 51.15] \\
Q3 & V & 8 & Num. & 157/400 & 39.25 [34.59, 44.12] \\
Q3 & V & 2 & Cat. & 396/400 & 99.00 [97.46, 99.61] \\
Q3 & V & 4 & Cat. & 389/400 & 97.25 [95.14, 98.46] \\
Q3 & V & 6 & Cat. & 387/400 & 96.75 [94.52, 98.09] \\
Q3 & V & 8 & Cat. & 361/400 & 90.25 [86.95, 92.79] \\
Q3 & V & 2 & Rel. & 400/400 & 100.00 [99.05, 100.00] \\
Q3 & V & 4 & Rel. & 247/400 & 61.75 [56.90, 66.38] \\
Q3 & V & 6 & Rel. & 89/400 & 22.25 [18.45, 26.58] \\
Q3 & V & 8 & Rel. & 47/400 & 11.75 [8.95, 15.28] \\
Q3 & T & 2 & Full & 400/400 & 100.00 [99.05, 100.00] \\
Q3 & T & 4 & Full & 278/400 & 69.50 [64.82, 73.81] \\
Q3 & T & 6 & Full & 126/400 & 31.50 [27.14, 36.21] \\
Q3 & T & 8 & Full & 65/400 & 16.25 [12.96, 20.18] \\
Q3 & T & 2 & Num. & 400/400 & 100.00 [99.05, 100.00] \\
Q3 & T & 4 & Num. & 319/400 & 79.75 [75.54, 83.40] \\
Q3 & T & 6 & Num. & 186/400 & 46.50 [41.67, 51.40] \\
Q3 & T & 8 & Num. & 132/400 & 33.00 [28.57, 37.75] \\
Q3 & T & 2 & Cat. & 400/400 & 100.00 [99.05, 100.00] \\
Q3 & T & 4 & Cat. & 386/400 & 96.50 [94.21, 97.90] \\
Q3 & T & 6 & Cat. & 383/400 & 95.75 [93.30, 97.33] \\
Q3 & T & 8 & Cat. & 360/400 & 90.00 [86.67, 92.57] \\
Q3 & T & 2 & Rel. & 400/400 & 100.00 [99.05, 100.00] \\
Q3 & T & 4 & Rel. & 326/400 & 81.50 [77.40, 85.00] \\
Q3 & T & 6 & Rel. & 202/400 & 50.50 [45.62, 55.37] \\
Q3 & T & 8 & Rel. & 157/400 & 39.25 [34.59, 44.12] \\
Q7 & V & 2 & Full & 400/400 & 100.00 [99.05, 100.00] \\
Q7 & V & 4 & Full & 298/400 & 74.50 [70.01, 78.52] \\
Q7 & V & 6 & Full & 121/400 & 30.25 [25.95, 34.92] \\
Q7 & V & 8 & Full & 53/400 & 13.25 [10.27, 16.92] \\
Q7 & V & 2 & Num. & 400/400 & 100.00 [99.05, 100.00] \\
Q7 & V & 4 & Num. & 345/400 & 86.25 [82.53, 89.28] \\
Q7 & V & 6 & Num. & 258/400 & 64.50 [59.69, 69.03] \\
Q7 & V & 8 & Num. & 198/400 & 49.50 [44.63, 54.38] \\
Q7 & V & 2 & Cat. & 400/400 & 100.00 [99.05, 100.00] \\
Q7 & V & 4 & Cat. & 396/400 & 99.00 [97.46, 99.61] \\
Q7 & V & 6 & Cat. & 390/400 & 97.50 [95.46, 98.64] \\
Q7 & V & 8 & Cat. & 381/400 & 95.25 [92.70, 96.94] \\
Q7 & V & 2 & Rel. & 400/400 & 100.00 [99.05, 100.00] \\
Q7 & V & 4 & Rel. & 354/400 & 88.50 [85.00, 91.27] \\
Q7 & V & 6 & Rel. & 198/400 & 49.50 [44.63, 54.38] \\
Q7 & V & 8 & Rel. & 119/400 & 29.75 [25.48, 34.41] \\
Q7 & T & 2 & Full & 400/400 & 100.00 [99.05, 100.00] \\
Q7 & T & 4 & Full & 330/400 & 82.50 [78.47, 85.91] \\
Q7 & T & 6 & Full & 160/400 & 40.00 [35.32, 44.87] \\
Q7 & T & 8 & Full & 83/400 & 20.75 [17.06, 24.99] \\
Q7 & T & 2 & Num. & 400/400 & 100.00 [99.05, 100.00] \\
Q7 & T & 4 & Num. & 329/400 & 82.25 [78.20, 85.68] \\
Q7 & T & 6 & Num. & 200/400 & 50.00 [45.12, 54.88] \\
Q7 & T & 8 & Num. & 170/400 & 42.50 [37.75, 47.39] \\
Q7 & T & 2 & Cat. & 400/400 & 100.00 [99.05, 100.00] \\
Q7 & T & 4 & Cat. & 396/400 & 99.00 [97.46, 99.61] \\
Q7 & T & 6 & Cat. & 379/400 & 94.75 [92.11, 96.54] \\
Q7 & T & 8 & Cat. & 377/400 & 94.25 [91.52, 96.14] \\
Q7 & T & 2 & Rel. & 400/400 & 100.00 [99.05, 100.00] \\
Q7 & T & 4 & Rel. & 351/400 & 87.75 [84.17, 90.61] \\
Q7 & T & 6 & Rel. & 211/400 & 52.75 [47.85, 57.59] \\
Q7 & T & 8 & Rel. & 166/400 & 41.50 [36.77, 46.39] \\
G4 & V & 2 & Full & 400/400 & 100.00 [99.05, 100.00] \\
G4 & V & 4 & Full & 226/400 & 56.50 [51.60, 61.27] \\
G4 & V & 6 & Full & 64/400 & 16.00 [12.73, 19.91] \\
G4 & V & 8 & Full & 47/400 & 11.75 [8.95, 15.28] \\
G4 & V & 2 & Num. & 400/400 & 100.00 [99.05, 100.00] \\
G4 & V & 4 & Num. & 259/400 & 64.75 [59.95, 69.27] \\
G4 & V & 6 & Num. & 183/400 & 45.75 [40.93, 50.65] \\
G4 & V & 8 & Num. & 149/400 & 37.25 [32.65, 42.09] \\
G4 & V & 2 & Cat. & 400/400 & 100.00 [99.05, 100.00] \\
G4 & V & 4 & Cat. & 394/400 & 98.50 [96.77, 99.31] \\
G4 & V & 6 & Cat. & 376/400 & 94.00 [91.23, 95.94] \\
G4 & V & 8 & Cat. & 364/400 & 91.00 [87.79, 93.43] \\
G4 & V & 2 & Rel. & 400/400 & 100.00 [99.05, 100.00] \\
G4 & V & 4 & Rel. & 331/400 & 82.75 [78.74, 86.14] \\
G4 & V & 6 & Rel. & 132/400 & 33.00 [28.57, 37.75] \\
G4 & V & 8 & Rel. & 56/400 & 14.00 [10.94, 17.74] \\
G4 & T & 2 & Full & 400/400 & 100.00 [99.05, 100.00] \\
G4 & T & 4 & Full & 288/400 & 72.00 [67.41, 76.17] \\
G4 & T & 6 & Full & 134/400 & 33.50 [29.05, 38.26] \\
G4 & T & 8 & Full & 61/400 & 15.25 [12.06, 19.10] \\
G4 & T & 2 & Num. & 400/400 & 100.00 [99.05, 100.00] \\
G4 & T & 4 & Num. & 311/400 & 77.75 [73.42, 81.55] \\
G4 & T & 6 & Num. & 207/400 & 51.75 [46.86, 56.61] \\
G4 & T & 8 & Num. & 192/400 & 48.00 [43.15, 52.89] \\
G4 & T & 2 & Cat. & 400/400 & 100.00 [99.05, 100.00] \\
G4 & T & 4 & Cat. & 393/400 & 98.25 [96.43, 99.15] \\
G4 & T & 6 & Cat. & 379/400 & 94.75 [92.11, 96.54] \\
G4 & T & 8 & Cat. & 356/400 & 89.00 [85.55, 91.70] \\
G4 & T & 2 & Rel. & 400/400 & 100.00 [99.05, 100.00] \\
G4 & T & 4 & Rel. & 321/400 & 80.25 [76.07, 83.86] \\
G4 & T & 6 & Rel. & 198/400 & 49.50 [44.63, 54.38] \\
G4 & T & 8 & Rel. & 148/400 & 37.00 [32.41, 41.83] \\
G12 & V & 2 & Full & 400/400 & 100.00 [99.05, 100.00] \\
G12 & V & 4 & Full & 287/400 & 71.75 [67.15, 75.94] \\
G12 & V & 6 & Full & 105/400 & 26.25 [22.18, 30.77] \\
G12 & V & 8 & Full & 57/400 & 14.25 [11.16, 18.02] \\
G12 & V & 2 & Num. & 400/400 & 100.00 [99.05, 100.00] \\
G12 & V & 4 & Num. & 320/400 & 80.00 [75.80, 83.63] \\
G12 & V & 6 & Num. & 209/400 & 52.25 [47.36, 57.10] \\
G12 & V & 8 & Num. & 223/400 & 55.75 [50.85, 60.54] \\
G12 & V & 2 & Cat. & 400/400 & 100.00 [99.05, 100.00] \\
G12 & V & 4 & Cat. & 395/400 & 98.75 [97.11, 99.46] \\
G12 & V & 6 & Cat. & 389/400 & 97.25 [95.14, 98.46] \\
G12 & V & 8 & Cat. & 383/400 & 95.75 [93.30, 97.33] \\
G12 & V & 2 & Rel. & 400/400 & 100.00 [99.05, 100.00] \\
G12 & V & 4 & Rel. & 347/400 & 86.75 [83.08, 89.73] \\
G12 & V & 6 & Rel. & 235/400 & 58.75 [53.86, 63.47] \\
G12 & V & 8 & Rel. & 192/400 & 48.00 [43.15, 52.89] \\
G12 & T & 2 & Full & 400/400 & 100.00 [99.05, 100.00] \\
G12 & T & 4 & Full & 242/400 & 60.50 [55.63, 65.17] \\
G12 & T & 6 & Full & 77/400 & 19.25 [15.69, 23.40] \\
G12 & T & 8 & Full & 46/400 & 11.50 [8.73, 15.00] \\
G12 & T & 2 & Num. & 400/400 & 100.00 [99.05, 100.00] \\
G12 & T & 4 & Num. & 309/400 & 77.25 [72.89, 81.09] \\
G12 & T & 6 & Num. & 250/400 & 62.50 [57.66, 67.10] \\
G12 & T & 8 & Num. & 257/400 & 64.25 [59.44, 68.79] \\
G12 & T & 2 & Cat. & 400/400 & 100.00 [99.05, 100.00] \\
G12 & T & 4 & Cat. & 387/400 & 96.75 [94.52, 98.09] \\
G12 & T & 6 & Cat. & 383/400 & 95.75 [93.30, 97.33] \\
G12 & T & 8 & Cat. & 374/400 & 93.50 [90.65, 95.53] \\
G12 & T & 2 & Rel. & 400/400 & 100.00 [99.05, 100.00] \\
G12 & T & 4 & Rel. & 322/400 & 80.50 [76.33, 84.08] \\
G12 & T & 6 & Rel. & 259/400 & 64.75 [59.95, 69.27] \\
G12 & T & 8 & Rel. & 233/400 & 58.25 [53.36, 62.98] \\
\end{longtable}\endgroup

\begingroup\normalsize\setlength{\tabcolsep}{4pt}\renewcommand{\arraystretch}{1.08}
\begin{longtable}{lrrrr}
\caption{Exact-path accuracy with 95\% Wilson intervals; 400 instances per row.}\label{tab:app_ci_path}\\
\toprule\rowcolor{meiSage} Model & Input & $k$ & Correct/$n$ & Rate [95\% CI] \\
\midrule\endfirsthead
\toprule\rowcolor{meiSage} Model & Input & $k$ & Correct/$n$ & Rate [95\% CI] \\
\midrule\endhead\bottomrule\endfoot
Q3 & V & 4 & 392/400 & 98.00 [96.10, 98.98] \\
Q3 & V & 8 & 324/400 & 81.00 [76.87, 84.54] \\
Q3 & V & 12 & 212/400 & 53.00 [48.10, 57.84] \\
Q3 & V & 16 & 156/400 & 39.00 [34.35, 43.86] \\
Q3 & T & 4 & 385/400 & 96.25 [93.91, 97.71] \\
Q3 & T & 8 & 224/400 & 56.00 [51.10, 60.78] \\
Q3 & T & 12 & 166/400 & 41.50 [36.77, 46.39] \\
Q3 & T & 16 & 111/400 & 27.75 [23.59, 32.33] \\
Q7 & V & 4 & 399/400 & 99.75 [98.60, 99.96] \\
Q7 & V & 8 & 325/400 & 81.25 [77.13, 84.77] \\
Q7 & V & 12 & 213/400 & 53.25 [48.35, 58.09] \\
Q7 & V & 16 & 170/400 & 42.50 [37.75, 47.39] \\
Q7 & T & 4 & 396/400 & 99.00 [97.46, 99.61] \\
Q7 & T & 8 & 332/400 & 83.00 [79.01, 86.36] \\
Q7 & T & 12 & 243/400 & 60.75 [55.88, 65.41] \\
Q7 & T & 16 & 182/400 & 45.50 [40.69, 50.40] \\
G4 & V & 4 & 376/400 & 94.00 [91.23, 95.94] \\
G4 & V & 8 & 193/400 & 48.25 [43.39, 53.14] \\
G4 & V & 12 & 92/400 & 23.00 [19.14, 27.37] \\
G4 & V & 16 & 53/400 & 13.25 [10.27, 16.92] \\
G4 & T & 4 & 384/400 & 96.00 [93.60, 97.52] \\
G4 & T & 8 & 308/400 & 77.00 [72.63, 80.86] \\
G4 & T & 12 & 227/400 & 56.75 [51.85, 61.52] \\
G4 & T & 16 & 172/400 & 43.00 [38.24, 47.90] \\
G12 & V & 4 & 397/400 & 99.25 [97.82, 99.74] \\
G12 & V & 8 & 338/400 & 84.50 [80.63, 87.72] \\
G12 & V & 12 & 188/400 & 47.00 [42.16, 51.90] \\
G12 & V & 16 & 162/400 & 40.50 [35.80, 45.38] \\
G12 & T & 4 & 397/400 & 99.25 [97.82, 99.74] \\
G12 & T & 8 & 353/400 & 88.25 [84.72, 91.05] \\
G12 & T & 12 & 323/400 & 80.75 [76.60, 84.31] \\
G12 & T & 16 & 285/400 & 71.25 [66.63, 75.47] \\
\end{longtable}\endgroup

\begingroup\normalsize\setlength{\tabcolsep}{4pt}\renewcommand{\arraystretch}{1.08}
\begin{longtable}{lrrr}
\caption{Complete isolated-edge recovery per graph. A success requires every field of every edge to be correct. The graph, rather than an edge query, is the sampling unit.}\label{tab:app_ci_path_local}\\
\toprule\rowcolor{meiSage} Model & Edges & Correct graphs/$n$ & Recovery [95\% CI] \\
\midrule\endfirsthead
\toprule\rowcolor{meiSage} Model & Edges & Correct graphs/$n$ & Recovery [95\% CI] \\
\midrule\endhead\bottomrule\endfoot
Q3 & 4 & 400/400 & 100.00 [99.05, 100.00] \\
Q3 & 8 & 400/400 & 100.00 [99.05, 100.00] \\
Q3 & 12 & 400/400 & 100.00 [99.05, 100.00] \\
Q3 & 16 & 400/400 & 100.00 [99.05, 100.00] \\
Q7 & 4 & 400/400 & 100.00 [99.05, 100.00] \\
Q7 & 8 & 400/400 & 100.00 [99.05, 100.00] \\
Q7 & 12 & 400/400 & 100.00 [99.05, 100.00] \\
Q7 & 16 & 400/400 & 100.00 [99.05, 100.00] \\
G4 & 4 & 400/400 & 100.00 [99.05, 100.00] \\
G4 & 8 & 400/400 & 100.00 [99.05, 100.00] \\
G4 & 12 & 400/400 & 100.00 [99.05, 100.00] \\
G4 & 16 & 400/400 & 100.00 [99.05, 100.00] \\
G12 & 4 & 400/400 & 100.00 [99.05, 100.00] \\
G12 & 8 & 400/400 & 100.00 [99.05, 100.00] \\
G12 & 12 & 400/400 & 100.00 [99.05, 100.00] \\
G12 & 16 & 400/400 & 100.00 [99.05, 100.00] \\
\end{longtable}\endgroup

\section{Relation to the Main Conclusions}
\label{app:relation_main}
The supplementary evidence follows the same sequence as the main analysis. Local and conditional breakdowns characterize the gap between recovering facts and producing a valid answer. Retrieval and constraint diagnostics identify task requirements associated with that gap. The controlled path task establishes a second setting with the same behavioral separation. Probes then distinguish solution membership in global-prompt states from prediction of decision correctness. Finally, paired replacement measures how changes to those states alter failed decisions across depth and prompt positions.

The empirical claim concerns the connection between recoverable evidence, decodable solution information, and successful decisions under these controlled tasks. The path construction, changing acceptance sets, distinct model-specific intervention cohorts, and finite layer grids specify the scope of that claim. Together, the main and supplementary results motivate evaluating evidence recovery and evidence use separately, and studying the internal computations that connect them.

\section{Prompting Controls}
\label{app:prompting_controls}

Before introducing deterministic execution, we test whether additional
instructions alone improve global decisions. All four conditions use the same
400 visual-grid instances for each model and candidate count. \emph{Direct}
requests the selected item IDs in one response. \emph{Generic Review} asks the
model to reconsider its initial answer. \emph{Extract--Select} (ES) first asks
the model to transcribe every card into structured JSON and then select a
subset from that transcription. \emph{Extract--Select--Verify} (ESV) adds an
explicit check of the numerical, categorical, and compatibility requirements
before the final answer.

Table~\ref{tab:prompting_controls} shows that structured prompting alone
produces uneven changes. ES improves the $k=4$ result for both Qwen
checkpoints, while its $k=6$ and $k=8$ accuracy remains close to Direct.
Explicit verification produces similar results. Generic Review helps both
Gemma checkpoints in several conditions, while its effect is model-dependent
and frequently yields responses outside the required output format. These
comparisons motivate separating evidence transcription from joint execution
in the Extract--Solve analysis.

\begin{table}[t]
\centering
\begingroup
\normalsize
\definecolor{meiSage}{HTML}{E5EFEB}
\definecolor{meiBlue}{HTML}{648CAF}
\definecolor{meiRose}{HTML}{BD7B89}
\definecolor{meiRule}{HTML}{A9B6BA}
\arrayrulecolor{meiRule}
\setlength{\tabcolsep}{7pt}
\renewcommand{\arraystretch}{1.08}
\caption{\textbf{Prompting controls produce model-dependent changes without
consistently resolving difficult global decisions.}
Entries are exact globally valid answers (\%) over 400 paired visual-grid
instances. ES denotes Extract--Select and ESV denotes
Extract--Select--Verify.}
\label{tab:prompting_controls}
\begin{tabular}{@{}llrrrr@{}}
\toprule
\rowcolor{meiSage}
Model & $k$ & Direct & Review & \textcolor{meiBlue}{ES} &
\textcolor{meiRose}{ESV} \\
\midrule
Qwen2.5-VL-3B
& 4 & 53.25 & 7.00 & 62.25 & 59.25 \\
& 6 & 17.50 & 2.75 & 16.50 & 15.50 \\
& 8 &  5.75 & 1.50 &  5.25 &  5.25 \\
\addlinespace
Qwen2.5-VL-7B
& 4 & 71.50 & 0.75 & 74.75 & 74.25 \\
& 6 & 28.75 & 0.25 & 26.25 & 26.25 \\
& 8 & 11.75 & 0.00 & 10.50 & 10.50 \\
\addlinespace
Gemma-3-4B
& 4 & 53.25 & 68.75 & 48.50 & 48.75 \\
& 6 & 16.25 & 25.00 & 11.25 & 11.75 \\
& 8 & 10.25 &  8.00 &  3.00 &  3.25 \\
\addlinespace
Gemma-3-12B
& 4 & 72.25 & 87.75 & 76.50 & 73.75 \\
& 6 & 24.75 & 61.75 & 24.25 & 23.25 \\
& 8 & 14.00 & 30.50 & 10.25 & 10.50 \\
\bottomrule
\end{tabular}
\endgroup
\end{table}

\IfFileExists{appendix_generated/reasoning_control.tex}{%
\subsection{Explicit Reasoning Budget}
\label{app:reasoning_control}

The preceding prompting controls vary response structure within the original checkpoints. We next test whether additional inference-time computation changes the difficult eight-candidate result. The evaluation uses 195 visual-grid instances shared by every condition (52 for conflict 0 and distractor ratio 0, 52 for conflict 0 and distractor ratio 0.25, 52 for conflict 1 and distractor ratio 0, 39 for conflict 1 and distractor ratio 0.25). Qwen3-VL-8B-Instruct and Qwen3-VL-8B-Thinking receive the same images and constraints. Each checkpoint is evaluated once with answer-only decoding and once with a two-stage protocol: the first stage reasons through every candidate and constraint, and the second stage emits the final JSON answer. Sampling is disabled in every condition. The paired sample design isolates the change associated with explicit reasoning within each checkpoint.

\begin{table}[t]
\centering
\begingroup
\normalsize
\definecolor{meiSage}{HTML}{E5EFEB}
\definecolor{meiBlue}{HTML}{648CAF}
\definecolor{meiRose}{HTML}{BD7B89}
\definecolor{meiRule}{HTML}{A9B6BA}
\arrayrulecolor{meiRule}
\setlength{\tabcolsep}{5pt}
\renewcommand{\arraystretch}{1.10}
\caption{\textbf{Explicit reasoning changes global validity on paired eight-candidate visual instances.}
Valid reports exact globally valid answers with pointwise Wilson 95\% intervals. Final JSON is the share of responses containing a parseable item-set object. At cap is the share whose final generation reaches its token budget.}
\label{tab:reasoning_control}
\begin{tabular}{@{}llrrrr@{}}
\toprule
\rowcolor{meiSage}
Checkpoint & Condition & $n$ & \shortstack{Valid [95\% CI]} & \textcolor{meiBlue}{Final JSON} & \textcolor{meiRose}{At cap} \\
\midrule
Qwen3-VL-8B Instruct & Answer only & 195 & 5.64 [3.18, 9.82] & 99.49 & 0.00 \\ 
 & Explicit reasoning & 195 & 48.21 [41.29, 55.19] & 53.33 & 0.00 \\ 
Qwen3-VL-8B Thinking & Answer only & 195 & 0.00 [0.00, 1.93] & 0.00 & 100.00 \\ 
 & Explicit reasoning & 195 & 0.51 [0.09, 2.85] & 0.51 & 100.00 \\ 
\bottomrule
\end{tabular}
\endgroup
\end{table}

\begin{figure}[t]
\centering
\includegraphics[width=0.94\textwidth]{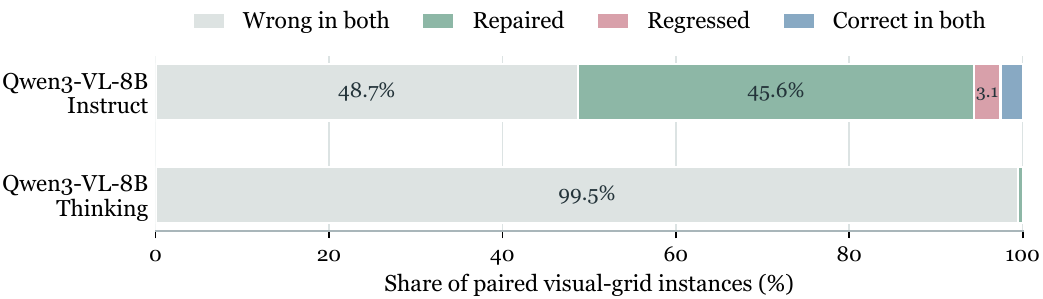}
\caption{\textbf{Explicit reasoning repairs some decisions and can also overturn valid answers.}
Each horizontal bar partitions the same 195 instances by their paired answer-only and explicit-reasoning outcomes. Repaired denotes an invalid answer that becomes globally valid; regressed denotes a valid answer that becomes invalid. Qwen3-VL-8B-Instruct and Qwen3-VL-8B-Thinking use identical instances and deterministic decoding.}
\label{fig:reasoning_control}
\end{figure}

\paragraph{What Additional Computation Changes}
Qwen3-VL-8B Instruct moves from 5.64\% with answer-only decoding to 48.21\% with explicit reasoning. The paired gain is +42.56 percentage points, with a 95\% interval of [+34.87, +50.26]. Figure~\ref{fig:reasoning_control} shows the sample-level source of this change: 89 previously incorrect decisions become valid and 6 previously valid decisions become invalid. Explicit reasoning still leaves 101 of 195 decisions invalid, corresponding to 51.79\% of the paired cohort.

The Thinking checkpoint reaches the 2,048-token answer budget on all 195 answer-only runs and the 256-token finalization budget on all 195 two-stage runs. Its final JSON rate is 0.00\% for answer-only decoding and 0.51\% after explicit reasoning. The resulting valid-answer rates of 0.00\% and 0.51\% characterize budget saturation and completion under this fixed protocol. The Instruct comparison supplies the completed paired test of additional reasoning, while the Thinking checkpoint exposes generation budget as a separate systems constraint.

\paragraph{Implication for the Integration Gap}
The 42.56-point Instruct gain establishes that inference procedure contributes substantially to global validity. The remaining 51.79\% error rate shows that a sizeable integration gap survives explicit step-by-step computation on the same visual evidence. Reasoning budget thus joins visual presentation, model scale, and constraint structure as a variable that shapes multi-evidence integration. Future evaluations can separate reasoning quality from completion by scaling both deliberation and finalization budgets while preserving the paired local-to-global protocol.
}{}

\section{A Constructive Direction: Extract--Solve}
\label{app:extract_solve}

The behavioral experiments reveal a persistent separation between recovering individual visual facts and combining those facts into a globally valid decision. Appendix~\ref{app:prompting_controls} shows that review, structured extraction, and explicit verification produce model-dependent gains while difficult global decisions remain unresolved. These findings motivate a direct division of labor: the vision-language model converts the image into structured evidence, and a deterministic solver performs the joint constraint computation.

\paragraph{Extract--Solve Pipeline}
For each instance, the model first transcribes every card into a JSON record containing its item ID, cost, weight, value, color, type, and compatibility relations. We reuse the transcription generated by the Extract--Select condition, giving 400 paired instances for every model and candidate count. The resulting evaluation contains 4,800 transcriptions across four models and three values of $k$.

A format audit identified several equivalent representations of compatibility relations. Models expressed the same set as a JSON list, a comma-separated string, a single item ID, or an empty-set marker such as \texttt{None} or \texttt{null}. The semantic parser maps these forms into a common set representation. Records with missing cards, malformed JSON, unknown item IDs, invalid attributes, duplicate relations, or inconsistent compatibility fields receive a failed extraction outcome. We report this semantic parsing analysis as an exploratory evaluation specified after the format audit.

The deterministic component receives the structured transcription and a machine-readable copy of the constraints displayed in the original prompt. It enumerates all feasible subsets and emits an answer when the extracted instance yields one unique solution. Zero-solution and multiple-solution cases receive separate outcome labels. Gold metadata enters after prediction for scoring and for measuring exact transcription accuracy.

\paragraph{Deterministic Integration Recovers Difficult Decisions}
Table~\ref{tab:extract_solve} compares Extract--Solve with the same-runtime Direct and Extract--Select (ES) conditions defined in Appendix~\ref{app:prompting_controls}. The largest gains appear at the candidate counts where direct global decisions are weakest. Qwen2.5-VL-3B rises from 5.75\% to 92.25\% at $k=8$, Qwen2.5-VL-7B rises from 11.75\% to 96.75\%, and Gemma-3-12B rises from 14.00\% to 89.50\%. At $k=6$, Qwen2.5-VL-7B reaches 94.50\% and Gemma-3-12B reaches 100.00\%. Their paired improvements over Direct range from 64.75 to 86.50 percentage points across these harder conditions.

\begin{table*}[t]
\centering
\begingroup
\normalsize
\definecolor{meiSage}{HTML}{E5EFEB}
\definecolor{meiBlue}{HTML}{648CAF}
\definecolor{meiRose}{HTML}{BD7B89}
\definecolor{meiRule}{HTML}{A9B6BA}
\arrayrulecolor{meiRule}
\setlength{\tabcolsep}{5pt}
\renewcommand{\arraystretch}{1.10}
\caption{\textbf{Extract--Solve converts usable transcriptions into globally valid decisions.}
All entries are percentages over 400 paired instances.
ES denotes Extract--Select.
Brackets beside Extract--Solve report pointwise Wilson 95\% intervals.
The final column reports 95\% intervals for the paired difference from Direct,
estimated with 10,000 bootstrap draws stratified by conflict intensity and distractor ratio.}
\label{tab:extract_solve}
\begin{tabular}{@{}llrrlrl@{}}
\toprule
\rowcolor{meiSage}
Model &
$k$ &
Direct &
ES &
\textcolor{meiBlue}{\shortstack{Extract--Solve\\{[95\% CI]}}} &
\textcolor{meiRose}{$\Delta$ Direct} &
95\% CI for $\Delta$ \\
\midrule
Qwen2.5-VL-3B
& 4 & 53.25 & 62.25 & 36.50 [31.93, 41.33] & $-16.75$ & [$-24.00,-9.50$] \\
& 6 & 17.50 & 16.50 & 82.25 [78.20, 85.68] & $+64.75$ & [$+59.25,+70.00$] \\
& 8 &  5.75 &  5.25 & 92.25 [89.21, 94.49] & $+86.50$ & [$+82.50,+90.01$] \\
\addlinespace
Qwen2.5-VL-7B
& 4 & 71.50 & 74.75 & 84.00 [80.09, 87.27] & $+12.50$ & [$+7.00,+18.00$] \\
& 6 & 28.75 & 26.25 & 94.50 [91.81, 96.34] & $+65.75$ & [$+61.25,+70.50$] \\
& 8 & 11.75 & 10.50 & 96.75 [94.52, 98.09] & $+85.00$ & [$+81.25,+88.50$] \\
\addlinespace
Gemma-3-4B
& 4 & 53.25 & 48.50 &  6.50 [4.47, 9.35] & $-46.75$ & [$-52.00,-41.50$] \\
& 6 & 16.25 & 11.25 &  2.00 [1.02, 3.90] & $-14.25$ & [$-18.25,-10.50$] \\
& 8 & 10.25 &  3.00 &  0.50 [0.14, 1.80] &  $-9.75$ & [$-13.00,-6.75$] \\
\addlinespace
Gemma-3-12B
& 4 & 72.25 & 76.50 &  99.50 [98.20, 99.86] & $+27.25$ & [$+23.00,+31.50$] \\
& 6 & 24.75 & 24.25 & 100.00 [99.05, 100.00] & $+75.25$ & [$+71.00,+79.50$] \\
& 8 & 14.00 & 10.25 &  89.50 [86.11, 92.14] & $+75.50$ & [$+71.00,+79.75$] \\
\bottomrule
\end{tabular}
\endgroup
\end{table*}

\begin{figure*}[t]
\centering
\includegraphics[width=\textwidth]
{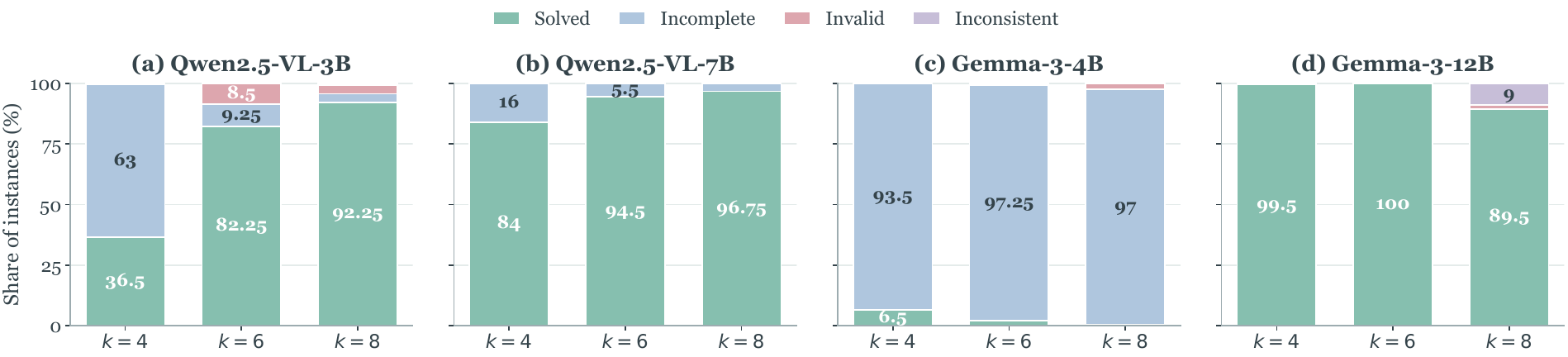}
\caption{\textbf{Outcome decomposition of the semantic Extract--Solve pipeline.}
Each bar contains 400 instances.
Solved denotes transcriptions that yield one unique feasible solution; all 3,137 instances in this category match the benchmark target.
Incomplete denotes missing item cards, Invalid covers malformed JSON or invalid fields,
and Inconsistent combines zero- and multiple-solution outcomes.
Q3 and Q7 denote Qwen2.5-VL-3B and Qwen2.5-VL-7B;
G4 and G12 denote Gemma-3-4B and Gemma-3-12B.}
\label{fig:extract_solve_outcomes}
\end{figure*}

\paragraph{Usable Transcriptions Lead to Correct Solutions}
Figure~\ref{fig:extract_solve_outcomes} separates successful solutions from extraction and solver outcomes. Across all models and candidate counts, 3,137 transcriptions yield one unique feasible solution. Every one of these solutions matches the benchmark target, giving 3,137 correct predictions out of 3,137 solved instances. The remaining outcomes comprise 1,553 incomplete card sets, 71 malformed or invalid records, and 39 transcriptions that produce zero or multiple feasible solutions.

\paragraph{Extraction Completeness Determines End-to-End Performance}
The model-level differences are concentrated in transcription completeness. Gemma-3-12B supplies a usable record for nearly every $k=4$ and $k=6$ instance, supporting 99.50\% and 100.00\% end-to-end accuracy. Qwen2.5-VL-7B also maintains high completion rates and reaches 84.00--96.75\% across candidate counts.

Gemma-3-4B lists a subset of the available cards in 1,151 of its 1,200 instances, producing end-to-end accuracies between 0.50\% and 6.50\%. Qwen2.5-VL-3B shows a related completeness pattern at $k=4$, where 252 of 400 transcriptions omit at least one card. Every incomplete response ends before the 768-token generation cap. The model terminates its structured response while cards remain unlisted, identifying extraction completeness as an interface-level failure.

\paragraph{Decision-Relevant Evidence}
Exact transcription of every field sets a stricter requirement than successful selection. For Gemma-3-12B at $k=8$, 61.50\% of transcriptions match every evaluated field, while Extract--Solve reaches 89.50\% selection accuracy. Some attribute and compatibility errors concern cards or relations outside the final feasible subset. The unique valid decision remains recoverable when those errors preserve the constraints that determine the target.

This pattern clarifies the role of extracted evidence. Global success depends on preserving the facts that define the feasible solution, while peripheral transcription errors can leave the final decision unchanged. The solver supplies an explicit test of whether the available transcription still supports a unique globally valid answer.

\paragraph{From Diagnosis to a Constructive Design Principle}
Extract--Solve turns the observed integration gap into a concrete system design. A reliable visual-to-structured interface supplies the evidence, and deterministic computation enforces joint consistency across that evidence. This combination nearly eliminates the difficult $k=6$ and $k=8$ failures for Qwen2.5-VL-7B and Gemma-3-12B. The Gemma-3-4B results identify extraction completeness as the next engineering target.

These results establish a practical path for multimodal systems that operate over multiple pieces of evidence: visual models can specialize in faithful structured recovery, while explicit execution modules carry out globally constrained decisions. This decomposition gives each component a clear responsibility, exposes its failures directly, and converts the local-to-global integration gap into an actionable system objective.

\section{Discussion}
\label{sec:discussion}

\paragraph{The Missing Step Between Perception and Reasoning}
Multimodal evaluation commonly treats the final answer as the primary unit of success, while compositional benchmarks reveal failures involving relations and combinations
\citep{yue2024mmmu,wang2025muirbench,liu2024mibench,meng2025mmiu, yuksekgonul2022and,ma2023crepe}.
Our results identify a more precise unit of analysis: the transformation from individually recoverable evidence into a decision that satisfies all constraints jointly.
This transformation deserves separate evaluation because local competence and global validity diverge sharply.
The gap survives same-grid local questioning, appears across visual and textual inputs, persists in larger checkpoints, and extends from subset selection to weighted-path reasoning.
These results locate a general challenge in coordinating available evidence during decision formation.

This perspective changes how multimodal capability should be measured. A model can identify every relevant object, recover its attributes, retrieve a specified target set, and still fail when several facts must jointly determine one answer. Evaluation based on constituent recognition can assign high competence to systems whose decisions remain globally inconsistent. Conditional local-to-global evaluation exposes this difference directly by asking how often demonstrated evidence recovery culminates in a valid decision. It provides a diagnostic axis that complements broader measures of perception, knowledge, retrieval, and end-to-end reasoning.

\paragraph{A Mechanistic Account Across Three Levels}
The internal analyses connect three levels of evidence.
First, local queries establish that individual facts are behaviorally recoverable.
Second, membership probes show that global-prompt hidden states contain information about which candidates belong to the valid solution.
Third, matched clean-state replacement recovers failed decisions across substantial ranges of network depth.
The combination reveals a progression from available facts, to solution-related representations, to decision sensitivity under intervention.

This progression distinguishes our mechanism analysis from studies centered on probe accuracy or activation replacement alone. Probe decodability measures information accessible to an external readout.
Matched and mismatched donors test whether the identity of the replacement state changes the generated decision.
Their agreement shows that solution-related structure exists in the computation and that corresponding clean states can redirect corrupted decisions.
Their separation shows that accessible solution information may remain disconnected from the model's final action.
The central mechanism is a conversion problem: internal states can carry useful evidence and solution membership while the generation process still produces a constraint-violating subset.

The depth and token-scope results further characterize this conversion.
Early image-token replacement produces strong recovery, while text-position replacement becomes effective later in Qwen2.5-VL-3B.
This profile is consistent with visual evidence becoming distributed across non-image prompt positions through successive attention layers.
Full-prompt replacement produces the strongest recovery near the later effective blocks, indicating that successful decisions depend on coordinated states across prompt positions.
The model-specific recovery ranges add a second insight: architectures organize recoverable decision states differently.
InternVL retains matched-donor effects deeper than Gemma, while larger Qwen and Gemma checkpoints change their recovery profiles in different directions.
Parameter count improves some behavioral outcomes, while architecture and training shape how long corrective information remains usable.

\paragraph{From Mechanistic Diagnosis to System Design}
The exploratory Extract--Solve analysis converts the diagnosis into a concrete design principle.
The vision-language model transcribes the visual evidence, and a deterministic component executes the joint constraints.
Across all evaluated cohorts, every transcription that yields a unique feasible solution produces the correct benchmark answer, covering 3,137 out of 3,137 solved instances.
For Qwen2.5-VL-7B and Gemma-3-12B, this pipeline raises difficult $k=6$ and $k=8$ conditions to 89.50--100.00\% accuracy.
The result demonstrates that explicit execution can transform available structured evidence into reliable global decisions.
The remaining errors reveal the next systems challenge.
Gemma-3-4B frequently terminates after listing a subset of the cards, and Qwen2.5-VL-3B shows a similar completeness pattern at $k=4$.
The bottleneck has moved from joint constraint execution to the reliability of the visual-to-structured interface.
This decomposition creates two measurable objectives: faithful and complete evidence extraction, followed by verifiable global execution.
It also suggests architectures in which multimodal representations feed explicit planning, optimization, or symbolic modules through a structured intermediate state.

\paragraph{Implications for Future Research}
The findings motivate a research agenda organized around evidence integration.
Benchmarks can pair local queries with globally verifiable decisions and report conditional global validity alongside aggregate accuracy.
Training objectives can reward consistency across all relevant facts, preserve solution-related information through generation, and penalize answers that violate jointly active constraints.
Architectural work can develop structured interfaces that expose evidence to execution modules while retaining uncertainty about ambiguous visual fields.
Interpretability studies can connect behavioral controls, probe targets, matched counterfactual donors, token scopes, and downstream decisions within one experimental chain.

The broader lesson concerns what it means for a multimodal model to reason successfully.
Recognition supplies the pieces of a decision; integration determines whether those pieces jointly support the action the model takes.
Our study makes this transition observable at the behavioral, representational, causal, and system levels.
Treating evidence integration as a first-class capability opens a path toward models whose final decisions remain consistent with the information they already possess.

\section{Extended Related Work}
\label{app:extended_related_work}

\paragraph{Compositional Generalization in Vision--Language Models}
Recent work directly tests whether vision--language models can compose skills
that are learned or evaluated separately. Li et al. train models on isolated
unimodal or reasoning capabilities and evaluate out-of-distribution
cross-modal and cross-task combinations \citep{li2025unveiling}. Their results
show a compositional ability gap between individual skills and their joint
use, and identify explicit visual description and progressive grounding as
useful training signals. Related studies evaluate sensitivity to
object--attribute binding, relations, spatial configurations, and controlled
recombination \citep{yuksekgonul2022and,ma2023crepe,kamath2023s}.

Our study examines composition within one exactly verifiable decision. Each
instance contains several visual facts whose numerical, categorical, and
relational constraints jointly determine a valid subset. Conditional
same-grid evaluation measures global validity after the same model has
successfully recovered every constituent fact. This setting connects local
evidence recovery, exact constraint satisfaction, candidate-level solution
information, clean-state replacement, and explicit execution without a
training phase between the constituent and composed tasks.

\paragraph{Evidence Use in Long and Multimodal Contexts}
Long-context evaluations reveal sensitivity to evidence position and
difficulty retrieving or combining distributed information
\citep{liu2024lost,bai2024longbench,wang2024leave}. NoLiMa targets retrieval
beyond literal matching \citep{modarressi2025nolima}, while MileBench studies
long-context multimodal understanding \citep{song2024milebench}. Multi-image
benchmarks similarly measure comparison, relational reasoning, and reasoning
across image collections and sequences
\citep{wang2025muirbench,liu2024mibench,meng2025mmiu,wang2024mementos}.
Our matched-retrieval control tests whether the target cards can be located
under the same visual presentation, and the conditional analysis asks whether
successful evidence recovery culminates in a jointly valid decision.

\paragraph{Probing and Causal Intervention}
Probes test which variables are accessible from hidden representations, while
activation patching tests whether replacing internal states changes model
behavior. Prior studies apply these tools to factual recall, latent reasoning,
and visual information processing
\citep{geva2023dissecting,yang2024large,neo2025towards}. Work on patching
methodology also highlights the importance of donor construction, intervention
scope, and controls for learned subspaces
\citep{zhang2024towards,makelov2024subspace}. Our analysis combines
candidate-level membership probes with matched and mismatched donors, depth
comparisons, and token-scope interventions. The supplementary generator-group
and construction-baseline controls test whether membership decoding extends
beyond fixed candidate frequencies and benchmark-generation features.

\end{document}